%% file: main.tex
\documentclass{article}

\usepackage{xcolor}

\newcommand{\best}[1]{%
  \begingroup
  \setlength{\fboxsep}{1.0pt}%
  \colorbox{Blue_1!18}{\strut\textbf{#1}}%
  \endgroup
}
\newcommand{\secondbest}[1]{\textbf{\underline{#1}}}

\usepackage{dojo_report}
\usepackage{algpseudocode}
\usepackage[utf8]{inputenc}
\usepackage[T1]{fontenc}
\usepackage{hyperref}
\usepackage{url}
\usepackage{booktabs}
\usepackage{amsfonts}
\usepackage{nicefrac}
\usepackage{amsmath}
\usepackage{microtype}
\usepackage{lipsum}
\usepackage{duckuments}
\usepackage{adjustbox}
\usepackage{array}
\usepackage{subcaption}
\usepackage{amsmath}
\usepackage{float}
\usepackage{multirow}
\usepackage{enumitem}
\usepackage{listings}
\usepackage{algpseudocode}
\usepackage{bibunits}
\usepackage{wrapfig}
\usepackage{changepage}
\usepackage{makecell}
\usepackage{wrapfig}
\usepackage{graphicx}
\usepackage{amssymb}
\usepackage{mathtools}
\usepackage{amsthm}
\usepackage{algorithm}
\usepackage{tocloft}
\usepackage{cleveref}
\usepackage{booktabs}
\usepackage{xspace}
\usepackage{fontawesome5}
\usepackage{booktabs}
\usepackage{graphicx}
\usepackage{pifont}
\usepackage{tabularx}
\usepackage{geometry}
\usepackage{tikz}

\usepackage{titlesec}

\titlespacing{\paragraph}{0pt}{0.35em}{0.6em}

\input{math_commands}

\definecolor{cornellred}{rgb}{0.7, 0.11, 0.11}
\definecolor{cadmiumgreen}{rgb}{0.0, 0.42, 0.24}
\definecolor{aliceblue}{rgb}{0.91, 0.94, 0.97}
\definecolor{darkblue}{rgb}{0.83, 0.89, 0.97}
\definecolor{Red7}{rgb}{0.941, 0.243, 0.243}
\definecolor{Green7}{RGB}{55, 178, 77}
\definecolor{Blue9}{rgb}{0.098,0.3,0.9}
\definecolor{SJViolet}{RGB}{105,100,171}
\definecolor{SJRed}{RGB}{237,109,107}

\usepackage[most]{tcolorbox}

\usepackage{array}
\usepackage{etoolbox}
\usepackage[most]{tcolorbox}

\usepackage{graphicx}
\usepackage{array}
\usepackage{etoolbox}
\usepackage[most]{tcolorbox}
\usepackage{tikz}
\newsavebox{\roundedimgbox}
\newcommand{\roundedimage}[2][5pt]{%
  \sbox{\roundedimgbox}{\includegraphics[width=\linewidth]{#2}}%
  \begin{tikzpicture}
    \clip[rounded corners=#1] (0,0) rectangle (\wd\roundedimgbox,\ht\roundedimgbox);
    \node[anchor=south west, inner sep=0pt, outer sep=0pt] at (0,0) {\usebox{\roundedimgbox}};
  \end{tikzpicture}%
}

\newcolumntype{C}[1]{>{\centering\arraybackslash}m{#1}}
\newcolumntype{L}[1]{>{\raggedright\arraybackslash}m{#1}}

\definecolor{DojoBlue}{HTML}{498ae5}
\definecolor{LinkBlue}{HTML}{0000ff}
\definecolor{Blue_1}{HTML}{1c6ef3}
\definecolor{Blue_2}{HTML}{d2e3fc}
\definecolor{Dojo_Yellow}{HTML}{ffdf7b}

\newsavebox{\taskimgbox}

\newcommand{\roundedtaskimage}[2][5pt]{%
  \sbox{\taskimgbox}{\includegraphics[width=0.95\linewidth]{#2}}%
  \begin{tikzpicture}
    \clip[rounded corners=#1]
      (0,0) rectangle (\wd\taskimgbox,\ht\taskimgbox);
    \node[
      anchor=south west,
      inner sep=0pt,
      outer sep=0pt
    ] at (0,0) {\usebox{\taskimgbox}};
  \end{tikzpicture}%
}

\newif\ifTaskDataSourcePending
\newcommand{\TaskDataSourceText}{}

\newcommand{\flushtaskdatasource}{%
    \ifTaskDataSourcePending
        \noindent\TaskDataSourceText\par\vspace{0.12em}%
        \TaskDataSourcePendingfalse
    \fi
}

\newcommand{\taskitem}[2]{%
    \ifstrequal{#1}{Data Source}{%
        \gdef\TaskDataSourceText{\textbf{#1}: #2}%
        \TaskDataSourcePendingtrue
    }{%
        \ifstrequal{#1}{Usage}{%
            \ifTaskDataSourcePending
                \noindent
                \TaskDataSourceText
                \hspace{2em}
                \textbf{#1}: #2
                \par\vspace{0.12em}%
                \TaskDataSourcePendingfalse
            \else
                \noindent\textbf{#1}: #2\par\vspace{0.12em}%
            \fi
        }{%
            \flushtaskdatasource
            \noindent\textbf{#1}: #2\par\vspace{0.12em}%
        }%
    }%
}

\newcommand{\taskshowcase}[3]{%
\begin{tcolorbox}[
    colback=white,
    colframe=black!12,
    boxrule=0.45pt,
    arc=1.5pt,
    left=4pt,
    right=4pt,
    top=4pt,
    bottom=4pt,
    boxsep=0pt,
    before skip=0.2em,
    after skip=0.25em
]
\noindent
\begin{tabular}{@{} C{0.20\linewidth} @{\hspace{0.018\linewidth}} L{0.782\linewidth} @{}}

\roundedtaskimage[5pt]{#1}
&
\begin{minipage}[c]{\linewidth}
    \setlength{\parskip}{0pt}
    \setlength{\baselineskip}{0.95\baselineskip}

    {\normalsize\bfseries\textcolor{Blue_1}{#2}}\par\vspace{0.25em}

    {\small
    \TaskDataSourcePendingfalse
    #3
    \flushtaskdatasource
    }
\end{minipage}

\end{tabular}
\end{tcolorbox}
}

\usepackage{listings}
\lstdefinestyle{compactpython}{
    language=Python,
    basicstyle=\ttfamily\scriptsize,
    keywordstyle=\color{blue!70!black}\bfseries,
    commentstyle=\color{green!40!black},
    stringstyle=\color{orange!80!black},
    numbers=none,
    frame=single,
    framerule=0.4pt,
    rulecolor=\color{black!20},
    backgroundcolor=\color{gray!3},
    breaklines=true,
    breakatwhitespace=true,
    showstringspaces=false,
    tabsize=2,
    columns=fullflexible,
    keepspaces=true,
    aboveskip=3pt,
    belowskip=3pt,
    xleftmargin=0pt,
    framexleftmargin=3pt,
    framexrightmargin=3pt,
    framesep=3pt,
    lineskip=-1pt
}

\hypersetup{
  colorlinks = true,
  linkcolor  = LinkBlue,
  citecolor  = LinkBlue,
  urlcolor   = LinkBlue
}

\colorlet{rlTitle}{rlwrldBlack}
\colorlet{rlSectionNum}{Blue_1}
\colorlet{rlHyperlink}{rlwrldDarkGreen}
\colorlet{rlRefLink}{rlwrldDarkGreen}
\colorlet{rlTocLink}{rlwrldBlack}
\definecolor{AbsBlue}{HTML}{498AE5}
\definecolor{AbsCyan}{HTML}{23B399}
\def\abstractstyle{background}
\def\abstractposition{intro}

\def\authordisplay{full}

\def\logotype{woMark}

\title{RoboDojo: A Unified Sim-and-Real Benchmark for Comprehensive Evaluation of Generalist Robot Manipulation Policies}

\ifdefstring{\authordisplay}{full}{%
\author{%
\normalsize \linespread{1.25}\selectfont 

    \textbf{Tianxing Chen}\textsuperscript{1${*}$\S}%
    \quad \textbf{Yue Chen}\textsuperscript{4${*}$\S\dag}%
    \quad \textbf{Zixuan Li}\textsuperscript{3${*}$}%
    \quad \textbf{Junyuan Tang}\textsuperscript{3${*}$}%
    \quad \textbf{Kailun Su}\textsuperscript{3${*}$}%
    \quad \textbf{Haoran Lu}\textsuperscript{4${*}$}%
    \quad \textbf{Weijie Wan}\textsuperscript{3${*}$}%
    \quad \textbf{Baijun Chen}\textsuperscript{1${*}$}%
    \quad \textbf{Songling Liu}\textsuperscript{4${*}$}%
    \quad \textbf{Haowen Yan}\textsuperscript{3}%
    \quad \textbf{Honghao Su}\textsuperscript{3}%
    \quad \textbf{Zhiyang Dou}\textsuperscript{6}%
    \quad \textbf{Kaixuan Wang}\textsuperscript{1}%
    \quad \textbf{Dandan Zhang}\textsuperscript{13}%
    \quad \textbf{Yunze Liu}\textsuperscript{3}%
    \quad \textbf{Yan Qin}\textsuperscript{16}%
    \quad \textbf{Qiwei Liang}\textsuperscript{16}%
    \quad \textbf{Qiwei Wu}\textsuperscript{16}%
    \quad \textbf{Zijian Lin}\textsuperscript{3}%
    \quad \textbf{Wenwei Lin}\textsuperscript{3}%
    \quad \textbf{Yuran Wang}\textsuperscript{10}%
    \quad \textbf{Minghua He}\textsuperscript{4}%
    \quad \textbf{Tianshu Wu}\textsuperscript{4}%
    \quad \textbf{Ruihai Wu}\textsuperscript{4}%
    \quad \textbf{Jingquan Zhou}\textsuperscript{18}%
    \quad \textbf{Kai-Chong Lei}\textsuperscript{3}%
    \quad \textbf{Haibao Yu}\textsuperscript{1}%
    \quad \textbf{Yuanfeng Ji}\textsuperscript{5}%
    \quad \textbf{Weiyang Jin}\textsuperscript{1}%
    \quad \textbf{Guanyu Lin}\textsuperscript{9}%
    \quad \textbf{Xiaofan Li}\textsuperscript{17}%
    \quad \textbf{Qi Xiong}\textsuperscript{3}%
    \quad \textbf{Renjing Xu}\textsuperscript{16}%
    \quad \textbf{Zhongyu Li}\textsuperscript{12}%
    \quad \textbf{Wenhao Chai}\textsuperscript{8}%
    \quad \textbf{Enze Xie}\textsuperscript{1}%
    \quad \textbf{Ziwei Wang}\textsuperscript{11}%
    \quad \textbf{Yao Mu}\textsuperscript{14}%
    \quad \textbf{Hao Dong}\textsuperscript{4}
    \quad \textbf{Wojciech Matusik}\textsuperscript{6}
    \quad \textbf{Mingyu Ding}\textsuperscript{7\dag}%
    \quad \textbf{Wenbo Ding}\textsuperscript{3\dag}%
    \quad \textbf{Ping Luo}\textsuperscript{1\dag}%
    \quad \textbf{Masayoshi Tomizuka}\textsuperscript{2\dag}%
    \par
    {\footnotesize
      \textsuperscript{1}MMLab@HKU
      \quad \textsuperscript{2}UC Berkeley
      \quad \textsuperscript{3}THU
      \quad \textsuperscript{4}PKU
      \quad \textsuperscript{5}Stanford
      \quad \textsuperscript{6}MIT
      \quad \textsuperscript{7}UNC
      \quad \textsuperscript{8}Princeton
      \quad \textsuperscript{9}CMU
      \quad \textsuperscript{10}NUS
      \quad \textsuperscript{11}NTU
      \quad \textsuperscript{12}CUHK
      \quad \textsuperscript{13}IC
      \quad \textsuperscript{14}SJTU
      \quad \textsuperscript{15}NU
      \quad \textsuperscript{16}HKUST (GZ)
      \quad \textsuperscript{17}ZJU
      \quad \textsuperscript{18}Yale
      \quad $^*$Co-first Authors
      \quad \textsuperscript{\S}Co-project Leaders
      \quad \textsuperscript{\dag}Corresponding Authors
      }%
  }%
}{%
  \author{ALINLAB $\times$ RLWRLD}%
}
\begin{document}

\maketitle

\makeatletter
\newcommand{\rlfootertext}[1]{%
  \g@addto@macro\rl@pagefootnotes{%
    \par\noindent #1%
  }%
}
\makeatother

\rlfootertext{}

\begin{figure}[h]
    \centering
    \includegraphics[width=1.0\textwidth]{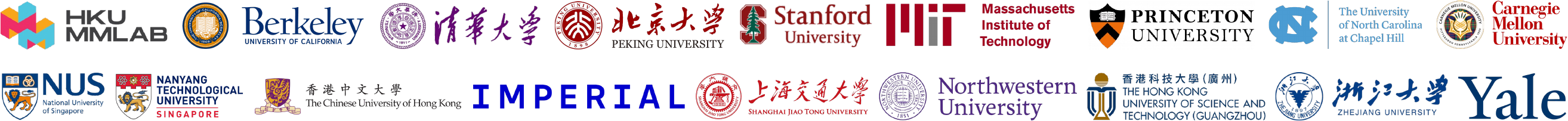}
\end{figure}

\newcommand{\linkicon}[3]{%
    \raisebox{-1.5pt}{\includegraphics[height=1.05em]{#1}}%
    ~\href{#2}{\texttt{#3}}%
}

\begingroup

\noindent
\makebox[\textwidth][c]{%
\begin{tabular}{@{}c@{\hspace{1.1em}}c@{\hspace{1.1em}}c@{\hspace{1.1em}}c@{\hspace{1.1em}}c@{\hspace{1.1em}}c@{}}

\raisebox{-1.5pt}{\includegraphics[height=1.05em]{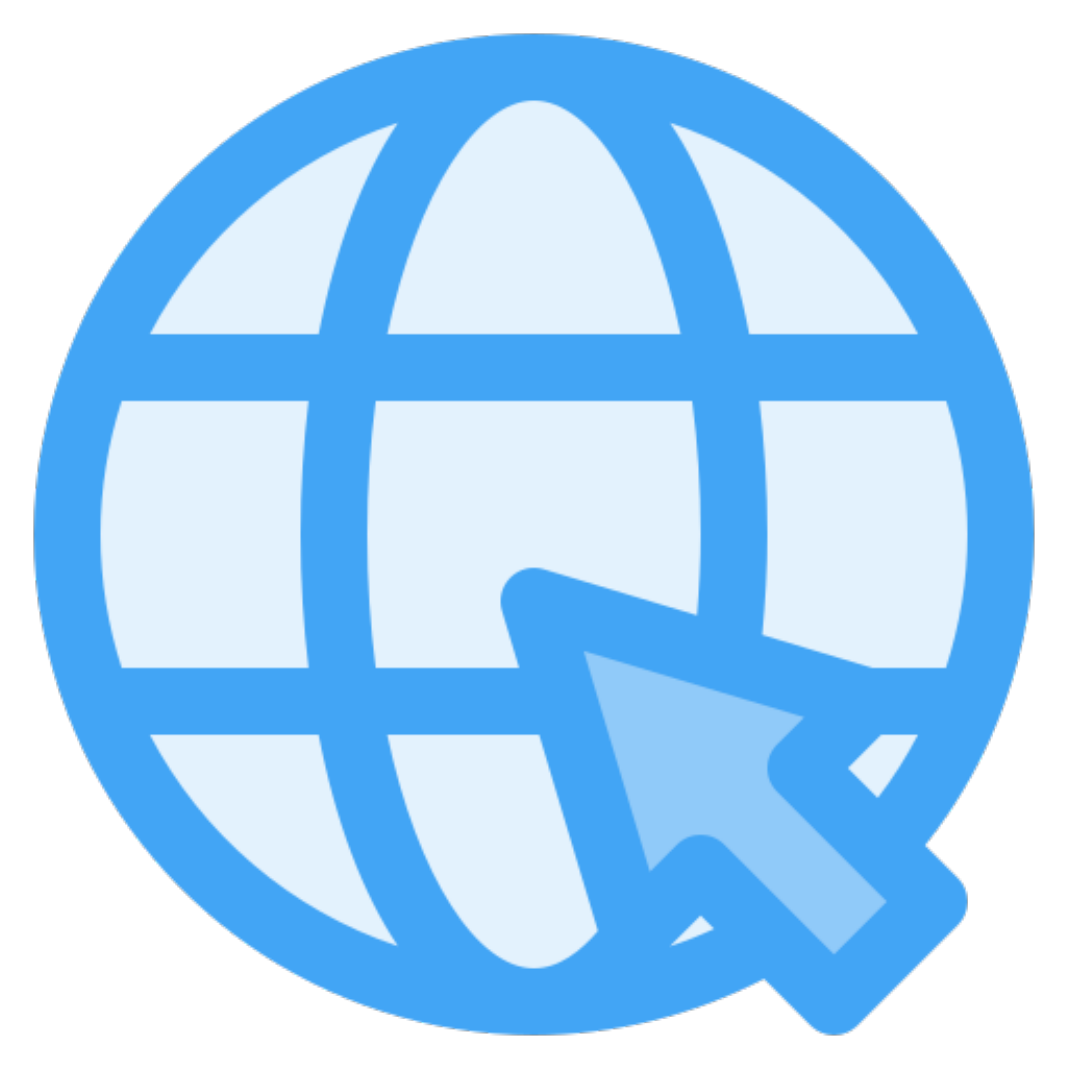}}%
~\textnormal{Website:}~\href{https://RoboDojo-Benchmark.com}{\texttt{RoboDojo-Benchmark.com}}
&
\raisebox{-1.5pt}{\includegraphics[height=1.05em]{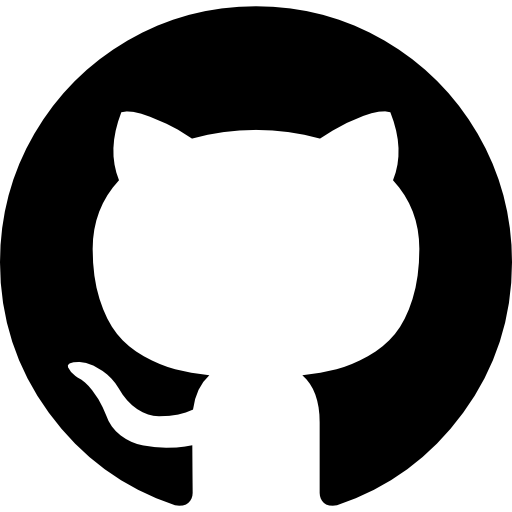}}%
~\textnormal{Code:}~\href{https://github.com/RoboDojo-Benchmark/RoboDojo}{\texttt{Benchmark}},~\href{https://github.com/XPolicyLab/XPolicyLab}{\texttt{XPolicyLab}}
&
\linkicon{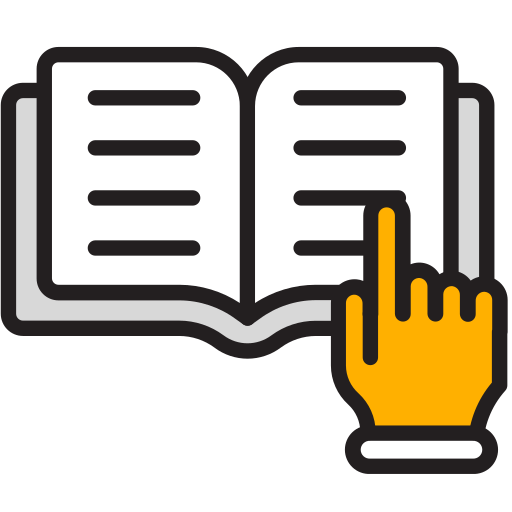}
{http://robodojo-benchmark.com/doc}
{Document}
&
\linkicon{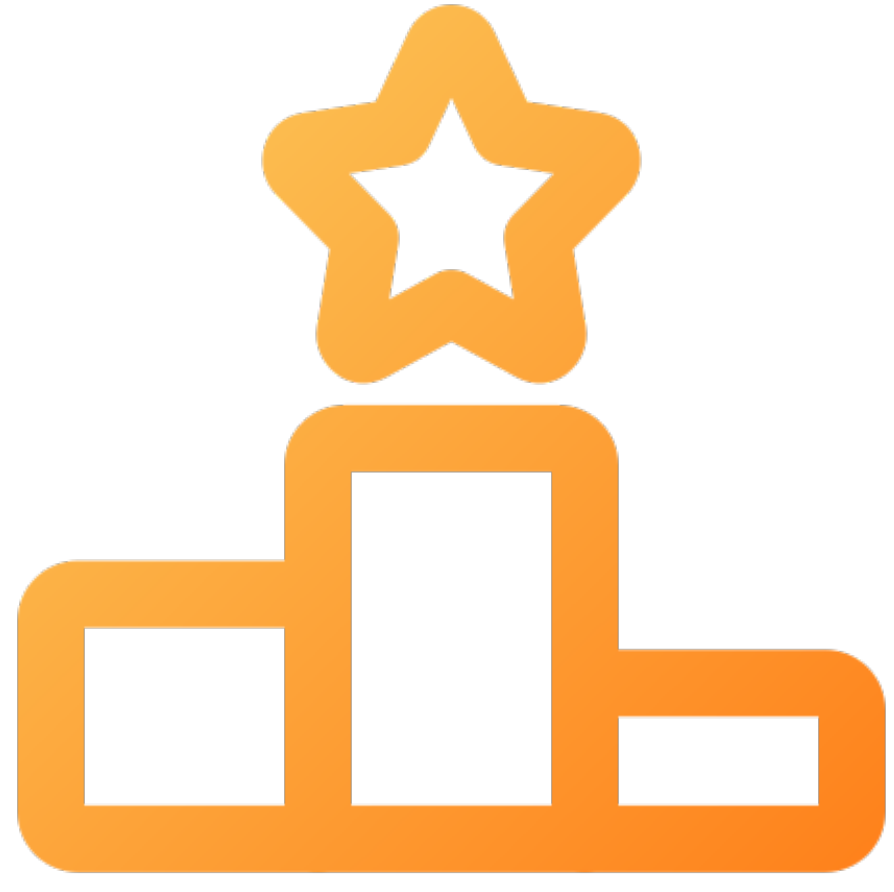}
{https://robodojo-benchmark.com/LeaderBoard}
{Leaderboard}

\end{tabular}%
}
\endgroup



\begin{figure}[h]
    \centering
    \includegraphics[width=0.98\textwidth]{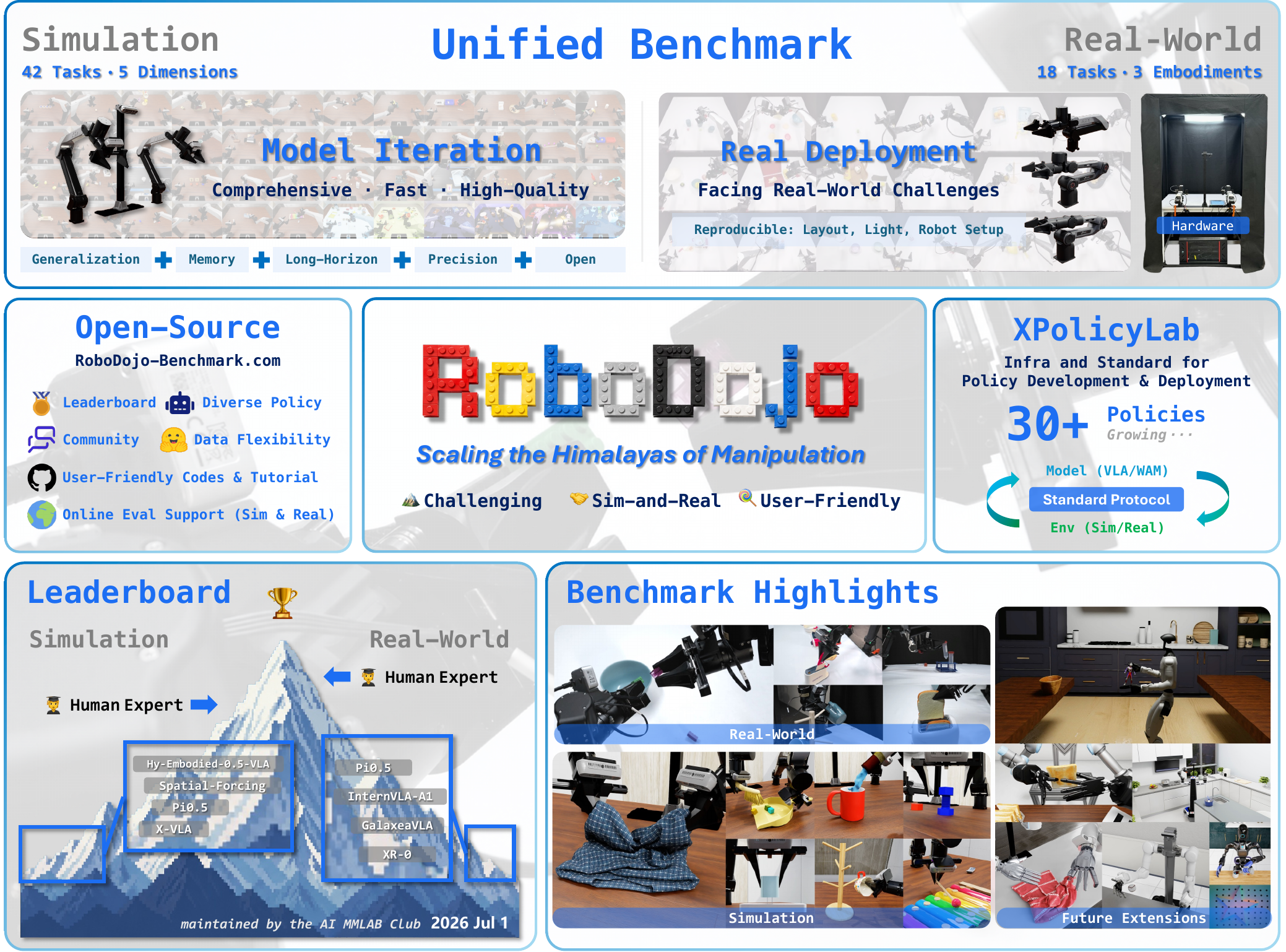}
\caption{\textbf{Overview of RoboDojo.} RoboDojo unifies efficient simulation evaluation and reproducible real-world testing for generalist robot manipulation, covering 42 simulation tasks, 18 real-world tasks, heterogeneous parallel simulation, RoboDojo-RealEval, XPolicyLab, and a continuously updated leaderboard.} 
    \label{fig:teaser}
\end{figure}

\ifdefstring{\abstractposition}{page1}{\input{sec/0_abstract}
}{}

\newpage

{\hypersetup{linkcolor=rlTocLink}\input{sec/index}}

\newpage

\ifdefstring{\abstractposition}{intro}{\input{sec/0_abstract}}{}
\input{sec/1_introduction}
\input{sec/2_related_work}

\input{sec/4_benchmark}
\input{sec/3_method}
\input{sec/5_XPolicyLab}
\input{sec/5.5_leaderboard}
\input{sec/6_experiment}
\input{sec/6.5_future_direction}
\input{sec/7_conclusion}

\makeatletter
\renewcommand\@biblabel[1]{[#1]}
\let\@bibsetup\NAT@bibsetnum
\makeatother

\newpage
\bibliographystyle{plainnat}
\bibliography{reference}

\input{sec/X_appendix}

\end{document}

%% file: math_commands.tex
\usepackage{amsmath,amsfonts,bm}

\def\eqref#1{equation~\ref{#1}}

\def\1{\bm{1}}

\DeclareMathAlphabet{\mathsfit}{\encodingdefault}{\sfdefault}{m}{sl}
\SetMathAlphabet{\mathsfit}{bold}{\encodingdefault}{\sfdefault}{bx}{n}



%% file: sec/0_abstract.tex
\begin{abstract}

Generalist robot manipulation policies have made substantial progress, yet existing benchmarks remain limited in their ability to systematically and comprehensively evaluate policy capabilities. Many benchmarks rely on simple, short-horizon, or skill-narrow tasks that often share similar manipulation patterns and cover only limited capability dimensions. Moreover, evaluations are commonly conducted either in simulation or in the real world alone: simulation provides efficient and scalable feedback but cannot fully capture physical deployment challenges, whereas real-world evaluation offers direct evidence of deployment performance but is costly, time-consuming, and difficult to reproduce. To address these limitations, we introduce \textbf{\textcolor{Blue_1}{RoboDojo}}, a unified sim-and-real benchmark for comprehensive evaluation of generalist robot manipulation policies. RoboDojo includes 42 simulation tasks and 18 real-world tasks designed to cover diverse, challenging, and complementary manipulation capabilities. The simulation benchmark evaluates five capability dimensions: generalization, memory, precision, long-horizon execution, and open-vocabulary instruction following, while the real-world benchmark exposes policies to challenging physical-world deployment conditions. To support large-scale evaluation, RoboDojo implements heterogeneous parallel simulation in Isaac Sim, substantially improving evaluation throughput. For real-world evaluation, RoboDojo provides \textbf{\textcolor{Blue_1}{RoboDojo-RealEval}}, a reproducible real-world evaluation system with remote cloud access. By standardizing the hardware setup, scene reset procedure, evaluation protocol, and deployment interface, RoboDojo-RealEval enables policies to be tested under consistent physical conditions. In parallel, \textbf{\textcolor{Blue_1}{XPolicyLab}} provides a unified infrastructure for policy development and deployment, allowing policies to be integrated once and evaluated across RoboDojo simulation and real-world settings with minimal policy-side adaptation. We integrate 30 policies into XPolicyLab and evaluate them on RoboDojo, establishing a public leaderboard with systematic analysis of current policy performance. The website is available at \href{http://robodojo-benchmark.com/}{http://robodojo-benchmark.com/}.

\vspace{0.6em}
\begin{tcolorbox}[
colback=Blue_1!5,
colframe=Blue_1!55,
boxrule=0.5pt,
arc=2pt,
left=6pt,
right=6pt,
top=4pt,
bottom=4pt
]
\small

\textbf{\textcolor{Blue_1}{Leaderboard Governance.}}
To ensure fair and independent evaluation, AI MMLab Club, a non-profit foundation, maintains the RoboDojo leaderboard. Global academic partners will co-govern and operate it without commercial funding or sponsorship. \end{tcolorbox}

\end{abstract}

%% file: sec/index.tex
\tableofcontents
\thispagestyle{fancy}

%% file: sec/1_introduction.tex
\section{Introduction}
\label{sec:introduction}

A central objective of embodied intelligence is to enable robots to perform manipulation tasks in the physical world. Recent advances in generalist robot policies and embodied foundation models have led to increasingly capable systems across diverse manipulation scenarios~\citep{intelligence2026pi,zhang2026joyaira01foundationmodel,li2026causalworldmodelingrobot,li2025simplevla,zhang2026hy,lyu2026lda,yuan2026qwen}. As these systems continue to advance, reliable evaluation becomes essential for understanding their capabilities, limitations, and remaining failure modes. Existing benchmarks have investigated important aspects of robot manipulation, including language-conditioned long-horizon execution~\citep{mees2022calvin}, generalization~\citep{chen2025robotwin}, memory-dependent manipulation~\citep{chen2026rmbench}, precise manipulation~\citep{chen2026univtac,lan2025autobio}, and real-world deployment~\citep{yakefu2025robochallenge,chen2026manipulationnet,atreya2025roboarena}. Despite this progress, current evaluation protocols remain insufficient for systematically assessing generalist manipulation policies. Many benchmarks rely on relatively simple or short-horizon tasks, and task variations are often introduced primarily through changes in objects, layouts, or language expressions. While such variations are useful for measuring distributional robustness, they often preserve similar underlying manipulation patterns, limiting diagnostic coverage over distinct challenges such as generalization, memory, precision, open-semantic grounding, sequential execution, bimanual coordination, and tool use.

Another key limitation is the separation between simulation and real-world evaluation. Simulation-based benchmarks are efficient and scalable, making them suitable for rapid feedback and model iteration; however, they cannot fully capture contact-rich dynamics, actuation errors, perception noise, and other physical factors that emerge during deployment. Real-world evaluation provides a more direct assessment of policy behavior under physical-world conditions, but is costly, time-consuming, and difficult to reproduce without standardized hardware setups, scene reset procedures, evaluation protocols, and deployment interfaces. These limitations motivate a unified benchmark that combines scalable simulation-based diagnosis with reproducible real-world validation.

We introduce \textbf{\textcolor{Blue_1}{RoboDojo}}, a unified sim-and-real benchmark for efficient, comprehensive, and reproducible evaluation of generalist robot manipulation policies. RoboDojo contains 42 simulation tasks and 18 real-world tasks. The simulation benchmark provides broad capability coverage across Generalization, Memory, Long-Horizon, Precision, and Open, and supports heterogeneous parallel evaluation in Isaac Sim, enabling different tasks and scenes to run concurrently. The real-world benchmark complements simulation by exposing policies to challenging physical deployment conditions, including contact-rich interactions, perception noise, actuation errors, and environmental variations. To support reproducible physical testing, we design \textbf{\textcolor{Blue_1}{RoboDojo-RealEval}}, a remote real-world evaluation system that standardizes the hardware setup, workspace layout, lighting condition, scene reset procedure, evaluation protocol, and deployment interface. We further provide \textbf{\textcolor{Blue_1}{XPolicyLab}}, a unified infrastructure for policy development and deployment, enabling policies to be integrated once and evaluated across RoboDojo simulation and real-world settings with minimal policy-side adaptation.

RoboDojo is designed not only to expand benchmark scale, but also to establish a unified evaluation loop for diagnosing and improving generalist robot policies. In simulation, it supports fast policy iteration through diverse task designs and capability-oriented analysis. In the real world, it examines whether policy improvements transfer to challenging physical deployment conditions under standardized and reproducible settings. Together with continuously updated test cases, diagnostic analysis, and a public leaderboard, RoboDojo provides a systematic platform for measuring progress toward robust generalist manipulation. Our contributions are summarized as follows:

\begin{itemize}
\item We develop a unified sim-and-real evaluation system for robot manipulation policies, using a shared policy interface and evaluation pipeline across simulation and real-world testing. The system supports heterogeneous parallel simulation for efficient feedback and RoboDojo-RealEval for standardized, reproducible, and remote physical evaluation.

\item We construct the \textbf{RoboDojo Benchmark}, including 42 simulation tasks and 18 real-world tasks. The simulation tasks cover five capability dimensions: Generalization, Memory, Long-Horizon, Precision, and Open, while the real-world tasks span three robot embodiments and evaluate policies under challenging physical deployment conditions.

\item We build \textbf{XPolicyLab}, a standardized infrastructure for policy development and deployment. XPolicyLab integrates 30 robot manipulation models into a shared framework, enabling policies to be developed, deployed, and evaluated across RoboDojo benchmarks with minimal policy-side adaptation.

\item We conduct extensive evaluations of existing robot manipulation policies on RoboDojo. Based on these results, we establish a public leaderboard and provide systematic analysis of current policy limitations across simulation and real-world settings, highlighting key directions for future generalist manipulation policies.

\end{itemize}

%% file: sec/2_related_work.tex
\section{Related Work}
\label{sec:related_work}

\subsection{Robot Learning in Manipulation}

Robot manipulation policies have rapidly evolved from classical imitation learning methods to large-scale generalist robot policies~\citep{kim2024openvla,chen2025benchmarking}. Representative imitation learning approaches~\citep{zhao2023learning,chiDiffusionPolicyVisuomotor2024,chen2025g3flow,ze20243ddiffusionpolicygeneralizable,liu2025avr} have achieved strong visuomotor manipulation performance by learning from expert demonstrations. More recent vision-language-action models and foundation policies, including $\pi_{0.5}$~\citep{intelligence2025pi_}, RDT2~\citep{rdt2}, JoyAI-RA~\citep{zhang2026joyai}, Wall-OSS-0.5~\citep{yu2026wall}, Wall-WM~\citep{li2026wall}, LDA-1B~\citep{lyu2026lda}, Motus~\citep{bi2026motus}, G0.5~\citep{galaxea2026g05}, MotuBrain~\citep{team2026motubrain}, Qwen-RobotManip~\citep{yuan2026qwen}, Hy-Embodied-0.5-VLA~\citep{zhang2026hy} and LingBot-VA~\citep{li2026causalworldmodelingrobot}, further improve open-vocabulary instruction following, cross-task generalization, and cross-embodiment transfer. In parallel, recent studies have explored specific capability bottlenecks, such as memory-augmented manipulation~\citep{shi2025memoryvla,torne2026memmultiscaleembodiedmemory}, dynamic manipulation~\citep{cai2026internvla}, and world-model-based control~\citep{intelligence2026pi07steerablegeneralistrobotic,bi2026motus,ye2026worldactionmodelszeroshot}.

Despite this progress, existing policies still exhibit limited robustness when deployed beyond controlled task distributions. Their failures can arise from diverse factors, including insufficient temporal reasoning, weak spatial precision, limited generalization, and sensitivity to physical-world variations. This motivates evaluation protocols that go beyond aggregate task success and provide structured diagnosis across complementary manipulation capabilities and deployment settings. RoboDojo follows this direction by evaluating generalist policies through both capability-oriented simulation tasks and standardized real-world tests.

\subsection{Evaluation for Robot Manipulation}

A wide range of benchmarks have been proposed for evaluating robot manipulation. Simulation benchmarks provide scalable and reproducible testbeds: CALVIN~\citep{mees2022calvin} focuses on language-conditioned long-horizon manipulation; LIBERO~\citep{liu2023libero} studies lifelong learning across spatial, object, goal, and long-horizon suites; RoboTwin~\citep{chen2025robotwin} evaluates bimanual manipulation and generalization under structured domain randomization; RMBench~\citep{chen2026rmbench} targets memory-dependent manipulation; GarmentLab~\citep{lu2024garmentlab} focuses on garment manipulation; and SimplerEnv~\citep{li2024evaluatingrealworldrobotmanipulation} studies simulation environments that correlate with real-world performance. Real-world benchmarks, including ManipulationNet~\citep{chen2026manipulationnet}, RoboChallenge~\citep{yakefu2025robochallenge}, and RoboArena~\citep{atreya2025roboarena}, further assess policy behavior under physical deployment conditions.

These benchmarks have substantially advanced robot policy evaluation, but they are often specialized to particular task families, capability axes, or evaluation environments. Simulation-only benchmarks enable efficient and controlled stress testing, yet cannot fully capture the physical uncertainties of real-world deployment. Real-world benchmarks provide more direct evidence of deployed performance, but reproducibility remains difficult when hardware setups, scene reset procedures, evaluation protocols, and deployment interfaces vary across users or sites. RoboDojo addresses this gap by providing a unified sim-and-real benchmark with a shared policy interface and evaluation pipeline. It combines capability-oriented simulation diagnosis with reproducible real-world validation, while supporting policy integration through XPolicyLab. A detailed comparison between RoboDojo and existing benchmarks is provided in Appendix~\ref{sec:benchmark_comparison}.

%% file: sec/4_benchmark.tex
\section{RoboDojo Benchmark}
\label{sec:benchmark}

Building on the system design in Section~\ref{sec:method}, we construct the \textbf{\textcolor{Blue_1}{RoboDojo Benchmark}} to evaluate generalist robot manipulation policies across simulation and the real world. RoboDojo consists of 42 simulation tasks and 18 real-world tasks. The simulation benchmark supports efficient model iteration and capability-oriented diagnosis across five dimensions: Generalization, Memory, Precision, Long-Horizon, and Open. These dimensions capture distinct manipulation requirements beyond variations in objects, layouts, or language instructions. The real-world benchmark complements simulation by evaluating policies under challenging physical deployment conditions with standardized and reproducible protocols. Together, RoboDojo provides a broad evaluation suite for systematically diagnosing the limitations of current embodied manipulation policies.

\begin{figure}[!t]
\centering
\includegraphics[width=1.0\textwidth]{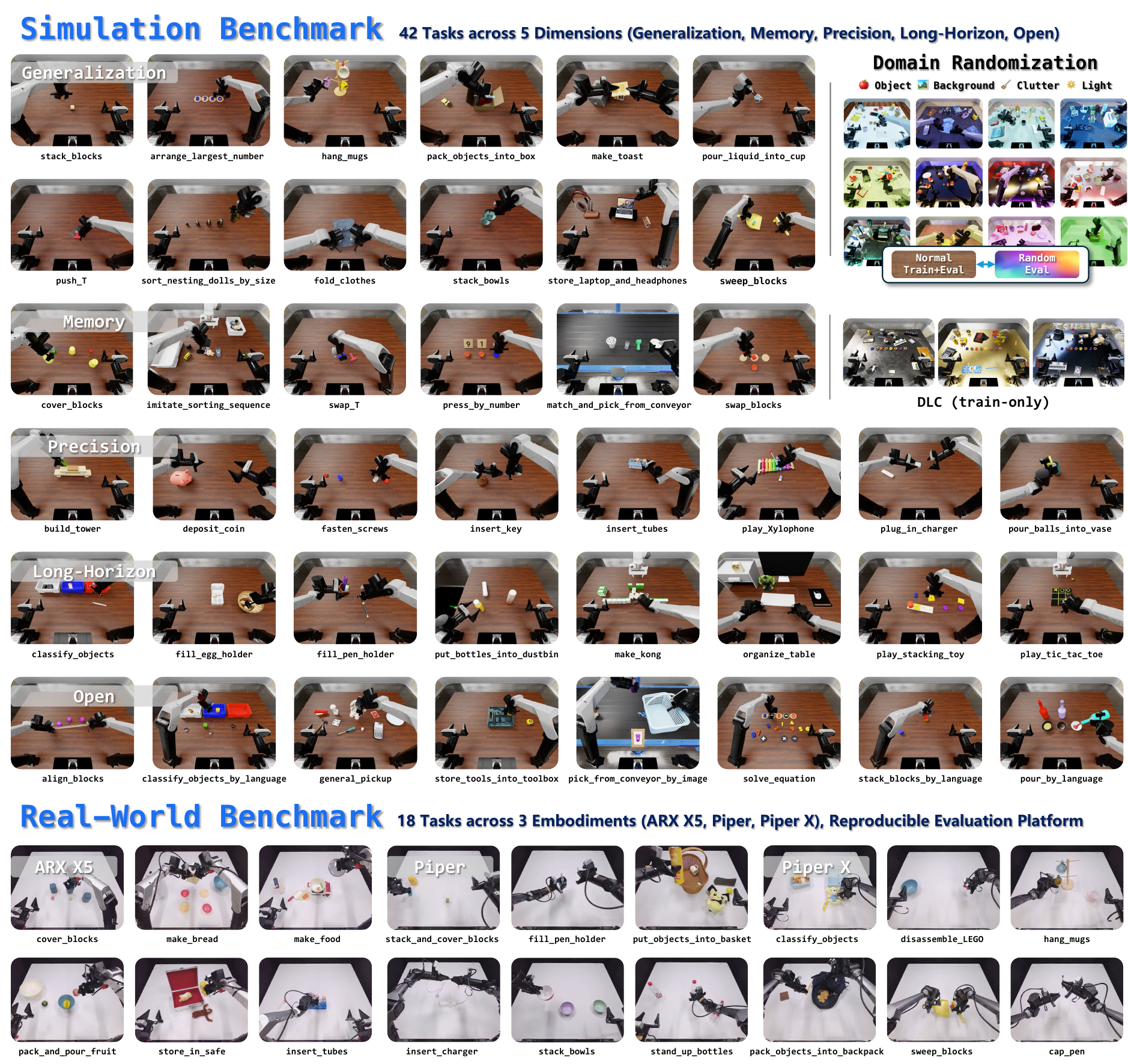}
\caption{\textbf{\textcolor{Blue_1}{Task Overview of RoboDojo.}}
RoboDojo includes 42 simulation tasks and 18 real-world tasks for evaluating generalist robot manipulation policies. The simulation tasks are organized into five capability dimensions: Generalization, Memory, Long-Horizon, Precision, and Open, enabling efficient capability-oriented diagnosis. The real-world tasks assess policy behavior under challenging and reproducible physical deployment conditions. Together, these tasks cover diverse manipulation skills, spatial configurations, and bimanual coordination patterns across simulation and the real world.}
\label{fig:all_tasks}
\end{figure}

\subsection{Simulation Benchmark}

We design 42 simulation tasks on the ARX X5 bimanual platform, with the two arm bases separated by 0.6,m. These tasks are organized into five capability dimensions: Generalization, Memory, Long-Horizon, Precision, and Open. The simulation benchmark is designed to provide rapid policy feedback and capability-oriented diagnosis beyond repeated variations of a narrow skill set. Together with the heterogeneous parallel simulation described in Section~\ref{sec:parallelism}, RoboDojo enables efficient model iteration across diverse task structures and manipulation requirements. An overview of all simulation tasks is shown in Fig.~\ref{fig:all_tasks}. Detailed task definitions and visualizations are provided in Appendix~\ref{appendix:sim_benchmark_tasks_details} and the \href{http://robodojo.com/doc}{project documentation}.

\subsubsection{Task Design}

\paragraph{Generalization.}
Robust deployment requires policies to complete the same task across diverse environments, rather than relying on fixed scenes or familiar visual contexts. The Generalization dimension includes 12 tasks that evaluate robustness to unseen variations in backgrounds, lighting, clutter, and target objects. Compared with RoboTwin 2.0~\citep{chen2025robotwin}, RoboDojo introduces stronger scene randomization and denser tabletop clutter. While RoboTwin 2.0 includes at most 10 clutter objects, RoboDojo randomizes up to 25 clutter objects, substantially increasing visual distraction and scene complexity. To further reduce overfitting to the default wooden-table scene, we provide 100 supplementary DLC trajectories for data-level augmentation. These trajectories are collected from simple pick-and-place behaviors under randomized backgrounds, lighting, and clutter layouts. Since they are independent of the evaluation tasks, they improve visual exposure without directly leaking task-specific solutions.

\paragraph{Memory.}
Many manipulation tasks are partially observable, where the correct action depends on information observed earlier rather than only the current frame. Inspired by RMBench~\citep{chen2026rmbench}, the Memory dimension includes 6 tasks that evaluate long-context observation modeling, key-frame memory, and non-Markovian decision making. These tasks cover both short-history settings, where a few key observations are sufficient for task completion, and longer interactive settings that require reasoning over multi-frame temporal sequences. For example, in \texttt{match\_and\_pick\_from\_conveyor}, the policy must remember the category of an object that disappears on the conveyor and later pick a matching object from subsequent candidates. In \texttt{imitate\_sorting\_sequence}, the policy observes another robot arm placing objects into a basket in a specific order, and must remember and reproduce the demonstrated ordering. These settings challenge policies that rely only on the current observation or a short fixed-length history.

\paragraph{Long-Horizon.}
Practical manipulation often requires executing a sequence of dependent steps before reaching the final goal. The Long-Horizon dimension includes 8 tasks that evaluate whether policies can infer task structure, maintain progress, and complete all required sub-steps without early termination or accumulated errors. For example, \texttt{classify\_objects} requires the robot to sort multiple objects into corresponding target locations. Due to the relatively large workspace, the task may also require handover behaviors between the two arms, resulting in longer execution horizons and increased temporal complexity.

\paragraph{Precision.}
Fine-grained manipulation requires accurate visual grounding, smooth motion prediction, and stable local control, since small spatial errors can directly lead to failure. The Precision dimension includes 8 tasks with strict spatial and motion accuracy requirements, focusing on fine-grained target localization, trajectory smoothness, and contact-rich control stability. For example, in \texttt{insert\_tubes}, the policy must continuously insert tubes into narrow holes. Even small localization or trajectory errors can prevent successful insertion, making this dimension particularly sensitive to action accuracy and motion stability.

\paragraph{Open.}
Generalist robot policies should understand diverse task specifications and transfer learned skills to new goals. The Open dimension includes 8 tasks that evaluate unseen task specifications whose required skills appear in the training data under different contexts. These tasks test open-semantic grounding, skill recombination, and language-conditioned transfer. For example, \texttt{general\_pickup} requires object grasping, a skill frequently observed during training, while the specific grasping goal is not directly included in the training tasks. This dimension examines whether policies can recombine learned manipulation skills and ground new language instructions to solve previously unseen tasks.

\subsubsection{Training Data Setting}

For simulation policy training, we provide 35 task directories with 3,500 trajectories, totaling 1,859,602 frames and 20.66 hours of bimanual manipulation data recorded at 25~Hz. These directories include 34 training tasks from the Generalization, Memory, Long-Horizon, and Precision dimensions, together with one auxiliary DLC data directory. Each training task contains 100 trajectories collected through either automated trajectory synthesis or VR-based teleoperation. The Open dimension is excluded from the training set, as it is designed to evaluate skill recombination and transfer to unseen task specifications rather than imitation of task-specific demonstrations.

For the Generalization dimension, demonstrations are collected under the standard setting without domain randomization. To reduce overfitting to a single visual configuration, we additionally include 100 auxiliary DLC trajectories for data-level augmentation. These trajectories are generated under complex domain randomization, including diverse backgrounds, lighting conditions, and clutter layouts. The auxiliary DLC data broadens visual exposure while remaining independent of the evaluation tasks, avoiding direct leakage of task-specific solutions. Detailed data modalities, collection procedures, and statistics are provided in Appendix~\ref{appendix:sim_training_data_statistics}.

\subsubsection{Evaluation Setting}

We evaluate policies on all 42 simulation tasks across five capability dimensions: Generalization, Memory, Long-Horizon, Precision, and Open. Each task is evaluated for 50 episodes, resulting in 2,100 episodes in total. We report both success rate and average score: success rate measures binary task completion, while average score captures partial task progress.

For the 12 Generalization tasks, the 50 episodes are split into 25 \textit{standard} episodes and 25 \textit{random} episodes. The \textit{standard} setting evaluates in-domain performance, whereas the \textit{random} setting introduces variations in backgrounds, clutter layouts, task-relevant objects, and lighting conditions to assess robustness under visual and scene-level distribution shifts. The overall performance is computed as the mean across the five capability dimensions, rather than across all tasks, preventing dimensions with more tasks from dominating the aggregate metric. Additional details on task horizons and episode settings are provided in Appendix~\ref{appendix:sim_evaluation_setting}.

\begin{figure}[t]
\centering
\includegraphics[width=1.0\textwidth]{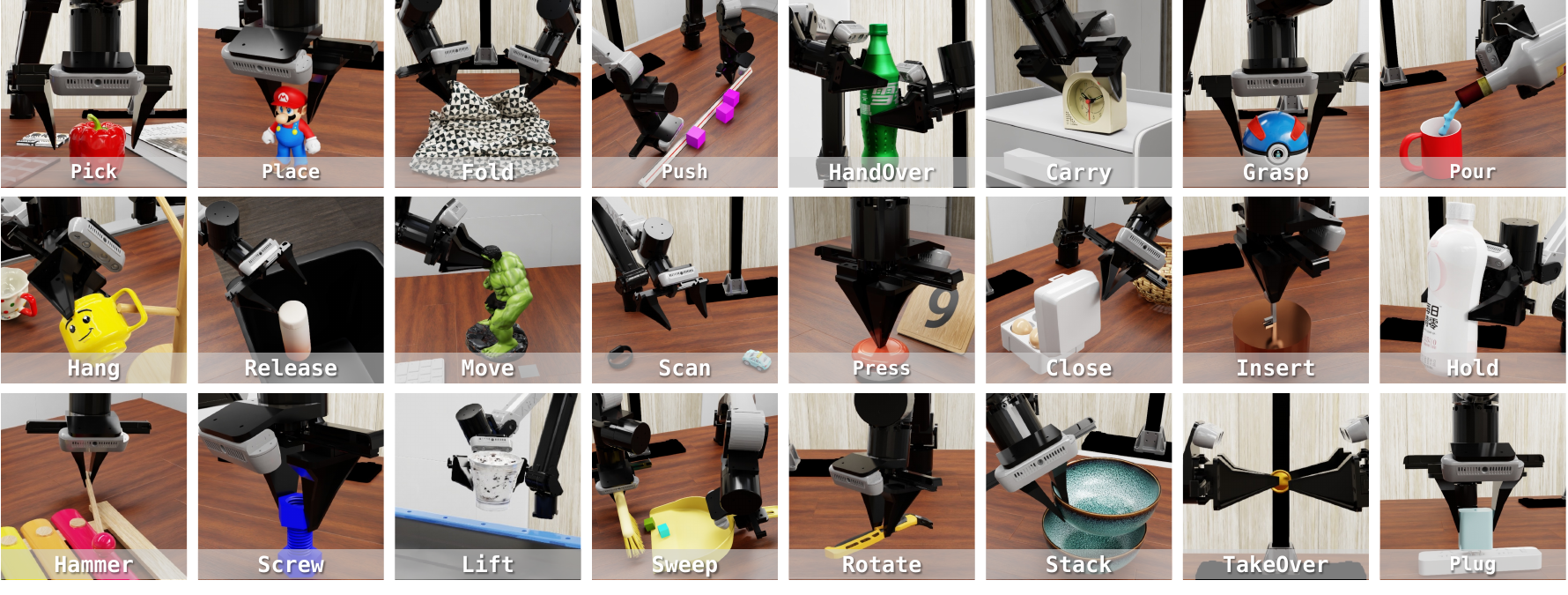}
\caption{\textbf{Skill Diversity in RoboDojo.}
Representative simulation tasks cover 24 manipulation skills, including grasping, placing, pushing, pulling, stacking, insertion, opening, closing, folding, alignment, tool use, and contact-sensitive operations. These tasks span diverse spatial configurations and bimanual coordination patterns, enabling skill-level diagnosis beyond repetitive pick-and-place behaviors.}
\label{fig:skill_diversity}
\end{figure}

\subsection{Real-World Benchmark}

To evaluate policy performance under physical deployment conditions, we construct the \textbf{\textcolor{Blue_1}{RoboDojo Real-World Benchmark}}. The benchmark consists of 18 real-world tasks across three commonly used collaborative bimanual robot platforms: ARX X5, Piper, and Piper X. This multi-embodiment design evaluates whether policies can maintain reliable performance across different robot kinematics, workspaces, camera placements, and data distributions, rather than being specialized to a single platform.

Unlike simulation tasks, which support rapid feedback and capability-oriented diagnosis, the real-world benchmark exposes policies to deployment challenges that are difficult to fully capture in simulation, including contact-rich interactions, perception noise, actuation errors, workspace constraints, and temporal variations. As described in Section~\ref{sec:real_world_hardware}, we open-source the hardware setup and evaluation specifications of \textbf{RoboDojo-RealEval} to support reproducible deployment. RoboDojo-RealEval standardizes the hardware configuration, workspace layout, lighting condition, scene reset procedure, evaluation protocol, and deployment interface, enabling policies to be tested under consistent physical conditions. We further provide cloud-based remote evaluation, allowing users to submit policies and evaluate them under shared physical test configurations. Detailed task information is provided in Appendix~\ref{appendix:real_benchmark_tasks_details} and the \href{https://robodojo-benchmark.com/doc}{project documentation}.

\begin{figure}[t]
\centering
\includegraphics[width=1.0\textwidth]{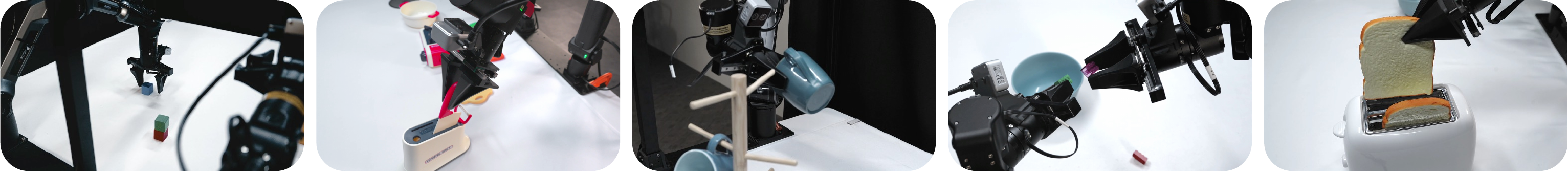}
\caption{\textbf{Real-World Task Highlights.} Representative key frames of challenging real-world manipulation tasks across different robot embodiments.}
\label{fig:real_highlight}
\end{figure}

\subsubsection{Task Design}

We design 6 tasks for each embodiment, resulting in 18 real-world tasks in total. Since real-world evaluation is constrained by hardware cost, execution time, human reset effort, and safety considerations, the benchmark is designed as a compact yet challenging physical test suite rather than an exhaustive enumeration of all capability dimensions. The selected tasks cover representative requirements, including long-horizon execution, precise manipulation, memory, and generalization.

Importantly, we do not construct one-to-one aligned task pairs between simulation and the real world. The real-world benchmark is not intended to measure direct sim-to-real transfer on matched tasks. Instead, it serves as a complementary evaluation setting for testing whether policies can handle physical deployment challenges beyond simulation success. The tasks involve contact-rich interactions, fine-grained spatial alignment, multi-object manipulation, partial observability, and embodiment-specific motion constraints, making evaluation sensitive to perception errors, actuation noise, accumulated trajectory drift, and unstable contact dynamics.

By evaluating policies across ARX X5, Piper, and Piper X under a shared protocol, RoboDojo assesses whether current generalist manipulation policies can operate reliably across diverse real-world deployment conditions. Detailed task specifications are provided in Appendix~\ref{appendix:real_benchmark_tasks_details}.

\subsubsection{Training Data Setting}

We collect real-world demonstrations using a homogeneous leader-follower teleoperation setup, where the leader arm has the same embodiment as the follower robot. This design aligns the teleoperation command space with the deployed robot embodiment, providing stable demonstrations for policy training. For each task, we collect 100 demonstrations from four operators to increase behavior diversity and reduce operator-specific bias. Across three embodiments and 18 tasks, the dataset contains 1,800 trajectories and 1,611,841 frames, corresponding to 17.91 hours of bimanual manipulation data recorded at 25~Hz.

Each demonstration includes robot states, language annotations, and synchronized RGB observations from one head camera and two wrist cameras. After collection, all trajectories are manually inspected for task completion, demonstration quality, and consistency with the RoboDojo-RealEval reset protocol. Detailed data modalities, filtering criteria, and embodiment-wise statistics are provided in Appendix~\ref{appendix:real_demo_statistics}.

\subsubsection{Evaluation Setting}

For real-world evaluation, we use the RoboDojo-RealEval platform introduced in Section~\ref{sec:real_world_hardware}. RoboDojo-RealEval supports both local deployment and remote cloud-based evaluation, where policies communicate with the real-robot client through the protocol described in Section~\ref{communication_protocol}. For each task, each policy is evaluated over 10 trials. Before each trial, a pre-collected evaluation layout is replayed to ensure consistent initial conditions across policies and sessions.

All trials are recorded and independently scored by three evaluators under a double-blind protocol. The scoring accounts for both final task success and intermediate sub-step completion, enabling fine-grained assessment of partial progress. The final score of each trial is computed as the average of the three evaluator scores. To improve transparency and fairness, we release evaluation videos and scores for leaderboard submissions, and provide an appeal mechanism for potential scoring errors. Additional details on motion planning, execution horizons, safety termination, and scoring protocols are provided in Appendix~\ref{appendix:real_evaluation_setting}.

\subsection{Evaluation Integrity and Anti-Gaming Protocols}

\begin{tcolorbox}[
colback=Blue_1!6,
colframe=Blue_1!65,
boxrule=0.6pt,
arc=2pt,
left=6pt,
right=6pt,
top=5pt,
bottom=5pt
]
\small
\textbf{\textcolor{Blue_1}{Leaderboard Governance.}}
The RoboDojo leaderboard is maintained by AI MMLab Club, a non-profit foundation dedicated to supporting open and community-driven AI research.
To ensure fairness, neutrality, and long-term credibility, RoboDojo evaluation will be jointly maintained and operated by global academic institutional partners.
No commercial company, including through sponsorship, funding, or operational participation, is involved in the official evaluation process.
As the benchmark evolves, the official evaluation and publication rules may be updated accordingly.
The latest leaderboard rules and submission protocol are available at
\href{http://RoboDojo-Benchmark.com/leaderboard}{http://RoboDojo-Benchmark.com/Leaderboard}.
\end{tcolorbox}

Although RoboDojo open-sources the simulation benchmark and the structural design of the real-world evaluation platform, official leaderboard publication requires a standardized verification protocol to ensure fairness, reproducibility, and community supervision. RoboDojo uses released public layouts as the primary leaderboard evaluation settings, allowing users to inspect task layouts, reproduce evaluations, and compare methods under a shared protocol. To reduce overfitting, hand-tuning, and leaderboard gaming on public layouts, submitted models are additionally evaluated on hidden verification layouts during official evaluation. These hidden layouts serve as an auxiliary consistency check rather than the primary leaderboard test set.

\begin{wrapfigure}{r}{0.35\linewidth}
    \vspace{-0.8em}
    \centering
    \includegraphics[width=0.95\linewidth]{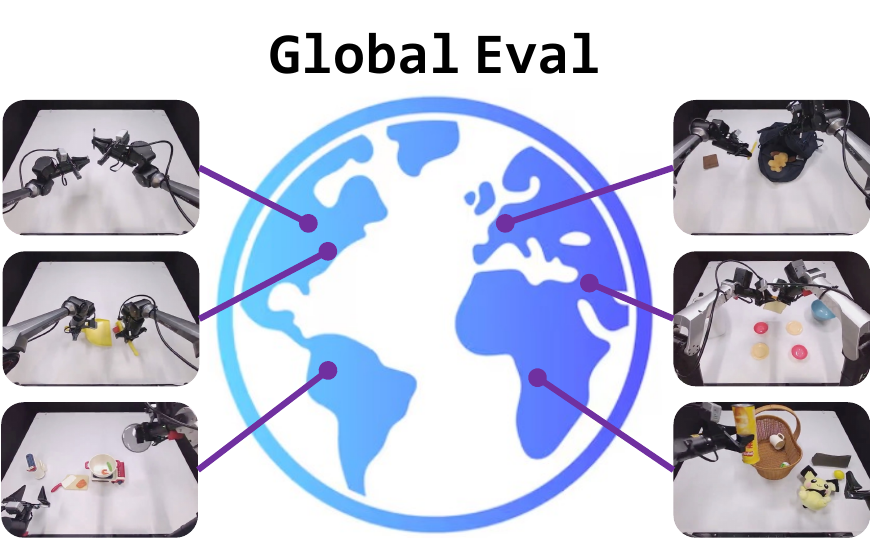}
    \caption{\small RoboDojo official evaluation is jointly supported by global academic institutional partners, with no commercial participation.}
    \label{fig:global_eval}
    \vspace{-1.0em}
\end{wrapfigure}

RoboDojo supports remote evaluation without requiring participants to release checkpoints or source code during private development. Participants can run a remote policy server and communicate with the official RoboDojo evaluation client through the standardized deployment protocol. To publish results on the official verified leaderboard, each submission must satisfy the following requirements:
\textbf{(1)} policies are evaluated through the official RoboDojo online evaluation system;
\textbf{(2)} simulation results are reported over three random seeds with mean and standard deviation, and real-world results cover all three robot embodiments, including ARX X5, Piper, and Piper X;
\textbf{(3)} submitted models pass hidden-layout verification;
\textbf{(4)} the evaluated checkpoint, training and deployment code, configuration files, and reproducibility instructions are released through \textbf{XPolicyLab} at the leaderboard publication stage; and
\textbf{(5)} evaluation videos are released for community inspection.

Participants may still use RoboDojo for private evaluation and iteration without releasing code or checkpoints. However, results without the evaluated checkpoint, implementation, configuration files, and reproducibility instructions are reported separately and are not considered verified leaderboard entries. In this way, RoboDojo balances flexible remote development with a verified publication protocol that emphasizes reproducible policy performance rather than hand-tuned results on specific released layouts. Additional details of the verification and publication protocol are provided in Appendix~\ref{appendix:evaluation_integrity}.

%% file: sec/3_method.tex
\section{Technical Implementation}
\label{sec:method}

In this section, we present the technical design of the \textbf{RoboDojo} evaluation platform. RoboDojo comprises three main components: a simulation evaluation platform, a real-world evaluation platform, and \textbf{XPolicyLab}, a unified infrastructure for policy development and deployment. The simulation platform supports diverse task construction, digital-twin asset generation, heterogeneous parallel evaluation, and simulation data collection, enabling rapid feedback and efficient policy iteration across manipulation capabilities. The real-world platform, \textbf{RoboDojo-RealEval}, enables reproducible physical evaluation by standardizing the hardware configuration, workspace layout, lighting condition, scene reset procedure, evaluation protocol, and deployment interface. XPolicyLab defines a shared policy interface for observation updates, action prediction, policy reset, and batched policy queries, allowing different robot manipulation policies to be integrated into a common codebase and evaluated under a unified protocol. Together, these components enable policies to be integrated once, iterated efficiently in simulation, and deployed to standardized real-world evaluation with minimal policy-side adaptation.

\input{sec/platform/simulation}

\input{sec/platform/real_world}

\input{sec/platform/xpolicylab}

%% file: sec/platform/simulation.tex
\subsection{Simulation Platform}
\label{sec:simulation_platform}

To support efficient and reproducible evaluation of generalist manipulation policies, RoboDojo builds a configurable simulation platform based on NVIDIA Isaac Sim~\citep{NVIDIA_Isaac_Sim} and Isaac Lab~\citep{mittal2025isaaclab}. The platform comprises three main components: a configuration-driven simulation stack, a physically grounded digital asset library, and a heterogeneous parallel execution mechanism. Together, these components support diverse task construction, scalable policy evaluation, and efficient demonstration collection. Additional implementation details are provided in Appendix~\ref{appendix:Simulation_Platform}.

\subsubsection{Simulation Platform Setup}

RoboDojo uses the vectorized execution interface of Isaac Lab~\citep{mittal2025isaaclab} as the simulation backend, and builds its simulation stack on the infrastructure of MagicSim~\citep{lu2026magicsim}, inheriting its modular manager architecture, configuration-driven scene construction pipeline, and object simulation backend. Each task is instantiated from modular YAML specifications that define task assets, scene layouts, initialization distributions, randomization ranges, and success conditions. This design decouples task specification from simulator execution, allowing different tasks to share the same runtime while varying task-specific objects, scene layouts, and reset distributions, details can be found in ~\ref{appendix:simulation_platform_details}.

At reset time, RoboDojo samples task-relevant objects, object poses, articulation states, clutter layouts, lighting conditions, and background textures with deterministic seed control. This enables diverse yet reproducible scene instances for both training and evaluation. The same scene construction pipeline supports rigid, articulated, and deformable assets through a unified interface. Details of the configuration format, randomization parameters, and scene construction pipeline are provided in Appendix~\ref{appendix:simulation_platform_setup}.

\subsubsection{Physically Grounded Digital Asset Library}

RoboDojo builds a digital asset library containing rigid, articulated, and deformable objects. These assets are used as both task-relevant objects and clutter distractors across simulation tasks, enabling diverse scene instantiations while maintaining reliable physical behavior for contact-rich manipulation, inspired by ManiTwin~\citep{wang2026manitwin}. Rigid objects are collected from online repositories and reconstructed from real-world daily objects using Meshy AI\footnote{\url{https://www.meshy.ai/}}. Each asset is annotated with semantic and task-level metadata, including object categories, language descriptions, placement regions, success-checking annotations, and manipulation affordances.

Articulated and deformable assets are collected from open-source platforms, 3D scanning, and selected assets from ClothesNet~\citep{zhou2023clothesnet}. Before being used in RoboDojo tasks, these assets are standardized in terms of kinematic structure, collision geometry, material parameters, and simulation-ready representations. Details on asset processing, annotation, and physical validation are provided in Appendix~\ref{appendix:asset_library_details}.

\begin{figure}[h]
    \centering
    \begin{subfigure}[t]{0.5\textwidth}
        \centering
        \includegraphics[width=\linewidth]{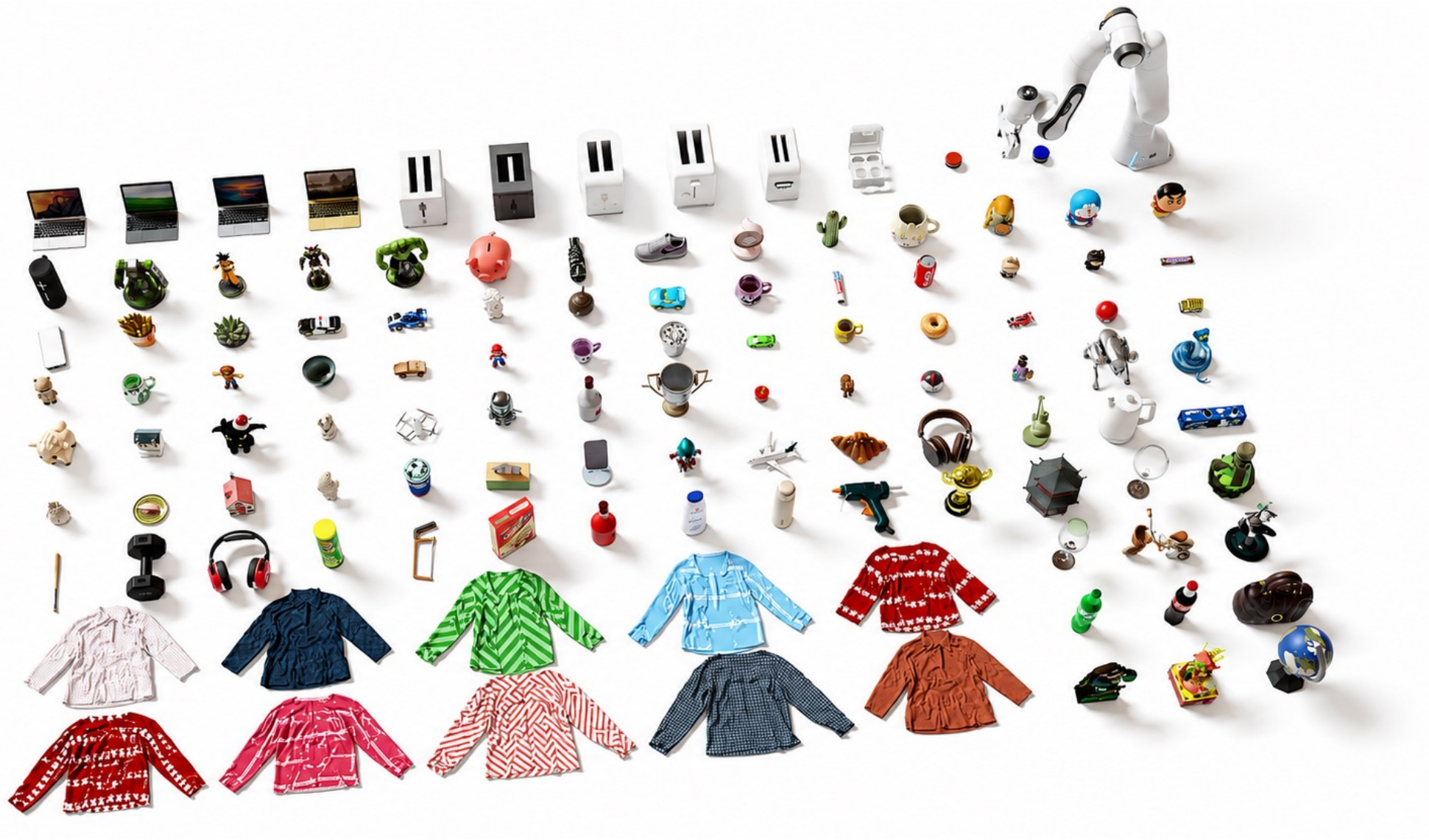}
        \caption{\textbf{Asset library.} Sample rigid, articulated, and deformable assets.}
        \label{fig:assets_library}
    \end{subfigure}
    \hfill
    \begin{subfigure}[t]{0.48\textwidth}
        \centering
        \roundedimage[6pt]{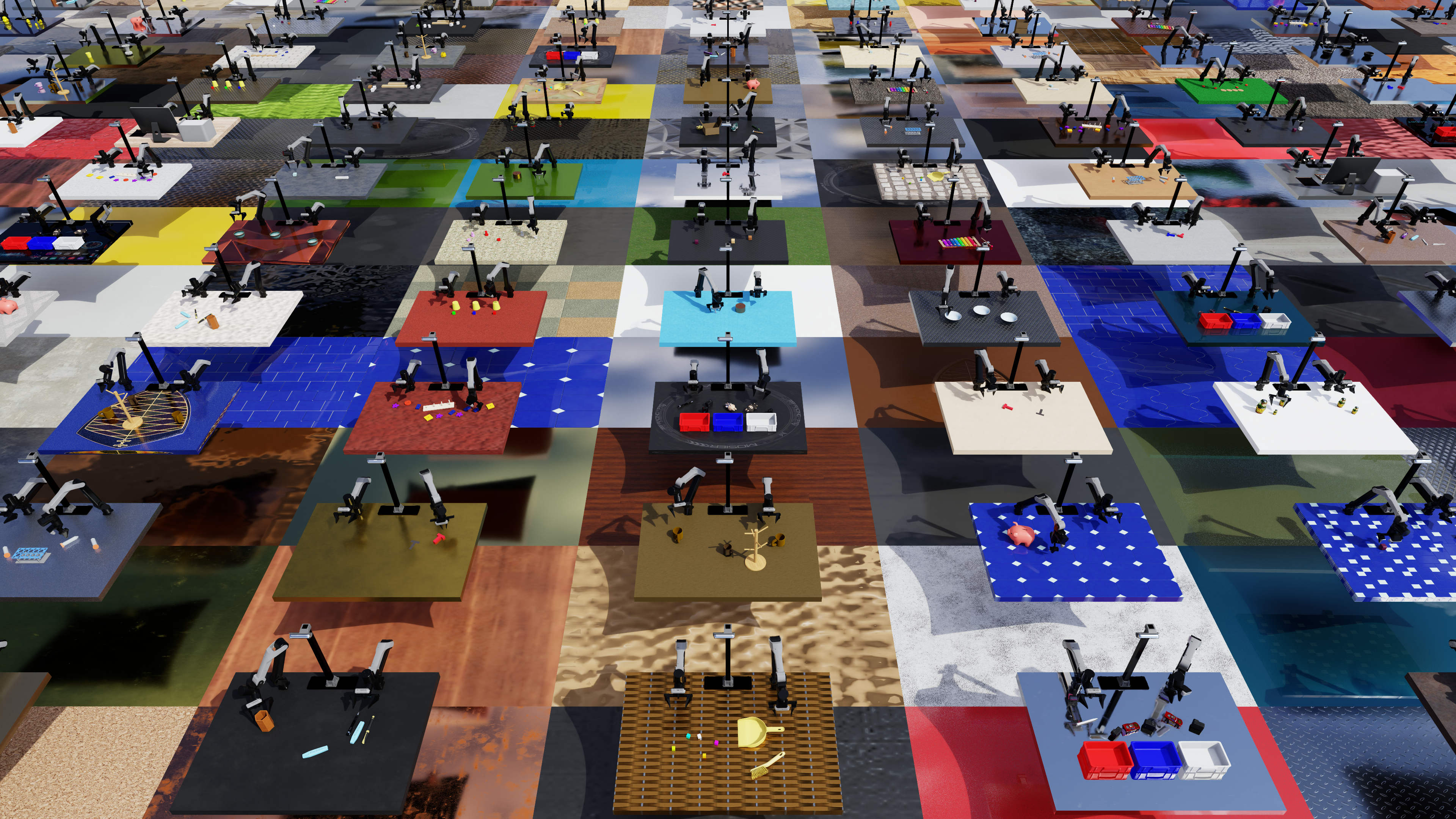}
        \caption{\textbf{Parallel simulation.} Concurrent heterogeneous task execution.}
        \label{fig:parallelism}
    \end{subfigure}
    \caption{\textbf{Assets and parallelism in RoboDojo.} RoboDojo combines physically grounded assets with heterogeneous parallel simulation for scalable benchmark construction and evaluation.}
    \label{fig:assets_parallelism}
\end{figure}

\subsubsection{Heterogeneous Parallelism}
\label{sec:parallelism}

Efficient benchmark evaluation requires parallel simulation over diverse scenes. Standard vectorized simulation often relies on homogeneous cloned environments, where parallel instances share the same scene template and differ mainly in poses or random seeds. While efficient, this setting is insufficient for evaluating generalist manipulation policies, which should be tested across diverse object geometries, object counts, clutter layouts, articulated structures, and background conditions.

RoboDojo therefore supports heterogeneous parallel simulation. Multiple environments are stepped under a shared vectorized interface, while each environment maintains an independently sampled scene configuration. Different parallel environments can contain different object categories, asset geometries, numbers of distractors, articulation structures, and task layouts. This design improves evaluation speed while preserving the scene-level diversity required for robust policy assessment. Implementation details of heterogeneous parallelism and multi-GPU sharding are provided in Appendix~\ref{appendix:heterogeneous_parallelism_details}.

\subsubsection{Simulation Data Collection}

RoboDojo supports two complementary data collection modes: automated trajectory synthesis and VR-based teleoperation. Both modes use a shared asset annotation layer that specifies manipulation-related affordances and task semantics, including graspable regions, placement regions, functional parts, and task-specific interaction points. These annotations support automated skill grounding and task validation.

For automated synthesis, RoboDojo composes demonstrations from reusable low-level skills, such as \texttt{grasp}, \texttt{place}, \texttt{handover}, \texttt{insert}, \texttt{open}, \texttt{close}, \texttt{stack}, and \texttt{push\_up}. Each skill is grounded in asset annotations and executed with the cuRobo v2 motion planner~\citep{sundaralingam2026curobov2dynamicsawaremotiongeneration}. For tasks that are difficult to synthesize automatically, RoboDojo provides a VR-based teleoperation interface for collecting high-quality demonstrations. Details on skill composition, motion planning, and teleoperation control are provided in Appendix~\ref{appendix:simulation_data_collection_details}.

%% file: sec/platform/real_world.tex
\subsection{Real-World Platform: RoboDojo-RealEval}
\label{sec:real_world_hardware}

\begin{figure}[h]
    \centering
    \includegraphics[width=1.0\textwidth]{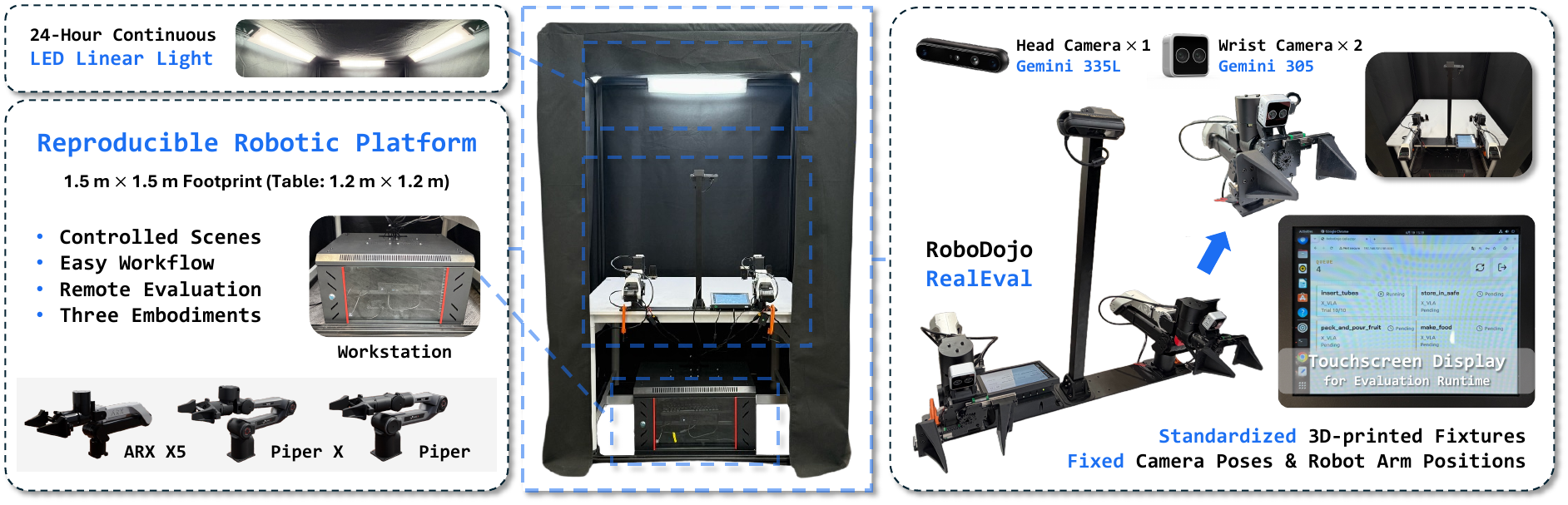}
    \caption{\textbf{Overview of the RoboDojo-RealEval system.}
    RoboDojo-RealEval provides a standardized physical platform for reproducible real-world robot manipulation evaluation, with controlled workspace geometry, fixed robot and camera mounts, stable lighting, a touchscreen evaluation interface, and support for three collaborative bimanual embodiments.}
    \label{fig:real_world_hardware}
\end{figure}

Real-world robot evaluation is highly sensitive to incidental factors such as lighting, robot placement, camera pose, manipulation surface condition, and scene reset accuracy, making reproducibility and fairness difficult to ensure across sessions and sites. To address this issue, we develop the \textbf{RoboDojo-RealEval Platform}, shown in Fig.~\ref{fig:real_world_hardware}, as a standardized hardware and software system for reproducible real-world evaluation. RoboDojo-RealEval fixes the relative poses of robot arms, cameras, lighting units, and the manipulation workspace through modular structural components, and provides a touchscreen-based web interface for scene reset, policy execution, emergency stop, video collection, and cloud-based scoring. During evaluation, predefined task layouts are replayed through a transparent overlay between the target layout image and the live observation stream, enabling evaluators to restore consistent initial conditions before each trial. The platform supports three collaborative bimanual embodiments, including ARX X5, Piper, and Piper X, and enables both local deployment and remote cloud-based evaluation under shared physical protocols. Together, these designs reduce uncontrolled variations and provide a reproducible physical foundation for standardized real-world policy evaluation. Additional hardware details, layout replay procedures, safety controls, and scoring protocols are provided in Appendix~\ref{appendix:realeval_details}.

%% file: sec/platform/xpolicylab.tex
\subsection{XPolicyLab}
\label{sec:xpolicylab}

XPolicyLab is a unified infrastructure and standard for robot policy development, training, evaluation, and deployment. It is designed to reduce the engineering burden of integrating heterogeneous robot policies into a shared benchmark ecosystem. Existing policies often rely on different data formats, preprocessing pipelines, training scripts, action representations, and runtime environments. XPolicyLab standardizes these external workflows through common data conversion tools, training templates, deployment procedures, and evaluation scripts, while preserving each policy's internal architecture and implementation.

Through XPolicyLab, we integrate 30 representative robot policy models into a shared codebase, enabling consistent training, debugging, deployment, and comparison under the RoboDojo protocol. For evaluation, XPolicyLab connects policy servers with RoboDojo simulation and RoboDojo-RealEval through a standardized observation-action interface. The same policy implementation can therefore be trained and iterated in simulation with fast feedback, and then deployed to remote real-world evaluation with minimal policy-side adaptation. The unified policy interface, communication protocol, and implementation details are provided in Appendix~\ref{appendix:xpolicylab_design}. The supported policies and project codebase are available at \href{https://XPolicyLab.github.io}{https://XPolicyLab.github.io}.

%% file: sec/5_XPolicyLab.tex

%% file: sec/5.5_leaderboard.tex
\section{RoboDojo Leaderboard}
\label{sec:leaderboard}

We establish RoboDojo leaderboards to provide standardized comparisons of current robot manipulation policies across both simulation and real-world settings. The simulation leaderboard evaluates broad capability coverage and supports efficient diagnosis through large-scale, repeated trials across five capability dimensions. The real-world leaderboard complements simulation by assessing policy behavior under standardized physical deployment conditions across multiple robot embodiments. Together, these leaderboards provide a unified view of policy performance, revealing both capability-level limitations in simulation and deployment-level challenges in the physical world.

\subsection{Simulation Benchmark Leaderboard}
\label{sec:sim_leaderboard}

Using \textbf{XPolicyLab}, we integrate, train, and evaluate 30 representative robot manipulation policies on the RoboDojo Simulation Benchmark Leaderboard. As a human-level reference, we additionally invite three expert teleoperators to perform VR-based simulation teleoperation under the same success criteria and execution horizons used for policy evaluation. Each expert has more than 1,000 hours of simulation teleoperation experience and participated in constructing the RoboDojo Simulation Benchmark dataset.

For most policies, we train three random seeds and evaluate each seed for 50 trials per task. For the Generalization dimension, the 50 trials are split into 25 trials under the standard setting and 25 trials under the randomized setting. Unless otherwise specified, each policy is therefore evaluated over 150 trials per task across three seeds. We report the mean success rate and mean score for each capability dimension, together with their corresponding standard deviations. The overall success rate and score are computed by averaging the corresponding metrics across the five capability dimensions.

The leaderboard is frozen for this paper on \textit{3 Jul. 2026}, while the latest results will be continuously updated at \href{http://RoboDojo-Benchmark.com/Leaderboard}{http://RoboDojo-Benchmark.com/Leaderboard}. Detailed training configurations for all evaluated models are provided in Appendix~\ref{appendix:train_details}. Models not trained with three random seeds are specially marked in the appendix.

\input{sec/benchmark_experiment/sim_bench_experiment}

\subsection{Real-World Benchmark Leaderboard}
\label{sec:real_leaderboard}

We evaluate 10 representative robot manipulation policies on the RoboDojo Real-World Benchmark Leaderboard. As a human-level reference, we additionally invite expert human teleoperators to perform real-world teleoperation using the same homogeneous leader-follower setup and task constraints as policy evaluation. Each teleoperator has more than 1,500 hours of real-world robot teleoperation experience and participated in collecting the RoboDojo Real-World Benchmark dataset.

For each evaluated policy, we train one random seed for each robot embodiment and evaluate it on the corresponding real-world tasks. The benchmark contains 18 tasks across three embodiments, with 10 trials per task, resulting in 180 real-world evaluation trials per policy. All evaluations are conducted through the RoboDojo-RealEval platform under the standardized scene reset procedure, evaluation protocol, and deployment interface described in Section~\ref{sec:real_world_hardware}. We report both task success rate and score, and compute aggregate performance across all real-world tasks.

The real-world leaderboard is frozen for this paper on \textit{3 Jul. 2026}, while the latest results will be continuously updated at \href{https://RoboDojo-Benchmark.com/Leaderboard}{https://RoboDojo-Benchmark.com/Leaderboard}. Detailed training configurations, hyperparameters, and policy-specific implementation details for all evaluated models are provided in Appendix~\ref{appendix:train_details}.

\input{sec/benchmark_experiment/real_bench_experiment}

%% file: sec/benchmark_experiment/sim_bench_experiment.tex
\input{table/sim_benchmark}

%% file: table/sim_benchmark.tex
\begin{table*}[!t]
\centering
\caption[\textbf{RoboDojo Simulation Benchmark Leaderboard}]{
\textbf{RoboDojo Simulation Benchmark Leaderboard}.
Each cell reports score / success rate for a capability dimension.
For each metric, rankings among evaluated policies are marked as \protect\best{Best} and \protect\secondbest{Second Best}.
Human teleoperation is reported as a reference and is excluded from policy ranking.
(3 Jul. 2026, continuously updated at 
\href{http://RoboDojo-Benchmark.com/Leaderboard}{RoboDojo-Benchmark.com/Leaderboard}.)}
\label{tab:robodojo_simulation_benchmark}
\begingroup
\small
\setlength{\tabcolsep}{5pt}
\renewcommand{\arraystretch}{1.08}

\newcommand{\thickline}{\noalign{\vspace{0.25em}\hrule height 1.2pt\vspace{0.25em}}}

\resizebox{\textwidth}{!}{
\begin{tabular}{l|c|c|c|c|c|c}
\thickline
\textcolor{Blue_1}{\textbf{Alg $\backslash$ Dim}} 
& \textcolor{Blue_1}{\textbf{Generalization}} 
& \textcolor{Blue_1}{\textbf{Precision}} 
& \textcolor{Blue_1}{\textbf{Long-Horizon}} 
& \textcolor{Blue_1}{\textbf{Memory}} 
& \textcolor{Blue_1}{\textbf{Open}} 
& \textcolor{Blue_1}{\textit{\textbf{Average}}}\\
\thickline

Hy-Embodied-0.5-VLA~\citep{zhang2026hy} 
& 11.77 / \secondbest{8.39\%} 
& 13.81 / 8.00\% 
& \best{25.74} / \best{14.92\%} 
& \best{13.37} / \best{12.11\%} 
& 0.65 / 0.58\% 
& \best{13.07} / \best{8.80\%} \\

Spatial Forcing~\citep{li2025spatial} 
& \best{14.12} / \best{9.33\%} 
& \secondbest{17.33} / \secondbest{10.58\%} 
& 23.26 / 14.58\% 
& 5.43 / 4.11\% 
& \secondbest{1.78} / \secondbest{1.58\%} 
& \secondbest{12.38} / \secondbest{8.04\%} \\

$\pi_{0.5}$~\citep{intelligence2025pi_} 
& \secondbest{13.37} / 8.17\% 
& 12.40 / 5.50\% 
& \secondbest{23.54} / \secondbest{14.67\%} 
& 5.78 / 4.56\% 
& \best{1.98} / \best{1.67\%} 
& 11.41 / 6.91\% \\

X-VLA~\citep{zheng2025x} 
& 10.48 / 6.78\% 
& \best{18.32} / \best{12.00\%} 
& 16.53 / 9.75\% 
& 4.76 / 3.56\% 
& 0.55 / 0.50\% 
& 10.13 / 6.52\% \\

X-WAM~\citep{guo2026unified4dworldaction} 
& 7.39 / 3.33\% 
& 6.72 / 1.83\% 
& 17.47 / 9.08\% 
& \secondbest{6.32} / 4.67\% 
& 0.57 / 0.25\% 
& 7.69 / 3.83\% \\

Xiaomi-Robotics-0~\citep{cai2026xiaomi} 
& 7.43 / 5.56\% 
& 8.42 / 4.58\% 
& 13.51 / 6.92\% 
& 5.07 / 3.67\% 
& 0.22 / 0.17\% 
& 6.93 / 4.18\% \\

StarVLA-$\alpha$~\citep{ye2026starvla} 
& 3.93 / 2.33\% 
& 9.90 / 4.33\% 
& 14.15 / 6.50\% 
& 3.34 / 2.44\% 
& 0.68 / 0.58\% 
& 6.40 / 3.24\% \\

GigaWorld-Policy~\citep{ye2026gigaworld} 
& 5.34 / 2.89\% 
& 6.15 / 1.83\% 
& 15.51 / 8.92\% 
& 3.46 / 2.22\% 
& 0.54 / 0.50\% 
& 6.20 / 3.27\% \\

GalaxeaVLA (G0)~\citep{jiang2025galaxea} 
& 4.53 / 2.83\% 
& 8.10 / 3.83\% 
& 12.60 / 5.58\% 
& 3.17 / 1.89\% 
& 0.70 / 0.67\% 
& 5.82 / 2.96\% \\

LingBot-VLA~\citep{wu2026pragmatic} 
& 6.71 / 4.28\% 
& 5.33 / 1.83\% 
& 10.89 / 5.25\% 
& 3.82 / 2.78\% 
& 0.72 / 0.67\% 
& 5.50 / 2.96\% \\

EventVLA~\citep{yang2026eventvla} 
& 3.94 / 1.94\% 
& 10.13 / 5.75\% 
& 5.05 / 0.83\% 
& 4.92 / \secondbest{4.78\%} 
& 0.80 / 0.75\% 
& 4.97 / 2.81\% \\

AHA-WAM~\citep{cai2026aha} 
& 5.79 / 3.28\% 
& 5.86 / 2.42\% 
& 8.61 / 2.67\% 
& 2.97 / 2.78\% 
& 0.88 / 0.83\% 
& 4.82 / 2.39\% \\

ABot-M0~\citep{yang2026abot} 
& 5.73 / 3.50\% 
& 5.50 / 1.75\% 
& 3.96 / 0.50\% 
& 2.44 / 2.22\% 
& 0.72 / 0.67\% 
& 3.67 / 1.73\% \\

Fast-WAM~\citep{yuan2026fast} 
& 2.34 / 1.11\% 
& 1.96 / 0.00\% 
& 9.14 / 5.17\% 
& 3.55 / 3.44\% 
& 0.42 / 0.42\% 
& 3.48 / 2.03\% \\

$\pi_0$~\citep{black2024pi_0} 
& 3.94 / 2.56\% 
& 3.56 / 0.75\% 
& 6.19 / 2.00\% 
& 3.47 / 2.11\% 
& 0.25 / 0.25\% 
& 3.48 / 1.53\% \\


GR00T-N1.7~\citep{bjorck2025gr00t} 
& 2.16 / 1.22\% 
& 2.54 / 0.67\% 
& 8.30 / 3.58\% 
& 1.06 / 0.89\% 
& 0.18 / 0.17\% 
& 2.85 / 1.31\% \\

InternVLA-A1~\citep{cai2026internvla} 
& 2.87 / 1.83\% 
& 3.00 / 0.92\% 
& 4.79 / 1.17\% 
& 1.58 / 1.33\% 
& 0.17 / 0.17\% 
& 2.48 / 1.08\% \\

SmolVLA (Single Task)~\citep{shukor2025smolvla} 
& 1.69 / 1.22\% 
& 2.87 / 0.33\% 
& 1.22 / 0.25\% 
& 3.35 / 2.44\% 
& 0.00 / 0.00\% 
& 1.83 / 0.85\% \\

LDA-1B~\citep{lyu2026lda} 
& 0.71 / 0.17\% 
& 3.21 / 0.50\% 
& 1.92 / 0.08\% 
& 2.08 / 1.78\% 
& 0.00 / 0.00\% 
& 1.58 / 0.51\% \\

MolmoAct2~\citep{fang2026molmoact2} 
& 0.39 / 0.06\% 
& 0.45 / 0.00\% 
& 2.32 / 0.00\% 
& 1.02 / 1.00\% 
& 0.91 / 0.83\% 
& 1.02 / 0.38\% \\

GO-1~\citep{bu2025agibot} 
& 1.58 / 1.22\% 
& 1.45 / 0.42\% 
& 1.13 / 0.25\% 
& 0.70 / 0.67\% 
& 0.08 / 0.08\% 
& 0.99 / 0.53\% \\

ACT (Single-Task)~\citep{zhao2023learning} 
& 0.69 / 0.56\% 
& 0.85 / 0.00\% 
& 1.73 / 0.92\% 
& 1.65 / 0.13\% 
& 0.00 / 0.00\% 
& 0.98 / 0.32\% \\

H-RDT~\citep{bi2026h} 
& 0.49 / 0.22\% 
& 0.41 / 0.00\% 
& 2.23 / 0.17\% 
& 0.12 / 0.11\% 
& 0.08 / 0.08\% 
& 0.67 / 0.12\% \\

RDT-1B~\citep{liu2025rdt} 
& 0.56 / 0.33\% 
& 0.38 / 0.00\% 
& 1.13 / 0.00\% 
& 0.49 / 0.33\% 
& 0.00 / 0.00\% 
& 0.51 / 0.13\% \\

DM0~\citep{yu2026dm0} 
& 0.49 / 0.06\% 
& 0.61 / 0.00\% 
& 0.97 / 0.08\% 
& 0.20 / 0.11\% 
& 0.00 / 0.00\% 
& 0.45 / 0.05\% \\

Dexora-1B~\citep{zhang2026dexora} 
& 0.49 / 0.11\% 
& 0.49 / 0.00\% 
& 0.82 / 0.00\% 
& 0.12 / 0.00\% 
& 0.01 / 0.00\% 
& 0.38 / 0.02\% \\

A1~\citep{zhang2026a1} 
& 0.16 / 0.00\% 
& 0.09 / 0.00\% 
& 1.07 / 0.00\% 
& 0.00 / 0.00\% 
& 0.08 / 0.08\% 
& 0.28 / 0.02\% \\

Spirit v1.5~\citep{spiritspirit} 
& 0.80 / 0.50\% 
& 0.03 / 0.00\% 
& 0.11 / 0.00\% 
& 0.22 / 0.22\% 
& 0.00 / 0.00\% 
& 0.23 / 0.14\% \\

TinyVLA~\citep{wen2025tinyvla} 
& 0.03 / 0.00\% 
& 0.05 / 0.00\% 
& 0.67 / 0.00\% 
& 0.11 / 0.11\% 
& 0.25 / 0.25\% 
& 0.22 / 0.07\% \\

OpenVLA-OFT~\citep{kim2025fine} 
& 0.04 / 0.00\% 
& 0.20 / 0.00\% 
& 0.70 / 0.00\% 
& 0.00 / 0.00\% 
& 0.08 / 0.08\% 
& 0.21 / 0.02\% \\


\thickline
Human Expert (Teleop) 
& 90.05 / 87.83\% 
& 68.06 / 64.00\% 
& 83.63 / 74.25\% 
& 75.25 / 74.33\% 
& 85.13 / 79.75\% 
& 80.42 / 76.03\% \\
\thickline

\end{tabular}
}
\endgroup
\end{table*}

%% file: sec/benchmark_experiment/real_bench_experiment.tex
\input{table/real_benchmark}

%% file: table/real_benchmark.tex
\begin{table*}[!t]
\newcommand{\thickline}{\noalign{\vspace{0.25em}\hrule height 1.2pt\vspace{0.25em}}}
\newcommand{\res}[2]{#1 / #2}
\newcommand{\na}{\textemdash}
\centering
\caption[\textbf{RoboDojo Real-World Benchmark Leaderboard}]{
\textbf{RoboDojo Real-World Benchmark Leaderboard}.
Each cell reports score / success rate.
For each embodiment-specific task, embodiment average, and overall average, rankings among evaluated policies are marked as \protect\best{Best} and \protect\secondbest{Second Best}; ties are marked consistently.
Human teleoperation is reported as a reference and is excluded from policy ranking.
(3 Jul. 2026, continuously updated at 
\href{http://RoboDojo-Benchmark.com/Leaderboard}{RoboDojo-Benchmark.com/Leaderboard}.)
}
\label{tab:robodojo_real_world_benchmark}
\begingroup
\scriptsize
\setlength{\tabcolsep}{2.5pt}
\renewcommand{\arraystretch}{1.12}

\resizebox{\textwidth}{!}{
\begin{tabular}{ll|cccccc|c|c}
\thickline
\textcolor{Blue_1}{\textbf{Policy}}
& \textcolor{Blue_1}{\textbf{Embodiment}}
& \textcolor{Blue_1}{\textbf{Task 1}}
& \textcolor{Blue_1}{\textbf{Task 2}}
& \textcolor{Blue_1}{\textbf{Task 3}}
& \textcolor{Blue_1}{\textbf{Task 4}}
& \textcolor{Blue_1}{\textbf{Task 5}}
& \textcolor{Blue_1}{\textbf{Task 6}}
& \textcolor{Blue_1}{\textbf{Emb. Avg.}}
& \textcolor{Blue_1}{\textbf{Overall Avg.}} \\
\midrule

\multirow{3}{*}{$\pi_{0.5}$~\citep{intelligence2025pi_}}
& ARX X5   
& \res{\best{24.6}}{\best{20.0}} 
& \res{\best{1.8}}{0.0} 
& \res{\best{25.8}}{\best{10.0}} 
& \res{\best{47.0}}{\best{20.0}} 
& \res{\secondbest{40.0}}{\best{20.0}} 
& \res{\best{26.8}}{\best{10.0}} 
& \res{\best{27.7}}{\best{13.3}} 
& \multirow{3}{*}{\res{\best{22.9}}{\best{12.8}}} \\
& Piper    
& \res{\best{10.0}}{\best{10.0}} 
& \res{\secondbest{28.0}}{0.0} 
& \res{\secondbest{10.0}}{\best{10.0}} 
& \res{0.0}{0.0} 
& \res{\secondbest{72.0}}{\secondbest{60.0}} 
& \res{\best{72.0}}{\best{50.0}} 
& \res{\best{32.0}}{\best{21.7}} 
& \\
& Piper X    
& \res{0.0}{0.0} 
& \res{\best{14.8}}{\best{10.0}} 
& \res{0.0}{0.0} 
& \res{\best{29.5}}{\best{10.0}} 
& \res{\best{7.5}}{0.0} 
& \res{3.0}{0.0} 
& \res{\best{9.1}}{\best{3.3}} 
& \\
\midrule

\multirow{3}{*}{InternVLA-A1~\citep{cai2026internvla}}
& ARX X5   
& \res{0.0}{0.0} 
& \res{0.0}{0.0} 
& \res{0.0}{0.0} 
& \res{0.0}{0.0} 
& \res{\best{48.0}}{\best{20.0}} 
& \res{12.0}{0.0} 
& \res{10.0}{3.3} 
& \multirow{3}{*}{\res{\secondbest{12.0}}{\secondbest{7.2}}} \\
& Piper    
& \res{0.0}{0.0} 
& \res{7.3}{0.0} 
& \res{0.0}{0.0} 
& \res{0.0}{0.0} 
& \res{\best{73.0}}{\best{70.0}} 
& \res{\secondbest{59.0}}{\secondbest{40.0}} 
& \res{\secondbest{23.2}}{\secondbest{18.3}} 
& \\
& Piper X    
& \res{0.0}{0.0} 
& \res{0.0}{0.0} 
& \res{\best{4.0}}{0.0} 
& \res{2.0}{0.0} 
& \res{0.0}{0.0} 
& \res{\best{10.0}}{0.0} 
& \res{\secondbest{2.7}}{0.0} 
& \\
\midrule

\multirow{3}{*}{GalaxeaVLA (G0)~\citep{jiang2025galaxea}}
& ARX X5   
& \res{1.0}{0.0} 
& \res{0.0}{0.0} 
& \res{3.0}{0.0} 
& \res{6.0}{0.0} 
& \res{0.0}{0.0} 
& \res{10.3}{0.0} 
& \res{3.4}{0.0} 
& \multirow{3}{*}{\res{9.0}{4.4}} \\
& Piper    
& \res{0.0}{0.0} 
& \res{\best{32.7}}{\best{10.0}} 
& \res{\best{13.3}}{\best{10.0}} 
& \res{0.0}{0.0} 
& \res{56.0}{50.0} 
& \res{30.0}{10.0} 
& \res{22.0}{13.3} 
& \\
& Piper X    
& \res{0.0}{0.0} 
& \res{0.0}{0.0} 
& \res{0.0}{0.0} 
& \res{\secondbest{4.0}}{0.0} 
& \res{0.0}{0.0} 
& \res{6.0}{0.0} 
& \res{1.7}{0.0} 
& \\
\midrule

\multirow{3}{*}{Xiaomi-Robotics-0~\citep{cai2026xiaomi}}
& ARX X5   
& \res{\secondbest{24.0}}{\best{20.0}} 
& \res{0.0}{0.0} 
& \res{3.0}{0.0} 
& \res{4.0}{0.0} 
& \res{\secondbest{40.0}}{\best{20.0}} 
& \res{\secondbest{19.0}}{\best{10.0}} 
& \res{\secondbest{15.0}}{\secondbest{8.3}} 
& \multirow{3}{*}{\res{7.9}{3.9}} \\
& Piper    
& \res{0.0}{0.0} 
& \res{0.0}{0.0} 
& \res{0.0}{0.0} 
& \res{0.0}{0.0} 
& \res{23.0}{20.0} 
& \res{22.7}{0.0} 
& \res{7.6}{3.3} 
& \\
& Piper X    
& \res{0.0}{0.0} 
& \res{0.7}{0.0} 
& \res{\best{4.0}}{0.0} 
& \res{0.0}{0.0} 
& \res{0.0}{0.0} 
& \res{2.7}{0.0} 
& \res{1.2}{0.0} 
& \\
\midrule

\multirow{3}{*}{X-VLA~\citep{zheng2025x}}
& ARX X5   
& \res{0.0}{0.0} 
& \res{0.0}{0.0} 
& \res{0.0}{0.0} 
& \res{2.0}{0.0} 
& \res{20.7}{\secondbest{10.0}} 
& \res{18.0}{0.0} 
& \res{6.8}{1.7} 
& \multirow{3}{*}{\res{7.6}{3.3}} \\
& Piper    
& \res{0.0}{0.0} 
& \res{5.3}{0.0} 
& \res{0.0}{0.0} 
& \res{0.0}{0.0} 
& \res{53.0}{50.0} 
& \res{37.0}{0.0} 
& \res{15.9}{8.3} 
& \\
& Piper X    
& \res{0.0}{0.0} 
& \res{0.0}{0.0} 
& \res{0.0}{0.0} 
& \res{0.0}{0.0} 
& \res{0.0}{0.0} 
& \res{0.7}{0.0} 
& \res{0.1}{0.0} 
& \\
\midrule

\multirow{3}{*}{GR00T-N1.7~\citep{bjorck2025gr00t}}
& ARX X5   
& \res{10.0}{\secondbest{10.0}} 
& \res{0.0}{0.0} 
& \res{\secondbest{9.0}}{0.0} 
& \res{14.7}{0.0} 
& \res{16.0}{0.0} 
& \res{6.0}{0.0} 
& \res{9.3}{1.7} 
& \multirow{3}{*}{\res{5.9}{1.7}} \\
& Piper    
& \res{0.0}{0.0} 
& \res{0.0}{0.0} 
& \res{0.0}{0.0} 
& \res{0.0}{0.0} 
& \res{10.0}{10.0} 
& \res{31.0}{10.0} 
& \res{6.8}{3.3} 
& \\
& Piper X    
& \res{0.0}{0.0} 
& \res{0.0}{0.0} 
& \res{0.0}{0.0} 
& \res{0.0}{0.0} 
& \res{0.0}{0.0} 
& \res{\secondbest{8.7}}{0.0} 
& \res{1.4}{0.0} 
& \\
\midrule

\multirow{3}{*}{$\pi_{0}$~\citep{black2024pi_0}}
& ARX X5   
& \res{9.0}{0.0} 
& \res{0.0}{0.0} 
& \res{3.0}{0.0} 
& \res{\secondbest{18.0}}{0.0} 
& \res{0.0}{0.0} 
& \res{18.7}{0.0} 
& \res{8.1}{0.0} 
& \multirow{3}{*}{\res{5.8}{1.7}} \\
& Piper    
& \res{0.0}{0.0} 
& \res{6.0}{0.0} 
& \res{0.0}{0.0} 
& \res{0.0}{0.0} 
& \res{42.0}{30.0} 
& \res{6.0}{0.0} 
& \res{9.0}{5.0} 
& \\
& Piper X    
& \res{0.0}{0.0} 
& \res{0.0}{0.0} 
& \res{0.0}{0.0} 
& \res{2.0}{0.0} 
& \res{0.0}{0.0} 
& \res{0.0}{0.0} 
& \res{0.3}{0.0} 
& \\
\midrule

\multirow{3}{*}{StarVLA-$\alpha$~\citep{ye2026starvla}}
& ARX X5   
& \res{0.0}{0.0} 
& \res{0.0}{0.0} 
& \res{0.0}{0.0} 
& \res{0.0}{0.0} 
& \res{0.0}{0.0} 
& \res{0.0}{0.0} 
& \res{0.0}{0.0} 
& \multirow{3}{*}{\res{4.1}{1.7}} \\
& Piper    
& \res{0.0}{0.0} 
& \res{14.0}{0.0} 
& \res{0.0}{0.0} 
& \res{0.0}{0.0} 
& \res{39.0}{30.0} 
& \res{18.7}{0.0} 
& \res{12.0}{5.0} 
& \\
& Piper X    
& \res{0.0}{0.0} 
& \res{\secondbest{2.7}}{0.0} 
& \res{0.0}{0.0} 
& \res{0.0}{0.0} 
& \res{0.0}{0.0} 
& \res{0.0}{0.0} 
& \res{0.5}{0.0} 
& \\
\midrule

\multirow{3}{*}{Spirit v1.5~\citep{spiritspirit}}
& ARX X5   
& \res{0.0}{0.0} 
& \res{0.0}{0.0} 
& \res{0.0}{0.0} 
& \res{0.0}{0.0} 
& \res{0.0}{0.0} 
& \res{0.0}{0.0} 
& \res{0.0}{0.0} 
& \multirow{3}{*}{\res{1.6}{0.6}} \\
& Piper    
& \res{0.0}{0.0} 
& \res{0.0}{0.0} 
& \res{0.0}{0.0} 
& \res{0.0}{0.0} 
& \res{10.0}{10.0} 
& \res{14.0}{0.0} 
& \res{4.0}{1.7} 
& \\
& Piper X    
& \res{0.0}{0.0} 
& \res{0.0}{0.0} 
& \res{0.0}{0.0} 
& \res{0.0}{0.0} 
& \res{0.0}{0.0} 
& \res{4.0}{0.0} 
& \res{0.7}{0.0} 
& \\
\midrule

\multirow{3}{*}{DM0~\citep{yu2026dm0}}
& ARX X5  
& \res{0.0}{0.0} 
& \res{0.0}{0.0} 
& \res{0.0}{0.0} 
& \res{0.0}{0.0} 
& \res{0.0}{0.0} 
& \res{0.0}{0.0} 
& \res{0.0}{0.0} 
& \multirow{3}{*}{\res{0.0}{0.0}} \\
& Piper   
& \res{0.0}{0.0} 
& \res{0.0}{0.0} 
& \res{0.0}{0.0} 
& \res{0.0}{0.0} 
& \res{0.0}{0.0} 
& \res{0.0}{0.0} 
& \res{0.0}{0.0} 
& \\
& Piper X 
& \res{0.0}{0.0} 
& \res{0.0}{0.0} 
& \res{0.0}{0.0} 
& \res{0.0}{0.0} 
& \res{0.0}{0.0} 
& \res{0.0}{0.0} 
& \res{0.0}{0.0} 
& \\
\thickline

\multirow{3}{*}{Human Expert (Teleop)}
& ARX X5  
& \res{100.0}{100.00} 
& \res{100.0}{100.00} 
& \res{100.0}{100.00} 
& \res{100.0}{100.00} 
& \res{100.0}{100.00} 
& \res{100.0}{100.00} 
& \res{100.0}{100.00} 
& \multirow{3}{*}{\res{100.0}{100.00}} \\
& Piper   
& \res{100.0}{100.00} 
& \res{100.0}{100.00} 
& \res{100.0}{100.00} 
& \res{100.0}{100.00} 
& \res{100.0}{100.00} 
& \res{100.0}{100.00} 
& \res{100.0}{100.00} 
& \\
& Piper X 
& \res{100.0}{100.00} 
& \res{100.0}{100.00} 
& \res{100.0}{100.00} 
& \res{100.0}{100.00} 
& \res{100.0}{100.00} 
& \res{100.0}{100.00} 
& \res{100.0}{100.00} 
& \\
\thickline
\end{tabular}
}
\begin{minipage}{0.98\textwidth}
\vspace{0.5em}
\footnotesize
\textit{Task order.} 
\textbf{ARX X5}: \texttt{cover\_blocks}, \texttt{make\_bread}, \texttt{make\_food}, \texttt{pack\_and\_pour\_fruit}, \texttt{store\_in\_safe}, \texttt{insert\_tubes}. 
\textbf{Piper}: \texttt{stack\_and\_cover\_blocks}, \texttt{fill\_pen\_holder}, \texttt{put\_objects\_into\_basket}, \texttt{insert\_charger}, \texttt{stack\_bowls}, \texttt{stand\_up\_bottles}. 
\textbf{Piper X}: \texttt{classify\_objects}, \texttt{disassemble\_LEGO}, \texttt{hang\_mugs}, \texttt{pack\_objects\_into\_backpack}, \texttt{sweep\_blocks}, \texttt{cap\_pen}.
\end{minipage}
\endgroup
\end{table*}

%% file: sec/6_experiment.tex
\section{Experiments}
\label{sec:experiments}

Beyond the leaderboard results in Section~\ref{sec:leaderboard}, our experiments further analyze what RoboDojo reveals about current robot manipulation policies and evaluate the benchmark system itself. We focus on three questions:
\textbf{\textcolor{Blue_1}{(1)}} what limitations RoboDojo reveals, by analyzing policy failures across capability dimensions and task types, and identifying key directions for future generalist manipulation policies;
\textbf{\textcolor{Blue_1}{(2)}} how efficiently RoboDojo supports benchmark execution, by examining evaluation efficiency in both heterogeneous parallel simulation and standardized real-world testing; and
\textbf{\textcolor{Blue_1}{(3)}} how reproducible RoboDojo evaluation is, by studying simulation-side stability across repeated seeds and real-world consistency under standardized RoboDojo-RealEval conditions.

\subsection{Analysis of RoboDojo Simulation Performance}

Based on the simulation leaderboard in Section~\ref{sec:sim_leaderboard}, we summarize key findings revealed by RoboDojo. Our analysis focuses on the current capabilities and limitations of generalist robot manipulation policies, including their performance gap from expert teleoperation and the remaining challenges toward robust general-purpose manipulation.

\input{sec/benchmark_experiment/sim_analysis}

\subsection{Analysis of RoboDojo Real-World Performance}

\input{sec/benchmark_experiment/real_analysis}

\subsection{Evaluation Efficiency}
\label{sec:evaluation_efficiency}

In this section, we evaluate the efficiency of RoboDojo in both simulation and real-world settings. For simulation, we measure the throughput improvement brought by heterogeneous parallel simulation under zero-action rollouts and large-policy inference. For real-world evaluation, we measure the wall-clock time required to complete physical trials, including scene reset, policy inference, and robot execution. These results quantify the practical cost of running RoboDojo at benchmark scale.

\subsubsection{Simulation Evaluation Efficiency}

\input{sec/efficiency_experiment/simulation_efficiency}

\subsubsection{Real-World Evaluation Efficiency}

\input{sec/efficiency_experiment/real_world_efficiency}

\subsection{Evaluation Stability}

We evaluate the stability of RoboDojo by measuring performance variation across both simulation and real-world evaluation settings. In simulation, we test whether the same policy obtains consistent results across different simulation workers and GPU deployments. In the real world, we examine cross-system consistency across RoboDojo-RealEval platforms, where hardware setup, lighting conditions, camera placement, and scene layout replay are standardized. This study assesses whether RoboDojo can provide stable and reliable scores for leaderboard comparison and remote real-world evaluation.

\subsubsection{Simulation Evaluation Stability}

\input{sec/consistency_experiment/simulation}

\subsubsection{Real-World Evaluation Stability}

\input{sec/consistency_experiment/real_world}

%% file: sec/benchmark_experiment/sim_analysis.tex
\textcolor{Blue_1}{\textbf{Finding 1: RoboDojo reveals a substantial gap toward balanced generalist robot manipulation.}}
Table~\ref{tab:robodojo_simulation_benchmark} summarizes the performance of current generalist robot manipulation policies on the RoboDojo simulation benchmark. Among the evaluated policies, Hy-Embodied-0.5-VLA, Spatial Forcing, $\pi_{0.5}$, X-VLA, Xiaomi-Robotics-0, X-WAM, GigaWorld-Policy, and StarVLA-$\alpha$ form the current leading group in terms of average performance. However, their results remain clustered in a low-score regime. Even the best-performing policy achieves only an $8.80\%$ average success rate and a $13.07$ average score, far below human experts, who reach a $76.03\%$ success rate and an $80.42$ score under the same evaluation protocol. This gap indicates that, although the tasks are feasible for human operators, current policies remain far from reliable task completion across diverse manipulation scenarios.

Beyond this overall gap, RoboDojo further reveals that current policies lack balanced capability development across dimensions. Different methods exhibit relative strengths in different areas: Spatial Forcing performs strongly on Generalization, X-VLA leads on Precision, $\pi_{0.5}$ is competitive on Open tasks, while Hy-Embodied-0.5-VLA achieves the best Long-Horizon, Memory, and overall performance. Nevertheless, these strengths remain dimension-specific and do not consistently translate across the full benchmark. Current policies perform relatively better on Generalization and Long-Horizon tasks, suggesting partial progress in visual grounding, goal localization, and multi-step execution. However, this advantage is only relative: even on these stronger dimensions, the best success rates remain below $15\%$. Precision, Memory, and Open tasks continue to expose more severe bottlenecks in fine-grained control, persistent scene understanding, and open-ended task execution. These results show that RoboDojo provides diagnostic signals beyond a single aggregate success rate, revealing fragmented progress across individual capabilities and a substantial gap toward balanced generalist robot manipulation.

\begin{figure}[t]
    \centering
    \includegraphics[width=1.0\textwidth]{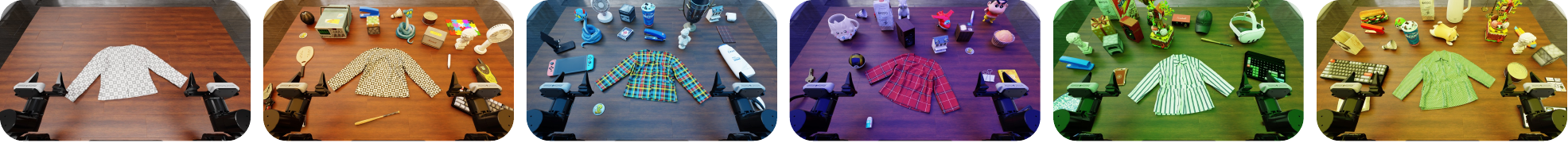}
    \caption{\textbf{Domain randomization in RoboDojo.}
    We visualize the effects of domain randomization in simulation, including variations in background, lighting, clutter layout, object appearance, and scene configuration. These randomized environments increase visual diversity and reduce overfitting to a fixed simulation setting.}
    \label{fig:domain_randomization}
\end{figure}

\textcolor{Blue_1}{\textbf{Finding 2: Scene-level randomization causes broad performance collapse, while visual-spatial grounding only partially improves robustness.}}

The Generalization dimension in RoboDojo evaluates policy robustness under visual and scene-level distribution shifts, including changes in object instances, layouts, clutter, lighting, and visual backgrounds. As shown in Table~\ref{tab:robodojo_simulation_benchmark}, Spatial Forcing achieves the strongest overall performance on this dimension, with a $9.33\%$ success rate and a $14.12$ score. The Standard--Random comparison in Table~\ref{tab:generalization_standard_random_score} further shows that the difficulty does not come from a few isolated hard cases; instead, scene randomization causes a broad performance collapse across nearly all policies.

A key observation is that strong performance in the Standard setting does not necessarily transfer to randomized scenes. Among the higher-performing methods, Hy-Embodied-0.5-VLA obtains the highest Standard score of $21.98$, but drops to $1.57$ under Random, corresponding to a $92.9\%$ relative drop. In contrast, Spatial Forcing improves over its $\pi_{0.5}$ base model in both absolute Random score ($6.98$ vs. $5.82$) and relative score retention, reducing the drop from $72.2\%$ to $67.2\%$. This suggests that explicit 3D spatial grounding and spatial representation alignment can improve robustness to scene-level variations, especially when object poses, spatial layouts, and visual backgrounds change.

However, this improvement remains limited in absolute terms. Even the strongest Random score is only $6.98$, and most competitive policies lose the majority of their Standard-setting performance after randomization. This indicates that current generalist manipulation policies remain highly sensitive to visual and spatial distribution shifts. Importantly, the failure is not merely a high-level recognition problem: even when policies can identify task-relevant objects or goals, distribution shifts may still degrade metric localization, action grounding, trajectory generation, and recovery from off-nominal states. Therefore, visual-spatial grounding is helpful for scene-level generalization, but reliable manipulation under randomized scenes also requires stable low-level control, robust closed-loop correction, and recovery-oriented policy behavior.

\input{table/generalization}

\textcolor{Blue_1}{\textbf{Finding 3: Current policies make partial progress on structured long-horizon tasks, but reliable skill composition remains difficult.}}
Long-Horizon is one of the relatively stronger dimensions in RoboDojo for leading policies. Hy-Embodied-0.5-VLA achieves the best performance, with a $14.92\%$ success rate and a $25.74$ score, followed closely by $\pi_{0.5}$ with a $14.67\%$ success rate and a $23.54$ score, and Spatial Forcing with a $14.58\%$ success rate and a $23.26$ score. Compared with Precision, Memory, and Open tasks, these results suggest that current policies can make more progress on structured multi-step workflows with clear intermediate goals and visually grounded substeps. Nevertheless, even the best policy succeeds in fewer than $15\%$ of Long-Horizon episodes, indicating that long-horizon manipulation remains far from reliable.

The gap between scores and success rates suggests that many policies can complete partial stages but fail to consistently reach the final goal. This reflects a central challenge in long-horizon manipulation: individual skills do not automatically compose into robust multi-step behavior. Policies must maintain task progress, select appropriate subtasks, switch stages at the right time, and recover from intermediate errors. Small execution errors can accumulate over extended sequences, causing later stages to fail even when early steps are partially successful.

The leading results further indicate that stronger action priors, embodied pretraining, and spatial grounding can help long-horizon execution. Hy-Embodied-0.5-VLA may benefit from large-scale embodied pretraining and temporally structured action prediction, while Spatial Forcing and $\pi_{0.5}$ suggest the value of spatial grounding and strong base policy training. However, these advantages remain limited. Overall, current policies show partial progress on structured long-horizon execution, but still lack reliable skill composition, stage-level decision making, and error recovery.

\textcolor{Blue_1}{\textbf{Finding 4: Precision remains a major bottleneck due to limited contact-aware control and weak off-trajectory correction.}}
The Precision dimension exposes a distinct failure mode from Generalization and Long-Horizon tasks. Policies must not only recognize task-relevant objects and target poses, but also generate spatially accurate, temporally smooth, and contact-sensitive actions under tight tolerances. As shown in Table~\ref{tab:robodojo_simulation_benchmark}, X-VLA achieves the best Precision performance, with a $12.00\%$ success rate and an $18.32$ score, followed by Spatial Forcing with a $10.58\%$ success rate and a $17.33$ score, and Hy-Embodied-0.5-VLA with an $8.00\%$ success rate and a $13.81$ score. However, even the strongest models remain far from reliable precision manipulation.

The Precision ranking differs from the overall leaderboard: Hy-Embodied-0.5-VLA obtains the best average performance, but is surpassed by X-VLA and Spatial Forcing on precision tasks. This suggests that stronger global task execution does not automatically yield accurate local control. Precision tasks require metric spatial reasoning, fine-grained motion generation, and stable contact transitions. Although stronger spatial grounding or action alignment may provide relative advantages, current policies still lack sufficiently robust 3D localization, smooth action prediction, and closed-loop contact control.

Our rollouts further suggest that many precision failures arise from weak state-conditioned correction rather than task misunderstanding. For example, in \texttt{insert\_key}, after failing to hand over or align the key, many policies still continue executing the subsequent insertion motion, indicating that they often follow an open-loop action sequence without verifying whether the current state satisfies the precondition for the next stage. We also observe action jitter in several policies across both end-effector and joint-space control interfaces, suggesting that such instability is not specific to a particular action representation, but is more likely related to insufficient temporal smoothness in action prediction. These results indicate that reliable precision manipulation requires smoother low-level action priors, contact-aware feedback, and correction-oriented training signals for recovering from off-trajectory states.

\textcolor{Blue_1}{\textbf{Finding 5: Memory mechanisms help, but memory-conditioned execution remains the bottleneck.}}
The Memory dimension shows that history- or memory-conditioned designs can improve performance, but are still far from reliable. Hy-Embodied-0.5-VLA achieves the best result, with a $12.11\%$ success rate and a $13.37$ score, while EventVLA also benefits from its KEM-based memory design, reaching a $4.78\%$ success rate and a $4.92$ score. However, the gap between these models suggests that explicit memory structure alone is insufficient. EventVLA stores sparse long-horizon evidence more explicitly, but our rollouts indicate that many failures arise from downstream manipulation errors or incorrect subtask execution rather than the complete absence of memory cues. In contrast, Hy-Embodied-0.5-VLA may benefit from stronger embodied pretraining and action priors, which help convert historical context into executable behavior. These results suggest that the key challenge is not only remembering past observations, but also using memory to guide state-conditioned decisions and robust action execution.

World-model-style prediction provides another form of implicit temporal memory. X-WAM achieves a $4.67\%$ success rate and a $6.32$ score on Memory tasks, slightly below EventVLA in success rate but higher in score. This suggests that predictive pretraining can help encode temporal continuity, object permanence, and task progress, leading to stronger partial completion. However, predictive models do not necessarily preserve or retrieve sparse historical evidence after it disappears from the current observation. Their limited final success rates indicate that temporal prediction alone is insufficient for tasks requiring reliable sparse-evidence recall and memory-conditioned control. Overall, RoboDojo suggests that future policies should combine predictive temporal modeling, explicit memory supervision, and action learning that directly conditions execution on remembered evidence.

\textcolor{Blue_1}{\textbf{Finding 6: Open-semantic manipulation remains largely unsolved.}}
The Open dimension is the most challenging split in RoboDojo. Current policies achieve near-zero performance on this dimension, and even the best-performing model, $\pi_{0.5}$, reaches only a $1.67\%$ success rate and a $1.98$ score. This result exposes a critical limitation of current generalist manipulation policies: they remain poorly equipped for tasks where language instructions require semantic interpretation beyond closed-set demonstrations. In real-world settings, users may specify novel goals, refer to objects by function or intent, or expect robots to recombine learned skills in unfamiliar ways. Unlike closed-set imitation tasks, Open tasks cannot be solved by simply matching familiar instructions to memorized action patterns.

These results suggest that current policies still lack robust semantic-to-action grounding. Although strong vision-language backbones can improve high-level perception and language understanding, they often fail to align open-ended instructions, visual affordances, and executable manipulation actions. In particular, policies must infer task intent, identify relevant objects or functional relations, select appropriate skills, and execute them under physical constraints. The near-zero Open performance indicates that this full semantic-to-action pipeline remains highly fragile. Future generalist manipulation policies therefore require stronger multimodal alignment, open-vocabulary affordance grounding, compositional skill learning, and execution-aware training signals that connect semantic goals to reliable robot behaviors.

%% file: table/generalization.tex
\begin{table}[t]
\centering
\small
\setlength{\tabcolsep}{3.5pt}
\renewcommand{\arraystretch}{1.25}
\caption{\textbf{Policy performance under standard and randomized visual settings.}
Each cell reports Standard Score / Random Score, with the relative score drop shown in parentheses. 
The drop is computed as $(\text{Standard}-\text{Random})/\text{Standard}$.}
\label{tab:generalization_standard_random_score}
\resizebox{\textwidth}{!}{
\begin{tabular}{cccccccc}
\toprule
\multicolumn{8}{c}{\textbf{Policy-wise Score} (Standard / Random, Relative Drop)} \\
\midrule

\makecell{\textbf{Hy-Embodied-0.5-VLA}\\ $21.98 / 1.57$ {\scriptsize $(92.9\%)$}}
&
\makecell{\textbf{Spatial Forcing}\\ $21.25 / 6.98$ {\scriptsize $(67.2\%)$}}
&
\makecell{\textbf{$\pi_{0.5}$}\\ $20.92 / 5.82$ {\scriptsize $(72.2\%)$}}
&
\makecell{\textbf{X\_VLA}\\ $17.92 / 3.04$ {\scriptsize $(83.0\%)$}}
&
\makecell{\textbf{Xiaomi-Robotics-0}\\ $13.81 / 1.05$ {\scriptsize $(92.4\%)$}}
&
\makecell{\textbf{X\_WAM}\\ $11.24 / 3.54$ {\scriptsize $(68.5\%)$}}
&
\makecell{\textbf{LingBot-VLA}\\ $10.87 / 2.55$ {\scriptsize $(76.5\%)$}}
&
\makecell{\textbf{AHA-WAM}\\ $10.32 / 1.26$ {\scriptsize $(87.8\%)$}}
\\

\addlinespace[3pt]

\makecell{\textbf{GigaWorldPolicy}\\ $10.28 / 0.41$ {\scriptsize $(96.0\%)$}}
&
\makecell{\textbf{ABot-M0}\\ $9.19 / 2.26$ {\scriptsize $(75.4\%)$}}
&
\makecell{\textbf{GalaxeaVLA}\\ $8.71 / 0.36$ {\scriptsize $(95.9\%)$}}
&
\makecell{\textbf{starVLA}\\ $7.53 / 0.33$ {\scriptsize $(95.6\%)$}}
&
\makecell{\textbf{$\pi_{0}$}\\ $7.18 / 0.71$ {\scriptsize $(90.1\%)$}}
&
\makecell{\textbf{EventVLA}\\ $6.66 / 1.22$ {\scriptsize $(81.7\%)$}}
&
\makecell{\textbf{InternVLA-A1}\\ $5.22 / 0.51$ {\scriptsize $(90.2\%)$}}
&
\makecell{\textbf{FastWAM}\\ $4.33 / 0.34$ {\scriptsize $(92.1\%)$}}
\\

\addlinespace[3pt]

\makecell{\textbf{GR00T-N1.7}\\ $3.97 / 0.35$ {\scriptsize $(91.2\%)$}}
&
\makecell{\textbf{SmolVLA}\\ $3.28 / 0.09$ {\scriptsize $(97.3\%)$}}
&
\makecell{\textbf{GO1}\\ $3.16 / 0.01$ {\scriptsize $(99.7\%)$}}
&
\makecell{\textbf{Spirit\_v15}\\ $1.60 / 0.00$ {\scriptsize $(100.0\%)$}}
&
\makecell{\textbf{ACT}\\ $1.37 / 0.00$ {\scriptsize $(100.0\%)$}}
&
\makecell{\textbf{LDA-1B}\\ $1.29 / 0.12$ {\scriptsize $(90.7\%)$}}
&
\makecell{\textbf{Dexbotic-DM0}\\ $0.93 / 0.04$ {\scriptsize $(95.7\%)$}}
&
\makecell{\textbf{H-RDT}\\ $0.89 / 0.09$ {\scriptsize $(89.9\%)$}}
\\

\addlinespace[3pt]

\makecell{\textbf{RDT-1B}\\ $0.82 / 0.29$ {\scriptsize $(64.6\%)$}}
&
\makecell{\textbf{Dexora-1B}\\ $0.80 / 0.18$ {\scriptsize $(77.5\%)$}}
&
\makecell{\textbf{MolmoACT2}\\ $0.73 / 0.04$ {\scriptsize $(94.5\%)$}}
&
\makecell{\textbf{A1}\\ $0.28 / 0.04$ {\scriptsize $(85.7\%)$}}
&
\makecell{\textbf{TinyVLA}\\ $0.06 / 0.01$ {\scriptsize $(83.3\%)$}}
&
\makecell{\textbf{OpenVLA-OFT}\\ $0.03 / 0.06$ {\scriptsize $(-100.0\%)$}}
&
&
\\

\bottomrule
\end{tabular}
}
\end{table}

%% file: sec/benchmark_experiment/real_analysis.tex
Before presenting the findings, we analyze the real-world results from three aspects: the gap to human teleoperation, the mismatch between simulation and real-world rankings, and deployment-specific failures such as action jitter and unsafe behaviors. These analyses show that RoboDojo-RealEval complements simulation by exposing physical-world bottlenecks that are difficult to capture in simulation alone.

\textcolor{Blue_1}{\textbf{Finding 1: Real-world deployment remains unreliable, with partial progress rarely translating into task completion.}}
Table~\ref{tab:robodojo_real_world_benchmark} shows that RoboDojo-RealEval remains highly challenging for current generalist manipulation policies. The best-performing policy, $\pi_{0.5}$, achieves only a $12.8\%$ overall success rate and a $22.9$ score across 18 real-world tasks, while human teleoperation reaches a $100.0\%$ success rate and a $100.0$ score on all tasks and embodiments. This gap shows that the tasks are feasible under the standardized protocol, yet current policies remain far from reliable physical deployment.

The higher scores relative to success rates further indicate that current policies can often make partial progress but fail to complete the full task. For example, $\pi_{0.5}$, InternVLA-A1, and GalaxeaVLA obtain scores of $22.9$, $12.0$, and $9.0$, respectively, but their success rates remain only $12.8\%$, $7.2\%$, and $4.4\%$. This suggests that real-world failures are driven not only by task misunderstanding, but also by execution bottlenecks such as precise alignment, contact handling, sub-step transition, and recovery from intermediate errors. Reliable real-world manipulation therefore requires closed-loop execution that can convert partial progress into final task success under physical uncertainty.

\textcolor{Blue_1}{\textbf{Finding 2: Simulation performance and real-world deployability are only partially aligned.}}
Since RoboDojo-RealEval currently evaluates a subset of the simulation leaderboard, we focus on policies evaluated in both settings. Among these overlapping policies, the two leaderboards show partial but not complete alignment. $\pi_{0.5}$ remains the strongest policy in RoboDojo-RealEval and is also among the leading methods in simulation, suggesting that simulation performance captures important aspects of general manipulation capability. However, the relative ordering changes for several policies: InternVLA-A1 and GalaxeaVLA achieve stronger real-world positions than their simulation rankings would suggest, while some policies with competitive simulation performance do not preserve the same advantage after physical deployment.

This mismatch should be interpreted carefully. RoboDojo is not designed as a paired sim-to-real transfer benchmark with one-to-one matched simulation and real-world task distributions. Instead, the two settings stress complementary aspects of generalist manipulation policies. Simulation provides scalable capability diagnosis across generalization, precision, long-horizon execution, memory, and open-semantic grounding, while RoboDojo-RealEval measures whether policies remain executable and reliable under physical deployment constraints.

Real-world evaluation introduces factors that are difficult to fully capture in simulation, including perception noise, camera and robot calibration errors, actuation latency, controller-specific dynamics, contact instability, and accumulated execution drift. These factors can cause policies with strong simulation performance to fail if their actions are jittery, poorly calibrated, insufficiently contact-aware, or unable to recover from small deviations. The partial alignment therefore supports the sim-and-real design of RoboDojo: simulation enables efficient large-scale capability analysis, while real-world evaluation exposes deployment-specific bottlenecks that cannot be inferred from simulation alone.

\textcolor{Blue_1}{\textbf{Finding 3: Real-world evaluation exposes execution instability and safety-critical behaviors beyond aggregate success metrics.}}
Real-world evaluation reveals deployment-specific failure modes that are not fully captured by aggregate success rates or simulation scores. Across several policies, we observe action jitter, oscillatory motions, repeated ineffective actions, contact instability, and occasional safety-critical behaviors. These failures indicate that physical deployment depends not only on high-level task understanding, but also on the temporal stability and safety of low-level action execution.

For example, DM0 frequently produces unstable control signals during deployment, leading to erratic motions that require close monitoring or safety intervention. This distinction is important: in simulation, unstable actions may only reduce task scores, whereas on physical robots they can introduce hardware risk, unsafe contacts, or unintended interactions with the environment. Therefore, final task success alone is insufficient to characterize real-world deployment quality.

These observations suggest that future generalist manipulation policies should optimize deployment-oriented properties beyond task completion, including action smoothness, contact-aware feedback, bounded control behavior, and recovery from off-trajectory states. RoboDojo-RealEval therefore provides complementary diagnostic value by revealing whether a policy is not only capable in principle, but also stable and safe enough for physical execution.

%% file: sec/efficiency_experiment/simulation_efficiency.tex
To evaluate the efficiency gain from heterogeneous parallel simulation, we conduct controlled experiments under five representative settings: RoboDojo with heterogeneous or non-heterogeneous parallel simulation under zero-action rollouts, RoboDojo with heterogeneous or non-heterogeneous parallel simulation coupled with $\pi_{0.5}$ policy inference, and RoboTwin 2.0 with zero-action rollouts.
All simulation and rendering workloads are executed on 8$\times$RTX 4090 GPUs.
In our setup, each RTX 4090 runs one RoboDojo simulation process with 10 parallel environments, while RoboTwin 2.0 runs two simulation processes per GPU.
Notably, RoboTwin 2.0 renders three camera views at a resolution of $320 \times 240$, whereas RoboDojo renders at $640 \times 480$, resulting in a higher per-frame rendering workload.
For non-zero-action RoboDojo settings, $\pi_{0.5}$ inference is performed on an A800 server within the same local network.
The zero-action setting measures raw simulation throughput without policy inference overhead, whereas the $\pi_{0.5}$ setting evaluates end-to-end efficiency under a realistic remote policy-deployment scenario.

As shown in Table~\ref{tab:simulation_efficiency}, heterogeneous parallel simulation substantially improves evaluation throughput. 
Under zero-action rollouts, RoboDojo achieves 77.4 interactions/s, compared with 40.0 interactions/s under non-heterogeneous parallel simulation, yielding a $1.94\times$ speedup. 
When $\pi_{0.5}$ inference is included, heterogeneous parallel simulation achieves 64.0 interactions/s, outperforming the non-heterogeneous setting by $1.63\times$. 
These results show that the proposed infrastructure improves both raw simulation throughput and practical end-to-end evaluation efficiency. 
Compared with RoboTwin 2.0 zero-action rollouts, RoboDojo achieves higher normalized throughput while supporting more diverse tasks and scenes. 
This efficiency enables large-scale policy evaluation and rapid feedback across comprehensive capability dimensions.

\begin{table}[h]
\centering
\caption{\textbf{Simulation evaluation efficiency.}
All simulation and rendering workloads are executed on 8$\times$RTX 4090 GPUs.
RoboDojo runs one simulation process with 10 parallel environments per RTX 4090, while RoboTwin 2.0 runs two simulation processes per GPU.
For non-zero-action RoboDojo settings, $\pi_{0.5}$ policy inference is performed on an A800 server within the same local network.}
\label{tab:simulation_efficiency}
\resizebox{0.9\textwidth}{!}{
\begin{tabular}{l p{0.45\textwidth} c c c}
\toprule
\textbf{Benchmark} & \textbf{Evaluation Setting} & \textbf{Frames} & \textbf{Time} & \textbf{Avg. Speed} \\
\midrule
RoboDojo 
& Heterogeneous parallel simulation, zero action 
& 1,640,000 & 5h 53m & 77.4 interactions/s \\

RoboDojo 
& Non-heterogeneous parallel simulation, zero action 
& 1,640,000 & 11h 22m & 40.0 interactions/s \\

RoboDojo 
& Heterogeneous parallel simulation + $\pi_{0.5}$ inference 
& 1,640,000 & 7h 07m & 64.0 interactions/s \\

RoboDojo 
& Non-heterogeneous parallel simulation + $\pi_{0.5}$ inference 
& 1,640,000 & 11h 38m & 39.2 interactions/s \\

RoboTwin 2.0 
& Zero-action rollout 
& 3,100,000 & 19h 19m & 44.6 interactions/s \\
\bottomrule
\end{tabular}
}
\end{table}

%% file: sec/efficiency_experiment/real_world_efficiency.tex
We further evaluate the efficiency of RoboDojo-RealEval using $\pi_{0.5}$ as the test policy. The policy server is deployed on an RTX 4090 server within the same local-area network as the real-world robot platform. For each task, we measure the wall-clock time required to complete 10 full evaluation trials, including scene reset, policy inference, and robot execution. All trials are executed to completion without manual interruption or early stopping, making the reported time a conservative estimate of the real-world evaluation cost.

As shown in Table~\ref{tab:real_task_eval_time}, RoboDojo-RealEval completes the 18-task real-world evaluation in 202.0 minutes, corresponding to approximately 3.4 hours for 180 physical trials. On average, each task requires 11.2 minutes for 10 trials. These results indicate that the standardized scene reset procedure, local RTX 4090 policy deployment, and unified evaluation interface substantially reduce the operational overhead of real-world benchmarking, enabling RoboDojo-RealEval to provide relatively rapid feedback while preserving full physical evaluation.

\begin{table}[h]
\centering
\caption{\textbf{Real-world evaluation time for each task.}
Each entry reports the wall-clock time required to complete 10 full evaluation trials, including scene reset, RTX 4090 policy inference within the same local-area network, and robot execution.}
\label{tab:real_task_eval_time}
\resizebox{0.87\linewidth}{!}{
\begin{tabular}{l c l c l c}
\toprule
Task & Time (min) & Task & Time (min) & Task & Time (min) \\
\midrule
connect\_charger & 13.5 & stand\_up\_bottles & 7.2 & cap\_pen & 6.1 \\
stack\_bowls & 7.3 & put\_objects\_into\_basket & 20.5 & pack\_and\_pour\_fruit & 11.1 \\
stack\_and\_cover\_blocks & 9.9 & hang\_mugs & 9.2 & store\_in\_safe & 9.8 \\
fill\_pen\_holder & 10.7 & pack\_objects\_into\_backpack & 12.1 & make\_food & 24.0 \\
sweep\_blocks & 6.0 & classify\_objects & 15.5 & insert\_tubes & 9.0 \\
disassemble\_LEGO & 5.7 & cover\_blocks & 10.8 & make\_bread & 13.6 \\
\midrule
\multicolumn{5}{r}{Total} & 202.0 \\
\bottomrule
\end{tabular}
}
\end{table}

%% file: sec/consistency_experiment/simulation.tex
Although RoboDojo uses standardized task configurations and evaluation protocols, Isaac Sim rendering and low-level simulation execution may still introduce minor nondeterminism across GPU devices. To evaluate the cross-GPU consistency of the simulation benchmark, we conduct stability experiments on three RTX 4090 GPUs. Specifically, we use layout 0 of each task and evaluate three representative policies, including $\pi_{0.5}$, Xiaomi-Robotics-0, and GalaxeaVLA, with three random seeds on each GPU.

\begin{wraptable}{r}{0.28\textwidth}
\vspace{-8pt}
\centering
\caption{\textbf{Real-world evaluation stability.}
We report the standard deviation across three repeated runs. Full per-task results are in Appendix Table~\ref{tab:real_world_consistency_std_full}.}
\label{tab:real_world_consistency_compact}
\vspace{-4pt}
\tiny
\setlength{\tabcolsep}{2pt}
\resizebox{\linewidth}{!}{
\begin{tabular}{lcccc}
\toprule
\multirow{2}{*}{Policy}
& \multicolumn{2}{c}{Overall}
& \multicolumn{2}{c}{Max Task} \\
\cmidrule(lr){2-3} \cmidrule(lr){4-5}
& SR & Score & SR & Score \\
\midrule
$\pi_{0.5}$  & 1.0 & 0.5 & 23.1 & 12.9 \\
GalaxeaVLA   & 1.2 & 1.0 & 15.3 & 16.6 \\
InternVLA-A1 & 1.3 & 1.2 & 17.3 & 10.4 \\
\bottomrule
\end{tabular}
}
\vspace{-10pt}
\end{wraptable}

Table~\ref{tab:sim_stability_std} reports the cross-GPU standard deviation of success rate and score. Overall, RoboDojo produces consistent simulation evaluation results across different GPUs. Across all policies and capability dimensions, the largest success-rate standard deviation is only $1.1$ percentage points, and the largest score standard deviation is $1.07$. For the overall average, the standard deviation is at most $0.5$ percentage points in success rate and $0.49$ in score. These results indicate that, under standardized evaluation layouts, RoboDojo yields stable simulation scores across GPU devices, making it suitable for fair leaderboard comparison despite minor nondeterminism from simulator rendering and low-level execution.

\begin{table}[h]
\centering
\caption{\textbf{Simulation evaluation stability across GPUs.}
We evaluate $\pi_{0.5}$, Xiaomi-Robotics-0, and GalaxeaVLA on three RTX 4090 GPUs using layout 0 of each task, with three random seeds on each GPU. Each cell reports the cross-GPU standard deviation in the format of success rate / score. Success-rate standard deviations are reported in percentage points.}
\label{tab:sim_stability_std}
\resizebox{0.75\linewidth}{!}{
\begin{tabular}{lcccccc}
\toprule
\textbf{Policy} & \textbf{Generalization} & \textbf{Precision} & \textbf{Long-Horizon} & \textbf{Memory} & \textbf{Open} & \textbf{Average} \\
\midrule
$\pi_{0.5}$ 
& $0.4$ / $0.46$ 
& $0.3$ / $0.15$ 
& $0.8$ / $0.54$ 
& $0.2$ / $0.18$ 
& $0.3$ / $0.31$ 
& $0.1$ / $0.19$ \\
\midrule
Xiaomi-Robotics-0 
& $0.3$ / $0.31$ 
& $1.1$ / $1.07$ 
& $0.7$ / $0.82$ 
& $0.5$ / $0.72$ 
& $0.2$ / $0.12$ 
& $0.5$ / $0.49$ \\
\midrule
GalaxeaVLA 
& $0.4$ / $0.45$ 
& $0.5$ / $1.06$ 
& $0.6$ / $0.68$ 
& $0.7$ / $0.54$ 
& $0.2$ / $0.13$ 
& $0.2$ / $0.37$ \\
\bottomrule
\end{tabular}
}
\end{table}

%% file: sec/consistency_experiment/real_world.tex
To examine the repeatability of RoboDojo-RealEval, we repeat real-world evaluation for three representative leaderboard policies: $\pi_{0.5}$, InternVLA-A1, and GalaxeaVLA. Each policy is evaluated for three independent rounds under the same RoboDojo-RealEval protocol, covering all 18 real-world tasks across three robot embodiments.

Table~\ref{tab:real_world_consistency_compact} summarizes the aggregate repeated-evaluation stability, and the full per-task breakdown is provided in Appendix Table~\ref{tab:real_world_consistency_std_full}. Overall, RoboDojo-RealEval produces stable aggregate evaluation results. Across the three policies, the standard deviation of the overall success rate is at most $1.3$ percentage points, and the standard deviation of the overall score is at most $1.2$ points. This indicates that the overall leaderboard results are not overly sensitive to a single evaluation round.

At the task level, several tasks exhibit larger variance, especially when the number of physical trials is limited and the task involves contact-rich or multi-stage execution. Such variance is expected in real-world robot evaluation, where small differences in object placement, contact state, perception noise, and accumulated execution error can affect the final outcome. Nevertheless, these variations are largely averaged out at the aggregate level, supporting the repeatability of RoboDojo-RealEval under its standardized hardware setup, scene reset procedure, evaluation protocol, and scoring process.

%% file: sec/6.5_future_direction.tex
\section{Future Extensions}

\begin{figure}[H]
    \centering
    \includegraphics[width=1.0\textwidth]{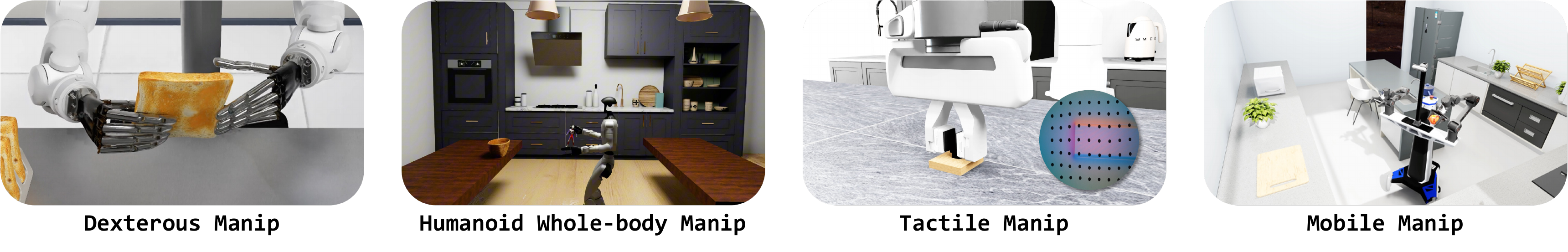}
    \caption{\textbf{Future extensions of RoboDojo.} RoboDojo will be continuously expanded to broader manipulation scenarios and embodiments, including dexterous hand manipulation, humanoid whole-body manipulation, tactile manipulation, and mobile manipulation.}

    \label{fig:future_work}
\end{figure}

RoboDojo is designed as an extensible benchmarking platform rather than a fixed task collection. In future releases, we will continuously expand RoboDojo toward broader scenarios, embodiments, and evaluation settings. As shown in Figure~\ref{fig:future_work}, we plan to introduce larger-scale benchmarks for dexterous hand manipulation, humanoid whole-body manipulation, tactile manipulation, and mobile manipulation. For each direction, RoboDojo will support multiple robot embodiments to improve accessibility and enable fair comparison across different hardware platforms. By progressively extending the task suite, embodiment coverage, and sensing modalities, RoboDojo aims to evolve into a general-purpose evaluation platform for embodied manipulation.

%% file: sec/7_conclusion.tex
\section{Conclusion}
\label{sec:conclusion}

We presented \textbf{RoboDojo}, a unified sim-and-real benchmark for evaluating generalist robot manipulation policies. RoboDojo includes 42 simulation tasks and 18 real-world tasks, covering diverse, long-horizon, precision-demanding, memory-dependent, and open-ended manipulation scenarios. The simulation benchmark supports efficient large-scale evaluation through heterogeneous parallel execution, enabling rapid feedback and fine-grained capability diagnosis. The real-world benchmark, built on RoboDojo-RealEval, provides standardized hardware configurations, scene reset procedures, evaluation protocols, and remote cloud-based access for reproducible physical assessment. Together with \textbf{XPolicyLab}, a unified standard and infrastructure for policy development and deployment, RoboDojo enables policies to be integrated, evaluated, and compared under a shared protocol.

Our experiments show that current robot policies remain far from reliable general-purpose manipulation. In simulation, RoboDojo reveals insufficient capability coverage across key dimensions, including generalization, long-horizon execution, precise manipulation, memory, and open-ended task understanding, indicating that current models are not yet comprehensive across diverse manipulation requirements. In the real world, the evaluated policies still perform poorly on complex manipulation tasks, showing that current models are not yet robust enough to handle challenging physical deployment conditions. RoboDojo will continue to update its leaderboard with new policies, tasks, evaluation results, and community submissions, providing a challenging testbed, reproducible evaluation platform, and open benchmark for tracking progress toward more capable omni-manipulation policies.

%% file: sec/X_appendix.tex
\newpage

\appendix
\crefalias{section}{appendix}
\crefname{appendix}{Appendix}{Appendices}
\Crefname{appendix}{Appendix}{Appendices}

\addcontentsline{toc}{section}{Appendix}
\addtocontents{toc}{\protect\setcounter{tocdepth}{-1}}

\begin{center}
    {\Large\bfseries Appendix}
\end{center}

\section{Acknowledgments}
\label{sec:Acknowledgments}
\subsection{Contributors}

\paragraph{RoboDojo-RealEval Infrastructure Contributors.}
We sincerely thank Tian Nian, Zijian Cai, Kehe Ye, Yukun Liao, Shaolong Zhu, Qiangyu Chen, Jiahao Zhang and Zichun Chen for their valuable support in developing the RoboDojo-RealEval real-world evaluation infrastructure.

\paragraph{Policy Contributors.}
We sincerely thank the following contributors for providing policy models and supporting their integration and evaluation in RoboDojo: Jun Guo, Zongzheng Zhang, Hongzhe Bi, Jisong Cai, Xiaofeng Wang, Zheng Zhu, Yuhang Tang, Weijie Ke, Mingleyang Li, Ganlin Yang, Shuai Yang, Hengtao Li, Wenxuan Song, Zhangzheng Tu, Kaidong Zhang, Yu Sun, Shuhe Huang, Junliang Guo, Tong Zhang, Yixing Chen, Pengxiang Ding, Rongxu Cui, and Hengkai Tan.

\subsection{Evaluation Integrity and Leaderboard Publication Details}
\label{appendix:evaluation_integrity}

\paragraph{Official remote evaluation.}
Policies must be evaluated through the RoboDojo online evaluation system, either by submitting a deployable policy package or by connecting a remote policy server. Both simulation and real-world evaluations are conducted through the official RoboDojo pipeline using the released task layouts, standardized scene reset procedure, evaluation protocol, and deployment interface. Reported scores are computed by the official evaluation system rather than self-reported by participants.

\paragraph{Repeated simulation evaluation and multi-embodiment real-world evaluation.}
For simulation, each submitted policy is evaluated under three random seeds to reduce the influence of stochastic initialization and evaluation noise. Participants may submit either three training-seed checkpoints or one checkpoint evaluated under three evaluation seeds. RoboDojo reports the mean and standard deviation across the three simulation runs. For real-world evaluation, each policy is evaluated on all three robot embodiments, including ARX X5, Piper, and Piper X. We report both per-embodiment and aggregate performance to reflect whether a policy generalizes across different collaborative bimanual platforms.

\paragraph{Hidden verification.}
During official evaluation, each submitted model is also evaluated on a hidden verification set with randomized layouts. This set is used to detect overfitting, hand-tuning, or gaming of the released public layouts. If the success rate on the hidden verification set differs significantly from that on the public evaluation set, the submission is considered invalid and is excluded from the official verified leaderboard.

\paragraph{Open-source artifact for verified publication.}
Participants who choose to publish results on the official verified leaderboard must release the full evaluation artifact through \textbf{XPolicyLab}. This includes the training and deployment code, the evaluated checkpoint, configuration files, and instructions for model loading, inference, deployment, and evaluation under the unified RoboDojo interface. This requirement applies only at the leaderboard publication stage, not during private evaluation. Releasing the exact artifact used for evaluation enables community inspection and independent reproduction.

\section{Comparison with Existing Benchmarks} \label{sec:benchmark_comparison}

We compare RoboDojo with existing robot manipulation benchmarks in Table~\ref{tab:robodojo_simulation_benchmark_compare}. The comparison highlights RoboDojo's distinctive design in terms of benchmark scale, sim-and-real evaluation, capability coverage, task diversity, real-world reproducibility, and remote evaluation support. Compared with prior benchmarks that are often specialized to either simulation or real-world evaluation, RoboDojo provides a unified benchmark suite that connects rapid simulation-based diagnosis with standardized real-world validation.

\input{table/compare}

\section{Real-World Evaluation Stability}

We evaluate $\pi_{0.5}$, GalaxeaVLA, and InternVLA-A1 over three independent RoboDojo-RealEval runs. Each cell reports the standard deviation of success rate and score across R1--R3, with success-rate deviations in percentage points. The ``Overall'' row gives the standard deviation of the average performance across runs.

\begin{table*}[h]
\centering
\caption{\textbf{Full per-task real-world evaluation stability across repeated runs.}}
\label{tab:real_world_consistency_std_full}
\scriptsize
\setlength{\tabcolsep}{4pt}
\resizebox{0.8\textwidth}{!}{
\begin{tabular}{lcccccc}
\toprule
\multirow{2}{*}{Task}
& \multicolumn{2}{c}{$\pi_{0.5}$}
& \multicolumn{2}{c}{GalaxeaVLA}
& \multicolumn{2}{c}{InternVLA-A1} \\
\cmidrule(lr){2-3} \cmidrule(lr){4-5} \cmidrule(lr){6-7}
& Std. SR (pp) & Std. Score
& Std. SR (pp) & Std. Score
& Std. SR (pp) & Std. Score \\
\midrule
\texttt{connect\_charger} & 0.0 & 0.0 & 0.0 & 0.0 & 0.0 & 0.0 \\
\texttt{stack\_bowls} & 5.8 & 4.0 & 15.3 & 16.6 & 5.8 & 3.5 \\
\texttt{stack\_and\_cover\_blocks} & 5.8 & 5.8 & 0.0 & 0.0 & 0.0 & 0.0 \\
\texttt{fill\_pen\_holder} & 10.0 & 1.2 & 5.8 & 11.7 & 0.0 & 3.5 \\
\texttt{stand\_up\_bottles} & 5.8 & 6.6 & 0.0 & 7.8 & 15.3 & 9.8 \\
\texttt{put\_objects\_into\_basket} & 5.8 & 5.8 & 5.8 & 5.8 & 5.8 & 5.8 \\
\texttt{hang\_mugs} & 0.0 & 1.2 & 0.0 & 2.3 & 0.0 & 2.3 \\
\texttt{pack\_objects\_into\_backpack} & 5.8 & 3.5 & 0.0 & 1.7 & 0.0 & 6.5 \\
\texttt{sweep\_blocks} & 5.8 & 5.8 & 5.8 & 5.8 & 0.0 & 0.0 \\
\texttt{disassemble\_LEGO} & 5.8 & 4.6 & 0.0 & 1.2 & 0.0 & 0.0 \\
\texttt{classify\_objects} & 0.0 & 2.3 & 0.0 & 0.0 & 0.0 & 0.0 \\
\texttt{cap\_pen} & 0.0 & 2.0 & 0.0 & 2.3 & 0.0 & 2.0 \\
\texttt{pack\_and\_pour\_fruit} & 5.8 & 3.1 & 0.0 & 3.5 & 0.0 & 10.0 \\
\texttt{store\_in\_safe} & 23.1 & 12.9 & 11.5 & 13.9 & 17.3 & 10.4 \\
\texttt{make\_food} & 5.8 & 5.1 & 0.0 & 1.7 & 0.0 & 0.0 \\
\texttt{insert\_tubes} & 5.8 & 6.4 & 0.0 & 8.4 & 0.0 & 4.6 \\
\texttt{cover\_blocks} & 5.8 & 4.6 & 0.0 & 0.3 & 0.0 & 0.0 \\
\texttt{make\_bread} & 0.0 & 0.0 & 0.0 & 0.0 & 0.0 & 0.0 \\
\midrule
\textbf{Overall} & \textbf{1.0} & \textbf{0.5} & \textbf{1.2} & \textbf{1.0} & \textbf{1.3} & \textbf{1.2} \\
\bottomrule
\end{tabular}
}
\end{table*}

\section{Simulation Training and Evaluation Details}

\subsection{Simulation Training Data Details}
\label{appendix:sim_training_data_statistics}

Table~\ref{tab:sim_training_data_statistics} summarizes the simulation training data used in RoboDojo. The training set contains 35 task directories and 3,500 trajectories, totaling 1,859,602 frames, corresponding to 20.66 hours of bimanual manipulation data recorded at 25~Hz. Each trajectory provides synchronized RGB-D observations from a head-mounted camera and two wrist cameras at a resolution of $640 \times 480$. Depth images are normalized and stored as integer values in millimeters. A third-view RGB video is also recorded for preview and visualization. The dataset further includes robot states and next-frame actions, covering end-effector poses, gripper states, and joint positions.

Demonstrations are collected through either automated trajectory synthesis or VR-based teleoperation, depending on task structure and execution difficulty. For tasks in the Generalization, Memory, Long-Horizon, and Precision dimensions, we collect 100 trajectories per task. The Open dimension is reserved for evaluation only and does not include task-specific training demonstrations, since it is designed to assess skill recombination and transfer to unseen task specifications rather than imitation of task-specific demonstrations.

For Generalization tasks, training trajectories are collected under the normal setting without domain randomization. To reduce overfitting to a single visual configuration during supervised fine-tuning, we additionally include one auxiliary DLC task with 100 domain-randomized trajectories. These trajectories include diverse backgrounds, lighting conditions, and clutter layouts, broadening visual exposure without leaking task-specific solutions from the evaluation tasks.

\begin{table}[h]
\centering
\caption{\textbf{Simulation training data statistics.} The simulation training set contains 1,859,602 frames from 3,500 trajectories, corresponding to 20.66 hours of bimanual manipulation data at 25\,Hz. The Open dimension is excluded from the training set and used only to evaluate skill recombination and transfer to unseen task specifications. DLC denotes an auxiliary domain-randomized task used to broaden visual exposure during training.}
\label{tab:sim_training_data_statistics}
\small
\setlength{\tabcolsep}{4pt}
\resizebox{0.8\textwidth}{!}{%
\begin{tabular}{lccccc}
\toprule
\textbf{Category} & \textbf{Frames} & \textbf{Duration} & \textbf{Trajectories} & \textbf{Avg. Frames / Traj.} & \textbf{Avg. Duration / Traj.} \\
\midrule
Generalization & 592,432 & 6.58 h & 1,200 & 494 & 19.75 s \\
Memory & 328,975 & 3.66 h & 600 & 548 & 21.93 s \\
Precision & 368,459 & 4.09 h & 800 & 461 & 18.42 s \\
Long-Horizon & 504,133 & 5.60 h & 800 & 630 & 25.21 s \\
Open & 0 & 0.00 h & 0 & -- & -- \\
DLC & 65,603 & 0.73 h & 100 & 656 & 26.24 s \\
\midrule
Total & 1,859,602 & 20.66 h & 3,500 & 531 & 21.25 s \\
\bottomrule
\end{tabular}%
}
\end{table}

\subsection{Simulation Evaluation Details}
\label{appendix:sim_evaluation_setting}

For each simulation task, the maximum interaction horizon is determined from the demonstration data. Specifically, we compute the 90th percentile of task-specific trajectory lengths and set the evaluation horizon to $1.2\times$ this value. For short tasks generated by automated trajectory synthesis, we use a larger multiplier of $1.5\times$ to avoid premature termination caused by small trajectory-length variations. This setting provides sufficient execution time while keeping the evaluation horizon consistent with the demonstrated task complexity.

\section{Real-World Training and Evaluation Details}

\subsection{Real-World Training Data Details}
\label{appendix:real_demo_statistics}

Table~\ref{tab:real_demo_statistics} summarizes the real-world demonstration data used in RoboDojo. We collect demonstrations using a homogeneous leader-follower teleoperation setup, where the leader arm has the same embodiment as the follower robot. For each task, we collect 100 demonstrations recorded by four different operators to improve behavior diversity and reduce operator-specific bias. Across three robot embodiments, including ARX X5, Piper, and Piper X, the dataset contains 1,800 trajectories and 1,611,841 frames, corresponding to 17.91 hours of bimanual manipulation data recorded at 25~Hz. Although each embodiment contributes the same number of trajectories, the total duration differs due to embodiment-specific execution speeds, workspace configurations, and task structures.

\begin{table}[t]
\centering
\caption{\textbf{Real-world demonstration statistics.} The real-world dataset contains 1,611,841 frames from 1,800 trajectories across three robot embodiments, corresponding to 17.91 hours of bimanual manipulation data at 25~Hz. Each embodiment contains 600 trajectories from 6 tasks, while the duration differs due to embodiment-specific execution speeds and task structures.}
\label{tab:real_demo_statistics}
\small
\setlength{\tabcolsep}{4pt}
\resizebox{0.8\textwidth}{!}{
\begin{tabular}{lccccc}
\toprule
\textbf{Embodiment} & \textbf{Frames} & \textbf{Duration} & \textbf{Trajectories} & \textbf{Avg. Frames / Traj.} & \textbf{Avg. Duration / Traj.} \\
\midrule
ARX X5 & 665,071 & 7.39 h & 600 & 1,108 & 44.34 s \\
Piper & 539,737 & 6.00 h & 600 & 900 & 35.98 s \\
Piper X & 407,033 & 4.52 h & 600 & 678 & 27.14 s \\
\midrule
Total & 1,611,841 & 17.91 h & 1,800 & 895 & 35.82 s \\
\bottomrule
\end{tabular}
}
\end{table}

Each demonstration includes robot joint states, end-effector poses, language annotations, and synchronized RGB observations from three camera views, including one head camera and two wrist cameras. All videos are recorded at a resolution of $640 \times 480$. After collection, all trajectories are manually inspected by multiple reviewers. The filtering criteria include whether the RoboDojo-RealEval platform is properly reset before the trial, whether the operator hesitates excessively during execution, and whether the task is successfully completed. The reset check covers protocol-specific conditions such as flattening the table cover and adjusting lighting to the standard intensity. This filtering process ensures that the demonstrations are task-complete, high-quality, and consistent with the real-world evaluation protocol.

\subsection{Real-World Evaluation Details}
\label{appendix:real_evaluation_setting}

For each real-world task, each policy is evaluated for 10 trials. Evaluation layouts are collected in advance and replayed before each trial to ensure consistent initial conditions across policies and sessions. When a policy outputs end-effector control commands, we use Pink~\citep{pink} as the robot motion planner to convert end-effector targets into executable robot motions.

The maximum execution horizon for each task is determined from the demonstration videos by multiplying the 90th percentile trajectory length by 1.5. Once the step limit is reached, the trial is automatically terminated and the robot returns to its reset pose. The evaluation manager may also manually stop a trial if the robot exhibits unsafe behavior that could damage the platform, such as hitting the table, colliding with the external frame, or self-collision. All evaluation videos are recorded for scoring.

Each video is scored independently by three evaluators under a double-blind protocol. The scoring considers both final task outcome and intermediate sub-step completion. The final trial score is obtained by averaging the three evaluator scores. To improve transparency and fairness, we release all evaluation videos and corresponding evaluator scores for leaderboard submissions, and allow users to appeal potential scoring errors.

\paragraph{Public evaluation videos and closed-source track.}
For transparency, RoboDojo releases evaluation videos for models included in the official verified leaderboard. These videos provide qualitative evidence of policy behavior and allow the community to inspect success cases, failure modes, and potential abnormal evaluation behavior. Results without the evaluated checkpoint, implementation, configuration files, and reproducibility instructions are reported only in a separate closed-source track, which is marked as non-official and unverified because the results cannot be independently reproduced or inspected by the community.

\section{Simulation Platform Details}
\label{appendix:simulation_platform_details}

This section provides additional implementation details of the RoboDojo simulation platform, including the configuration-driven task setup, digital asset processing, heterogeneous parallel simulation, and demonstration collection pipeline.

\subsection{Simulation Platform Foundation}
\label{appendix:Simulation_Platform}
The RoboDojo simulation platform is built upon the MagicSim infrastructure ~\citep{lu2026magicsim}, whose codebase provides the foundation of our simulation stack. RoboDojo adopts MagicSim as its core simulation infrastructure, inheriting its modular manager-based runtime design and physically based object abstraction system. Specifically, RoboDojo reuses MagicSim’s environment backbone for launching simulation, configuring physics and rendering, managing deterministic seeds, and exposing standardized hooks for task-specific scene setup, stepping, action application, and reset. Through MagicSim’s manager architecture, key simulation functions such as scene construction, object management, layout generation, camera configuration, data capture, and robot control are modularized into reusable components. RoboDojo further connects this infrastructure with Isaac Lab’s vectorized reinforcement learning interface, while preserving MagicSim’s manager-driven execution pattern for observation, reward, and termination computation.

RoboDojo also builds on MagicSim’s physically grounded object system, which provides unified abstractions for rigid bodies, articulated mechanisms, static geometry, garments, fluids, and scene fixtures. These MagicSim components encapsulate USD asset loading, PhysX-backed simulation setup, material configuration, state querying, and deterministic utility support, allowing RoboDojo to focus on robotic task construction rather than low-level simulator engineering. We thanks again to MagicSim who provides the foundational simulation architecture, object model, and simulation backend upon which RoboDojo’s embodied robotic environments are developed.

\subsection{Configuration-Driven Simulation Setup}
\label{appendix:simulation_platform_setup}

RoboDojo follows a configuration-first design, where each task is specified by modular YAML files rather than hard-coded simulator scripts. Each configuration defines task-relevant assets, distractor assets, object initialization distributions, robot initialization, camera settings, lighting conditions, background textures, articulation states, success conditions, and evaluation seeds. This design separates task specification from simulator execution, allowing different tasks to share the same runtime while changing task-specific objects, scene layouts, and randomization ranges.

At reset time, the simulator samples scene instances according to the task configuration under deterministic seed control. This allows RoboDojo to generate diverse scene layouts while remaining reproducible across evaluation runs. For example, the same task can be evaluated under different object poses, clutter configurations, lighting conditions, and background textures by changing the evaluation seed, while keeping the sampling process deterministic.

The platform supports rigid, articulated, and deformable assets through a unified scene construction pipeline. Rigid objects are loaded with collision geometry and physical parameters. Articulated objects are instantiated with joint configurations, joint limits, and initial articulation states. Deformable objects are instantiated with estimated material parameters and simulation-ready mesh representations. This unified pipeline enables different asset types to be used under the same task and policy interface.

\subsection{Digital Asset Processing and Validation}
\label{appendix:asset_library_details}

RoboDojo builds a digital asset library containing rigid, articulated, and deformable objects for task construction and clutter generation. For rigid objects, we collect assets from online repositories and reconstruct additional real-world daily objects using Meshy AI from reference images or text prompts. Reconstructed meshes are converted into simulation-ready USD assets and manually inspected in Isaac Sim before being added to the benchmark library.

Each asset is annotated with semantic and task-level metadata, including language descriptions, object categories, placement annotations, success-checking annotations, and manipulation affordances for automated data generation. These annotations support both task construction and demonstration generation. For example, graspable regions, placement regions, functional parts, and task-specific interaction points are used to ground low-level skills during automated trajectory synthesis and to define task-specific success checks.

For articulated objects, we inspect the kinematic structure, joint configuration, collision geometry, joint limits, and physical stability. For deformable objects, we assign estimated material parameters and realistic textures, and supplement the asset set with selected assets from ClothesNet~\citep{zhou2023clothesnet}. All assets are validated through simulation rollouts to reduce unstable contacts, incorrect collision geometry, and physically implausible behavior. This validation process improves the physical reliability of contact-rich manipulation tasks and reduces failures caused by asset artifacts rather than policy limitations.

\subsection{Heterogeneous Parallelism Implementation}
\label{appendix:heterogeneous_parallelism_details}

RoboDojo implements heterogeneous parallelism by disabling strict physics replication across cloned environments and allowing each environment to instantiate its own scene modules from seed-controlled configurations. Unlike homogeneous vectorized environments, where parallel instances usually share the same scene template, RoboDojo allows different environments in the same simulator process to contain different object categories, asset geometries, object counts, clutter layouts, articulation structures, and task configurations.

Each environment maintains an independent task state, random seed, robot initialization, observation stream, and success condition, while sharing the same batched stepping and action dispatch interface. This design preserves the efficiency of vectorized execution while allowing parallel evaluation over diverse scene configurations. For example, one environment may evaluate a rigid-object pick-and-place task with clutter distractors, while another may evaluate manipulation of an articulated object with a different joint structure.

For larger-scale evaluation and data collection, RoboDojo partitions episode seeds across multiple GPUs and launches independent simulator processes. Each process runs a subset of tasks and seeds, and the results are aggregated after evaluation. Combining intra-process heterogeneous environments with inter-process GPU sharding enables scalable evaluation without forcing all parallel environments to share the same scene template. This is especially important for evaluating generalist manipulation policies, since the benchmark should expose policies to diverse object geometries, layouts, and physical configurations rather than repeated rollouts of a single cloned scene.

\subsection{Simulation Demonstration Collection}
\label{appendix:simulation_data_collection_details}

RoboDojo supports two complementary demonstration collection modes: automated trajectory synthesis and VR-based teleoperation. Both modes use the same asset annotation layer, which specifies manipulation-related affordances and task semantics such as graspable regions, placement regions, functional parts, and task-specific interaction points. These annotations support automated skill grounding and task validation.

For automated trajectory synthesis, each task is decomposed into an ordered sequence of manually specified low-level skills. The skill library includes primitives such as \texttt{push\_up}, \texttt{place}, \texttt{handover}, \texttt{grasp}, \texttt{insert}, \texttt{open}, \texttt{close}, and \texttt{stack}. Each skill is grounded in the corresponding object annotations and executed using the cuRobo v2 motion planner~\citep{sundaralingam2026curobov2dynamicsawaremotiongeneration}, which generates physically feasible robot motions. This design enables complex demonstrations to be generated through reusable skill composition.

For tasks that are difficult to synthesize automatically, RoboDojo provides a VR-based teleoperation interface using devices such as Meta Quest and Pico. The system captures the delta 6D pose of the VR controller and maps it to target end-effector motion in simulation, which is solved online by cuRobo v2. For a single 6-DoF ARX X5 arm, spatial motion solving latency can reach approximately 3~ms per planning step. The gripper is controlled through controller buttons, while foot pedals are used to trigger task start and termination, fix robot arms, and provide other efficiency-oriented controls. This interface enables efficient collection of high-quality demonstrations for complex simulation tasks where automated synthesis is unreliable.

\section{RoboDojo-RealEval Platform Details}
\label{appendix:realeval_details}

\paragraph{Hardware setup.}
RoboDojo-RealEval uses a fixed-height white evaluation table with a 1.2\,m $\times$ 1.2\,m workspace. A replaceable white table cover maintains a consistent visual background and reduces replacement cost caused by surface damage during repeated evaluations. The workspace is enclosed by an external frame of 1.5\,m $\times$ 1.5\,m $\times$ 2.1\,m. Black curtains are mounted on the left, right, front, and top sides to reduce external illumination, while three linear LED light sources with fixed mounts provide stable lighting over the manipulation workspace. Custom structural components fix the relative geometry between the table, frame, robot arms, and cameras. A bottom connector fixes the relative distance, position, and orientation of the two robot arms, while a steel tube and custom 3D-printed mount fix the head-camera pose. The platform also includes a touchscreen interface and an integrated workstation box for the control computer and cooling system.

\paragraph{Layout replay and evaluation workflow.}
For each real-world task, we pre-collect reference evaluation layouts. During evaluation, the target layout image is overlaid with transparency on the live observation stream through the web interface. Evaluators compare the current scene with the reference layout and adjust object positions until the live observation aligns with the target image. This provides an intuitive mechanism for replaying initial layouts and reducing reset variation across policies and sessions. In our test with five data collectors, restoring a scene with five objects takes 14 seconds on average.

\paragraph{Safety control and scoring.}
After scene reset, the evaluator starts policy execution through the touchscreen interface. Since submitted policies may behave unexpectedly due to code errors or poor training, RoboDojo-RealEval provides an emergency stop function. Once triggered, policy control is immediately disabled and the robot slowly returns to a safe reset state. All observation videos are automatically collected and uploaded to the cloud for scoring. Each trajectory is scored by three independent raters who are blind to the identity of the evaluated policy. Samples with large scoring discrepancies are filtered, and the remaining scores are averaged to obtain the final result.

\subsection{Hardware Components of RoboDojo-RealEval}
\label{appendix:realeval_hardware_components}

\begin{figure}[h]
\centering
\includegraphics[width=1.0\textwidth]{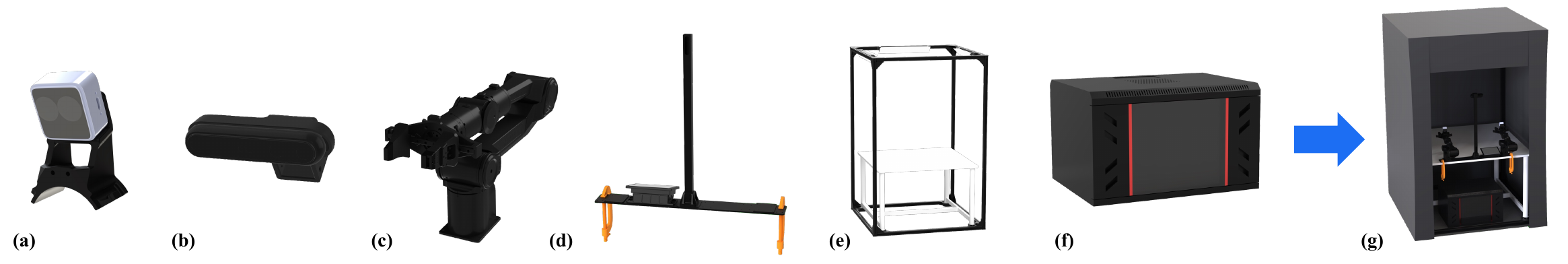}
\caption{\textbf{\textcolor{Blue_1}{Hardware components of RoboDojo-RealEval.}}
The platform consists of (a) two Gemini 305 wrist cameras, (b) one Gemini 335L head camera, (c) replaceable collaborative bimanual robot embodiments including ARX X5, Piper, and Piper X, (d) robot and camera support structure with a touchscreen interface, (e) external frame with LED lighting and curtains, (f) integrated workstation box, and (g) the assembled evaluation platform. These components standardize sensing, robot placement, illumination, and workspace layout for reproducible real-world evaluation.}
\label{fig:real_world_hardware_sperate}
\end{figure}

Fig.~\ref{fig:real_world_hardware_sperate} shows the hardware components of the \textbf{RoboDojo-RealEval} platform. The platform integrates wrist cameras, a head camera, replaceable collaborative bimanual robot embodiments, a robot and camera support structure, an external frame with controlled lighting and curtains, an integrated workstation box, and a touchscreen-based operation interface. These components jointly standardize sensing, robot placement, illumination, and workspace layout, providing a reproducible hardware foundation for real-world policy evaluation.

\section{XPolicyLab Design Details}
\label{appendix:xpolicylab_design}

XPolicyLab standardizes the external workflow of robot policy development while allowing each policy to preserve its own internal architecture and implementation. Its design covers three levels: data format, policy interface, and deployment protocol. This separation enables heterogeneous policies, including diffusion policies, VLA models, world-model-based policies, and classical imitation learning baselines, to be trained, debugged, deployed, and evaluated under the same RoboDojo protocol.

\paragraph{Data format and policy development.}
XPolicyLab converts raw RoboDojo trajectories into policy-specific training formats, such as HDF5 files, LeRobot-style datasets, precomputed language embeddings, and model-specific normalization statistics. At the benchmark level, each trajectory follows a unified observation and action format. Each observation may contain language instructions, multi-view visual inputs, robot states, and metadata. Visual inputs include head, wrist, and third-view cameras, while robot states may include joint positions, end-effector poses, TCP poses, gripper states, and mobile base states. All pose values follow the same convention, represented as position and quaternion in the form $[x, y, z, q_w, q_x, q_y, q_z]$. This unified format separates policy-specific data requirements from the benchmark protocol, allowing different policies to be connected through lightweight adapters and compared under consistent task, robot, action, and seed settings.

\paragraph{Policy interface and debugging.}
XPolicyLab defines a common policy interface through a standardized \texttt{Model} abstraction. Each policy implements a small set of required methods, including \texttt{update\_obs}, \texttt{get\_action}, and \texttt{reset}. For parallel simulation evaluation, XPolicyLab additionally supports batched interfaces such as \texttt{update\_obs\_batch} and \texttt{get\_action\_batch}. The environment client only calls these standardized functions and does not need to access the internal policy architecture. XPolicyLab also provides an offline debug mode that generates correctly shaped observations and validates returned actions, allowing developers to test model loading, observation parsing, action dimensions, parameter passing, and server-client communication before full simulation or real-world evaluation.

\paragraph{Package organization and deployment protocol.}
XPolicyLab standardizes the organization of each policy package. A policy package is expected to provide installation, data processing, training, model serving, and evaluation scripts under a consistent structure, such as \texttt{install.sh}, \texttt{process\_data.sh}, \texttt{train.sh}, \texttt{eval.sh}, \texttt{model.py}, \texttt{deploy.py}, and \texttt{deploy.yml}. The same set of experiment parameters, including dataset name, task name, checkpoint name, robot configuration, action type, and random seed, is used across data processing, training, and evaluation. This convention improves reproducibility and makes it easier to inspect, rerun, and compare experiments across policies.

\paragraph{Communication protocol.}
\label{communication_protocol}
XPolicyLab implements policy-environment communication through a lightweight client-server protocol based on \textbf{WebSocket} and \textbf{MessagePack}. The policy runs as a model server, while the simulator or real robot runs as an environment client. WebSocket provides persistent bidirectional communication for low-latency action queries, and MessagePack serializes observations and actions into a compact binary format that supports images, robot states, nested dictionaries, and action commands. This protocol enables efficient remote communication across offline debugging, heterogeneous parallel simulation, and cloud-based real-world evaluation, while keeping the policy runtime independent of the underlying evaluation environment.

\paragraph{Evaluation loop.}
The standard evaluation loop is shown in Table~\ref{tab:xpolicylab_eval_protocol}. At the beginning of each episode, the environment resets the policy state. The environment then repeatedly obtains an observation, sends it to the policy server, requests an action chunk, and executes the returned actions until the episode terminates. The same logic is extended to batched simulation evaluation by maintaining a list of active environment indices and querying the policy with batched observations. This minimal interface supports single-environment evaluation, vectorized simulation, action chunking, and remote real-world deployment.

\input{table/standard}

\section{Simulation Benchmark Tasks Details}
\label{appendix:sim_benchmark_tasks_details}

We provide a complete specification for each simulation task, including the task name, the language instructions used for data collection and evaluation, the task description, the source of expert demonstrations, and the task split indicating whether it is used for training or testing. This task-level documentation is intended to make the benchmark transparent and reproducible, and to help users clearly understand the evaluation protocol of each task. More details are available at \href{https://robodojo-benchmark.com/doc}{https://robodojo-benchmark.com/doc}.

\input{table/sim_task_details}

\section{Real-World Benchmark Tasks Details}
\label{appendix:real_benchmark_tasks_details}

We provide a complete specification for each real-world task, including the task name, the language instructions used for data collection and evaluation, the task description, the source of expert demonstrations, and the task split indicating whether it is used for training or testing. This task-level documentation makes the real-world benchmark transparent and reproducible, and helps users understand the standardized evaluation protocol for each physical task. More details are available at \texttt{RoboDojo-Benchmark.com/doc}.

\input{table/real_task_details}

\section{Policy Training Details}
\label{appendix:train_details}

In this section, we provide the training settings for all evaluated policies in RoboDojo. For each policy, we report the initialization checkpoint, training data, batch size, number of training steps, and other model-specific configurations when applicable. These details are included to improve reproducibility and to clarify how each policy is trained for simulation and real-world evaluation.

\input{sec/benchmark_experiment/train_details}



%% file: table/compare.tex
\begin{table*}[!ht]
\centering
\caption{\textbf{Comparison between RoboDojo and other benchmarks.}
``Skill'' is the number of operation primitives covered; ``Num. Policies'' is the number of baseline policies the benchmark integrates/evaluates. N/R indicates not reported. HP indicates heterogeneous parallelization.}
\label{tab:robodojo_simulation_benchmark_compare}
\begingroup
\footnotesize
\setlength{\tabcolsep}{3.2pt}
\renewcommand{\arraystretch}{1.08}
\definecolor{bestblue}{RGB}{210,243,247}
\newcommand{\thickline}{\noalign{\vspace{0.25em}\hrule height 1.2pt\vspace{0.25em}}}
\resizebox{\textwidth}{!}{
\begin{tabular}{l|c|c|c|c|c|c|c|c}
\thickline
\textbf{Benchmark} &
\textbf{Platform} &
\textbf{Skill} &
\textbf{HP} &
\textbf{Bimanual} &
\textbf{Tasks} &
\textbf{Data Source} &
\textbf{Num. Policies} &
\textbf{Sim-and-Real} \\
\midrule
RoboTwin 1.0~\citep{mu2025robotwin} & SAPIEN & 7 & $\times$ & $\checkmark$ & 15 & Teleop+DataGen & 4 & $\times$ \\
RoboTwin 2.0~\citep{chen2025robotwin} & SAPIEN & 13 & $\times$ & $\checkmark$ & 50 & DataGen & 11 & $\times$ \\
RMBench~\citep{chen2026rmbench} & SAPIEN & 3 & $\times$ & $\checkmark$ & 9 & Teleop+DataGen & 5 & $\times$ \\
AutoBio~\citep{lan2025autobio} & MuJoCo & 3 & $\times$ & $\checkmark$ & 16 & DataGen & 2 & $\times$ \\
DexJoCo~\citep{wang2026dexjoco} & MuJoCo & 12 & $\times$ & $\checkmark$ & 11 & Teleop & 4 & $\times$ \\
LIBERO~\citep{liu2023libero} & MuJoCo & 6 & $\times$ & $\times$ & 130 & Teleop & 3 & $\times$ \\
CALVIN~\citep{mees2022calvin} & PyBullet & 10 & $\times$ & $\times$ & 34 & Teleop & 1 & $\times$ \\
RLBench~\citep{james2019rlbench} & CoppeliaSim & 18 & $\times$ & $\times$ & 100 & DataGen & N/R & $\times$ \\
RoboCasa365~\citep{nasiriany2026robocasa365} & MuJoCo & 8 & $\times$ & $\times$ & 365 & Teleop+DataGen & 4 & $\times$ \\
SimplerEnv~\citep{li2024evaluating} & SAPIEN & 4 & $\times$ & $\times$ & 8 & Teleop & 4 & $\times$ \\
VLABench~\citep{zhang2025vlabench} & MuJoCo & 10 & $\times$ & $\times$ & 100 & DataGen & 3 & $\times$ \\
BEHAVIOR-1K~\citep{li2023behavior} & Isaac Sim & 19 & $\times$ & $\checkmark$ & 1000 & Teleop & 6 & $\times$ \\
RoboSuite~\citep{zhu2020robosuite} & MuJoCo & 9 & $\times$ & $\checkmark$ & 9 & Teleop & 1 & $\times$ \\
Meta-World~\citep{yu2020meta} & MuJoCo & 13 & $\times$ & $\times$ & 50 & DataGen & 7 & $\times$ \\
EBench~\citep{gao2026ebench} & Isaac Sim & 11 & $\times$ & $\checkmark$ & 26 & Teleop+DataGen & 4 & $\times$ \\
RoboLab~\citep{yang2026robolab} & Isaac Lab & 4 & $\times$ & $\times$ & 120 & Teleop & 5 & $\times$ \\
ManipulationNet~\citep{chen2026manipulationnet} & Real Robot & 5 & $\times$ & $\checkmark$ & 5 & N/R & N/R & $\times$ \\
RoboArena~\citep{atreya2025roboarena} & Real Robot & 11 & $\times$ & $\times$ & N/R & N/R & 7 & $\times$ \\
RoboChallenge~\citep{yakefu2025robochallenge} & Real Robot & 13 & $\times$ & $\checkmark$ & 30 & Teleop & 5 & $\times$ \\
\midrule
RoboDojo (ours) & Isaac Sim & 24 & $\checkmark$ & $\checkmark$ & 60 & Teleop+DataGen & 35+ & $\checkmark$ \\
\thickline
\end{tabular}
}
\endgroup
\end{table*}

%% file: table/standard.tex

\begin{table*}[t]
\centering
\caption{\textbf{\textcolor{Blue_1}{XPolicyLab unified policy interface and evaluation protocol.}}
XPolicyLab provides a shared policy-side interface for RoboDojo simulation and real-world evaluation. 
The same policy implementation can be deployed in single-environment real-world evaluation and batched simulation evaluation, where the batched interface enables heterogeneous parallel execution for faster feedback.}
\label{tab:xpolicylab_eval_protocol}
\small
\setlength{\tabcolsep}{5pt}
\renewcommand{\arraystretch}{1.18}
\begin{tabularx}{\textwidth}{p{0.16\textwidth}X X}
\toprule
\textbf{\textcolor{Blue_1}{Stage}} 
& \textbf{\textcolor{Blue_1}{Single-Environment Deployment}} 
& \textbf{\textcolor{Blue_1}{Batched Parallel Deployment}} \\
\midrule

Policy initialization
& \texttt{policy.reset()} resets the policy state before each episode. This mode is used for real-world evaluation and standard single-task simulation. 
& \texttt{policy.reset()} resets the policy state before evaluating multiple simulation environments in parallel. \\

Observation update
& XPolicyLab receives the latest observation $o_t$ from the RoboDojo environment and updates the policy by \texttt{policy.update\_obs($o_t$)}. 
& XPolicyLab collects active environment indices $\mathcal{I}_t$, receives batched observations $\mathbf{o}_{\mathcal{I}_t}$, and updates the policy by \texttt{policy.update\_obs\_batch($\mathbf{o}_{\mathcal{I}_t}$)}. \\

Action prediction
& The policy predicts an action chunk $\mathbf{a}_{t:t+K}$ through \texttt{policy.get\_action()}. The same interface is used for simulation and real-world deployment. 
& The policy predicts batched action chunks $\mathbf{a}_{\mathcal{I}_t,t:t+K}$ through \texttt{policy.get\_action\_batch($\mathcal{I}_t$)}, allowing multiple environments to share one policy service. \\

Environment execution
& RoboDojo executes each action in the chunk through \texttt{env.step($a_k$)} until the episode terminates or the action chunk is exhausted. 
& RoboDojo executes the $k$-th action for all active environments through \texttt{env.step\_batch($\mathbf{a}_{\mathcal{I}_t,k}$, $\mathcal{I}_t$)}. Finished environments are removed from the active set. \\

Evaluation role
& Used for remote real-world evaluation in RoboDojo-RealEval, where policies are tested under standardized and reproducible physical conditions. 
& Used for RoboDojo simulation evaluation, where different tasks, scenes, and processes run concurrently to improve evaluation speed and support rapid policy iteration. \\

\bottomrule
\end{tabularx}
\end{table*}

%% file: table/sim_task_details.tex
\subsection{Generalization}

\taskshowcase
{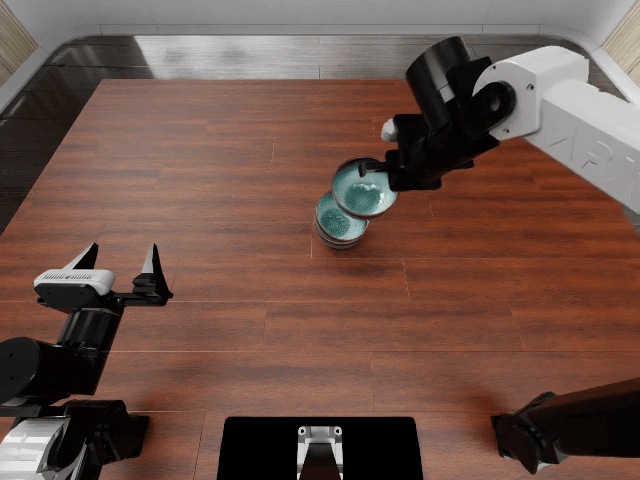}
{stack\_bowls}
{
\taskitem{Instruction}{
Stack the three bowls together.
}

\taskitem{Description}{There are three bowls. The robot needs to stack all the bowls together.}

\taskitem{Data Source}{Teleop}

\taskitem{Usage}{Train \& Eval}
}

\taskshowcase
{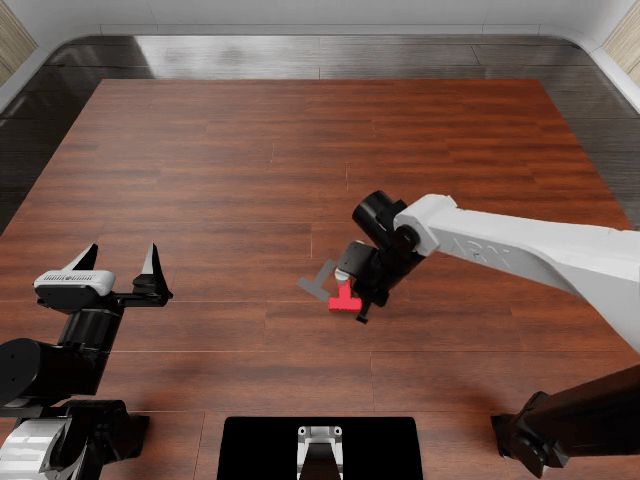}
{push\_T}
{
\taskitem{Instruction}{
Push the T-shaped block to align it precisely with the gray T-shaped pad.
}

\taskitem{Description}{There is a thin gray T-shaped pad and a T-shaped block. The robot needs to push the T-shaped block until it is precisely aligned with and fitted onto the gray pad.}

\taskitem{Data Source}{Teleop}

\taskitem{Usage}{Train \& Eval}
}

\taskshowcase
{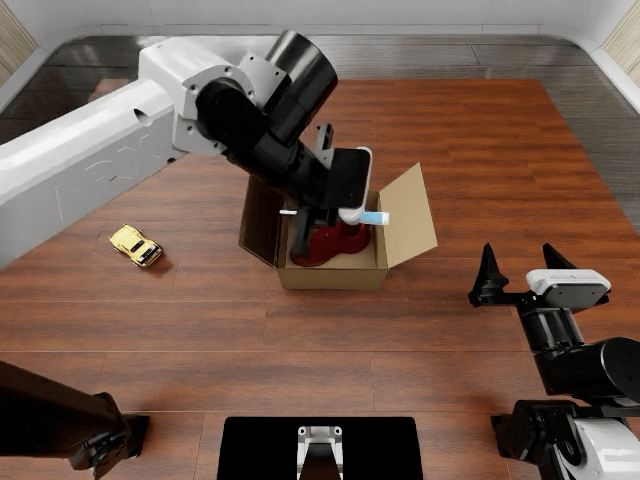}
{pack\_objects\_into\_box}
{
\taskitem{Instruction}{
Place all the objects into the box with their front sides facing left.
}

\taskitem{Description}{There are several objects on the table and a box. The robot needs to pick up all the objects, place them into the box, and ensure that each object is oriented with its front side facing left.}

\taskitem{Data Source}{Teleop}

\taskitem{Usage}{Train \& Eval}
}

\taskshowcase
{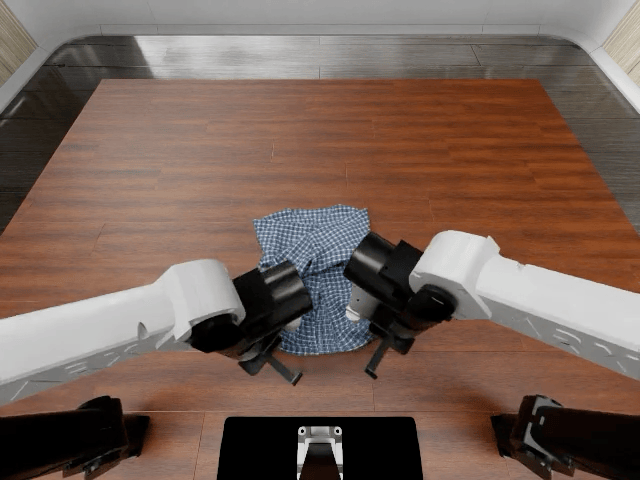}
{fold\_clothes}
{
\taskitem{Instruction}{
Fold the clothes neatly.
}

\taskitem{Description}{There is a piece of clothing. The robot needs to fold it neatly.}

\taskitem{Data Source}{AutoGen}

\taskitem{Usage}{Train \& Eval}
}

\taskshowcase
{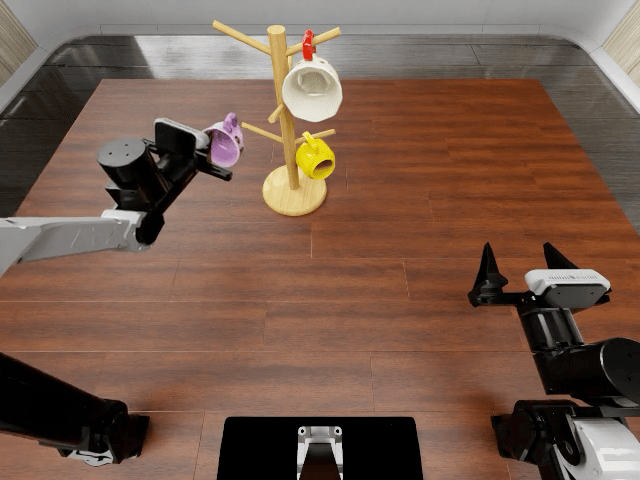}
{hang\_mugs}
{
\taskitem{Instruction}{
Hang all the mugs on the mug rack.
}

\taskitem{Description}{There are three mugs and one mug rack. The robot needs to pick up each mug and hang all of them on the rack.}

\taskitem{Data Source}{Teleop}

\taskitem{Usage}{Train \& Eval}
}

\taskshowcase
{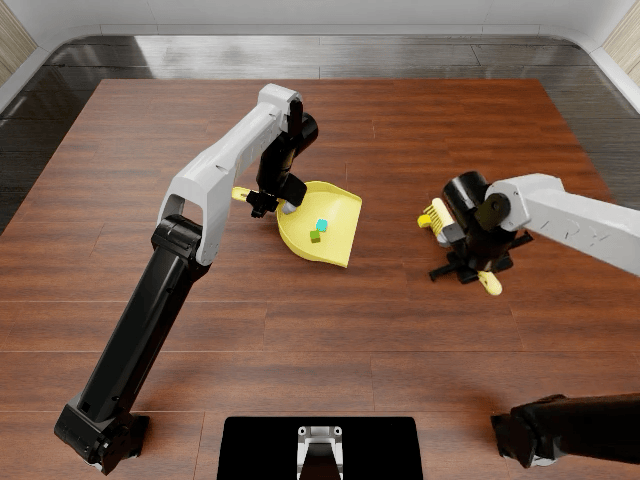}
{sweep\_blocks}
{
\taskitem{Instruction}{
Pick up the broom, hand it over to the right hand, then use the dustpan to sweep the blocks.
}

\taskitem{Description}{There is a broom and a dustpan on the left side, and the blocks are on the right side. The robot needs to first pick up the broom, hand it over to the right hand, then grasp the dustpan with the left hand, and finally sweep the blocks successfully.}

\taskitem{Data Source}{Teleop}

\taskitem{Usage}{Train \& Eval}
}

\taskshowcase
{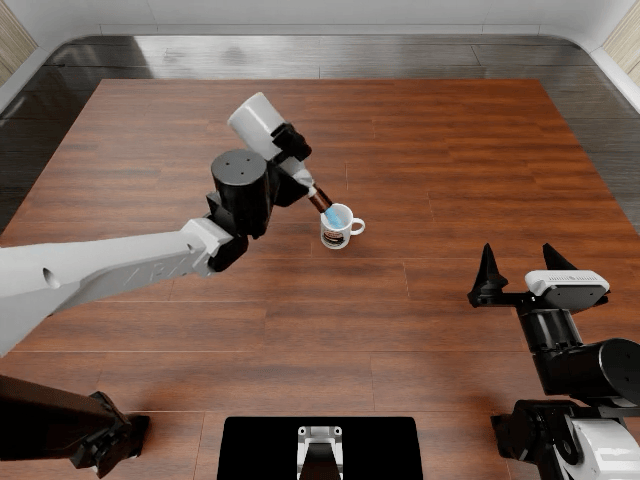}
{pour\_liquid\_into\_cup}
{
\taskitem{Instruction}{
Pour the liquid from the bottle into the cup.
}

\taskitem{Description}{There is a bottle containing liquid and a cup. The robot needs to pick up the bottle and pour the liquid into the cup.}

\taskitem{Data Source}{Teleop}

\taskitem{Usage}{Train \& Eval}
}

\taskshowcase
{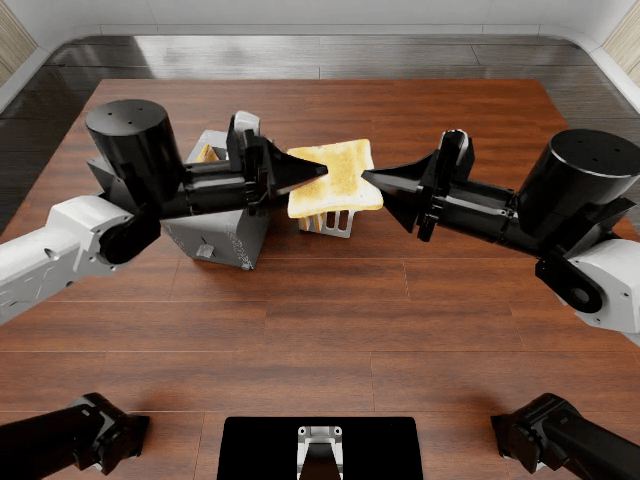}
{make\_toast}
{
\taskitem{Instruction}{
Pick up two slices of bread, place them into the toaster, and press the lever down.
}

\taskitem{Description}{There is a basket containing multiple slices of bread and a toaster. The robot needs to pick up two slices one by one, place them into the toaster, and then press the lever down to start toasting.}

\taskitem{Data Source}{Teleop}

\taskitem{Usage}{Train \& Eval}
}

\taskshowcase
{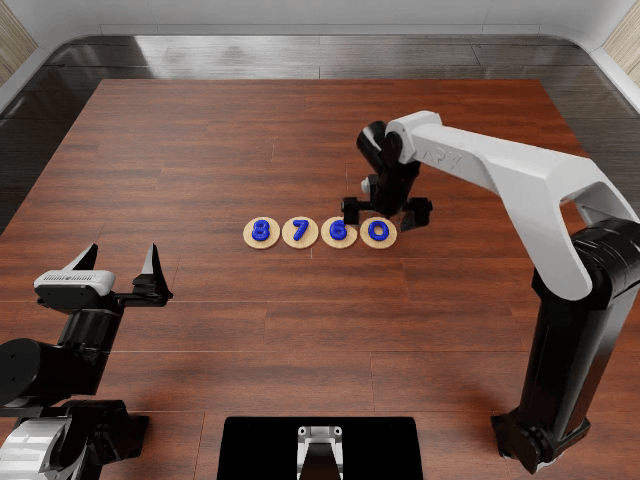}
{arrange\_largest\_number}
{
\taskitem{Instruction}{
Arrange the numbers from left to right to form the largest possible number, and place them on the pad.
}

\taskitem{Description}{There are several number tiles on the table and a pad for placement. The robot needs to determine the order that forms the largest possible number, then place the four numbers on the pad from left to right in that order.}

\taskitem{Data Source}{AutoGen}

\taskitem{Usage}{Train \& Eval}
}

\taskshowcase
{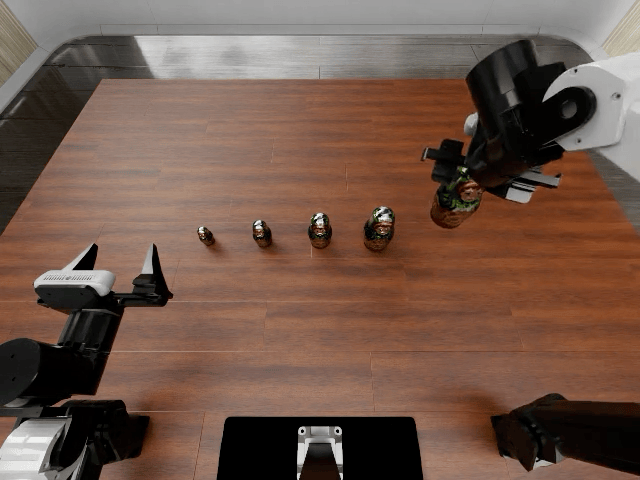}
{sort\_nesting\_dolls\_by\_size}
{
\taskitem{Instruction}{
Arrange the five nesting dolls in a row from left to right in descending size order.
}

\taskitem{Description}{There are five nesting dolls of different sizes on the table. The robot needs to sort them by size and arrange them in a straight row from left to right, from largest to smallest.}

\taskitem{Data Source}{AutoGen}

\taskitem{Usage}{Train \& Eval}
}

\taskshowcase
{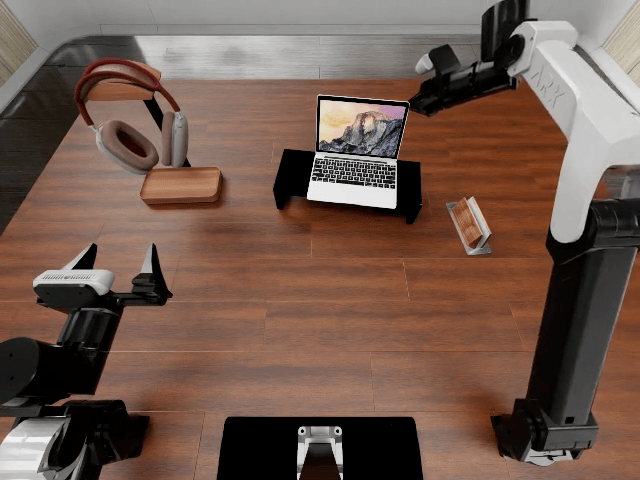}
{store\_laptop\_and\_headphones}
{
\taskitem{Instruction}{
Hang the headphones on the headphone stand, close the laptop, then place it into the vertical laptop stand.
}

\taskitem{Description}{There is an open laptop on a laptop stand, a vertical laptop stand, a pair of headphones, and a headphone stand. The laptop opening angle is random but greater than 30 degrees. The robot needs to first hang the headphones on the headphone stand, then close the laptop, pick it up from the stand, and insert it into the vertical laptop stand.}

\taskitem{Data Source}{Teleop}

\taskitem{Usage}{Train \& Eval}
}

\taskshowcase
{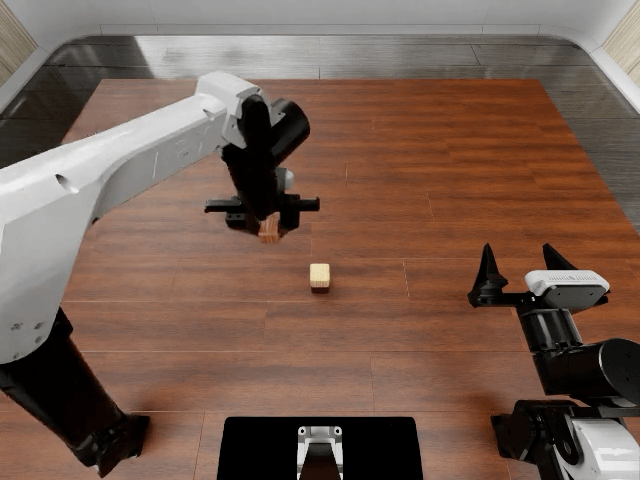}
{stack\_blocks}
{
\taskitem{Instruction}{
Stack the three blocks with different textures.
}

\taskitem{Description}{There are three blocks with different textures on the table. The robot needs to pick them up and stack them into a stable pile.}

\taskitem{Data Source}{AutoGen}

\taskitem{Usage}{Train \& Eval}
}

\subsection{Open}

\taskshowcase
{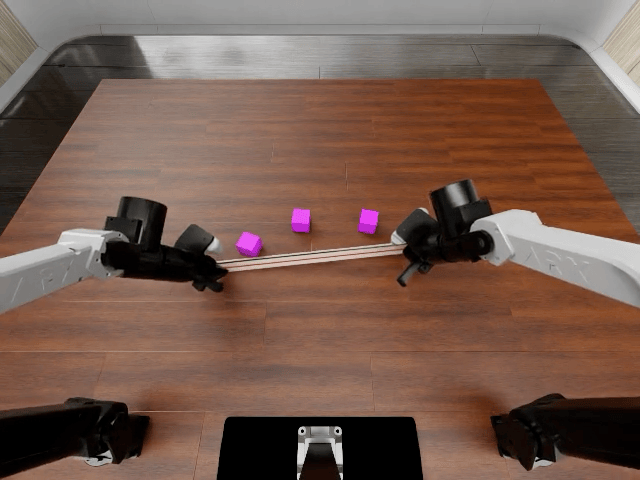}
{align\_blocks}
{
\taskitem{Instruction}{
Use the set square to push the three blocks into a straight, aligned row, then reset the robot arm.
}

\taskitem{Description}{There is a set square and three blocks. The robot needs to use the set square to push the blocks until they are aligned in parallel in a straight row.}

\taskitem{Data Source}{null (eval-only)}

\taskitem{Usage}{Eval}
}

\taskshowcase
{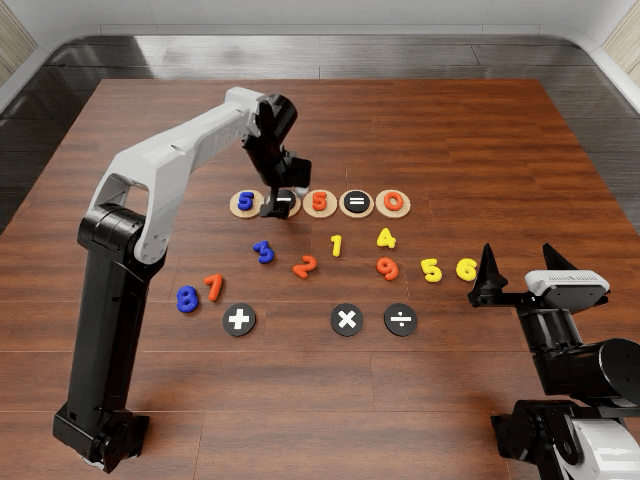}
{solve\_equation}
{
\taskitem{Instruction}{
Complete the equation by selecting the correct missing number or operator and placing it on the pad, then reset the robot arm.
}

\taskitem{Description}{There is an arithmetic equation on the table and a pad for the answer. The equation is missing either a number or an operator. The robot needs to choose the correct missing item from the randomly arranged numbers and operators on the table and place it on the pad to complete the equation.}

\taskitem{Data Source}{null (eval-only)}

\taskitem{Usage}{Eval}
}

\taskshowcase
{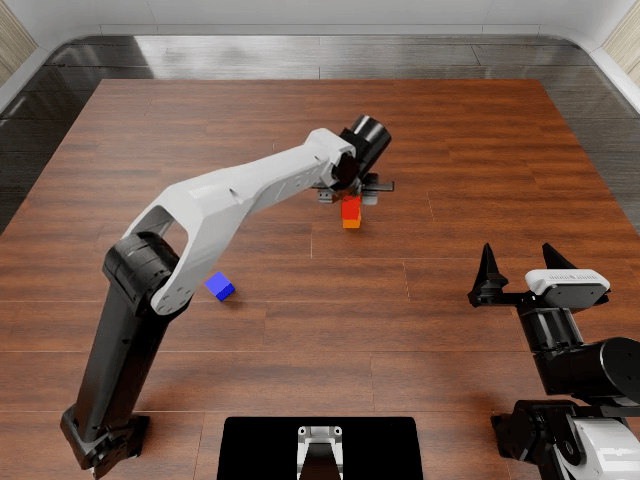}
{stack\_blocks\_by\_language}
{
\taskitem{Instruction}{
Stack the three blocks on top of each other in the order of {}, {}, and {}, then reset the robot arm.
}

\taskitem{Description}{There are several colored blocks. The robot needs to understand the language instruction and stack the blocks in the specified color order.}

\taskitem{Data Source}{null (eval-only)}

\taskitem{Usage}{Eval}
}

\taskshowcase
{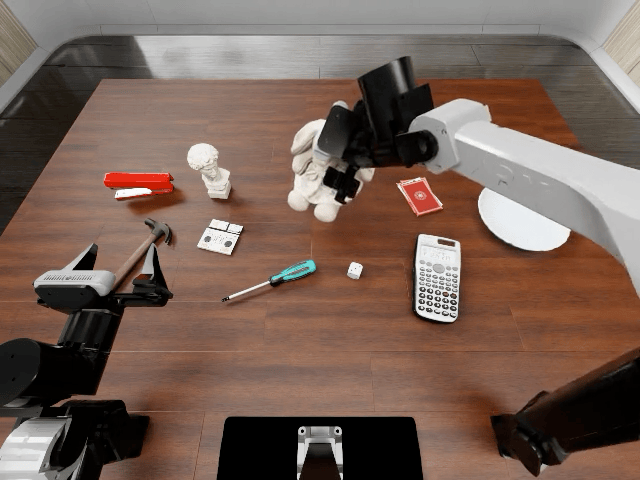}
{general\_pickup}
{
\taskitem{Instruction}{
Pick up the {} by 10 cm.
}

\taskitem{Description}{There are multiple objects. The robot needs to understand the language instruction, identify the target object, and pick it up.}

\taskitem{Data Source}{null (eval-only)}

\taskitem{Usage}{Eval}
}

\taskshowcase
{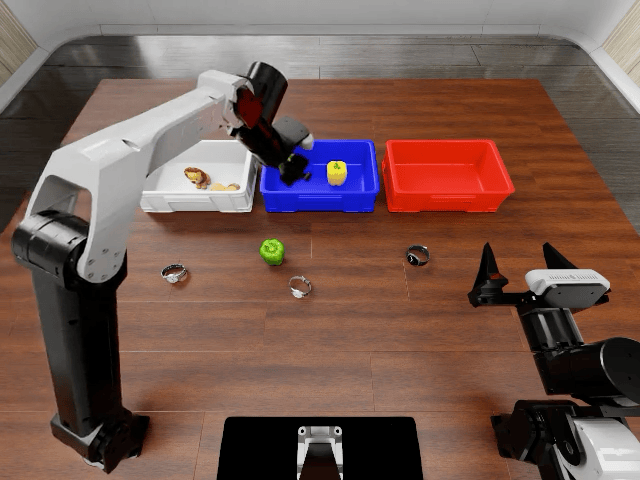}
{classify\_objects\_by\_language}
{
\taskitem{Instruction}{
Put {} objects into the left basket, {} objects into the middle basket, and {} objects into the right basket, then reset the robot arm.
}

\taskitem{Description}{There are three baskets and three categories of unseen objects. The robot needs to understand the language instruction, identify the category of each object, and place the objects into the specified baskets from left to right.}

\taskitem{Data Source}{null (eval-only)}

\taskitem{Usage}{Eval}
}

\taskshowcase
{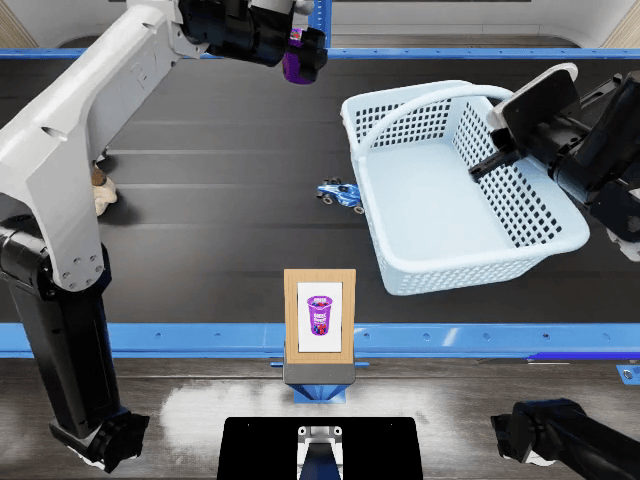}
{pick\_from\_conveyor\_by\_image}
{
\taskitem{Instruction}{
Lift the basket more than 8 cm, identify the target object on the conveyor according to the image on the board, pick it up, and place it into the basket.
}

\taskitem{Description}{There is a board displaying an image of the target object, a basket, and a conveyor carrying multiple objects. The robot needs to first lift the basket more than 8 cm, identify the target object on the conveyor based on the image shown on the board, pick up the target object, and place it into the basket.}

\taskitem{Data Source}{null (eval-only)}

\taskitem{Usage}{Eval}
}

\taskshowcase
{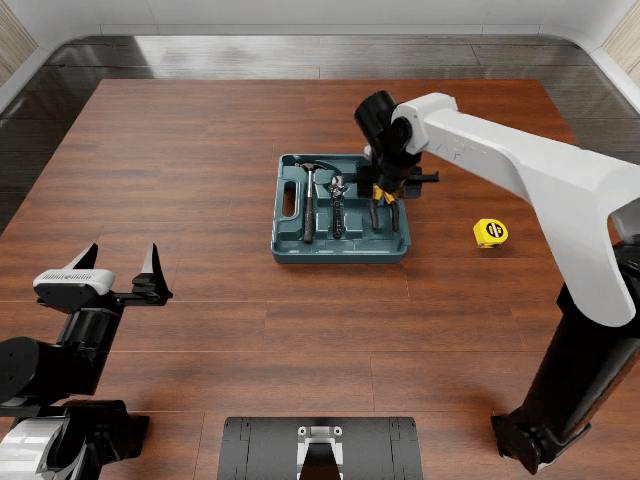}
{store\_tools\_in\_toolbox}
{
\taskitem{Instruction}{
Place each tool into its matching position in the toolbox, then reset the robot arm.
}

\taskitem{Description}{There are several tools and a toolbox with designated slots. The robot needs to pick up each tool and place it into the corresponding position in the toolbox.}

\taskitem{Data Source}{null (eval-only)}

\taskitem{Usage}{Eval}
}

\taskshowcase
{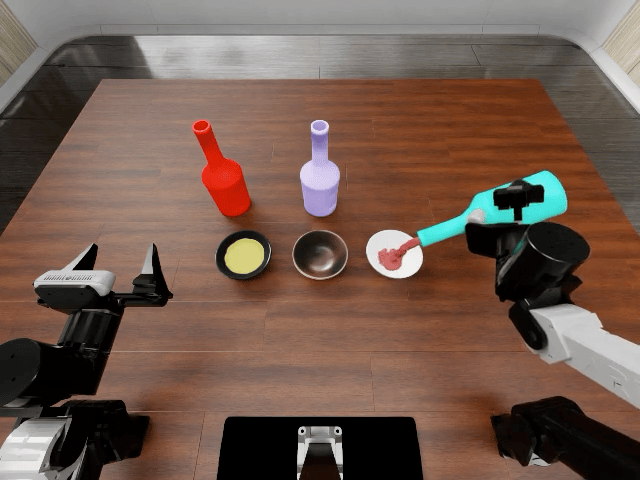}
{pour\_by\_language}
{
\taskitem{Instruction}{
Pour the liquid from the first {} bottle into the first {} bowl, from the second {} bottle into the second {} bowl, and from the third {} bottle into the third {} bowl. Then reset the robot arm.
}

\taskitem{Description}{There are several bottles, cups, and containers on the table. The robot needs to follow the language instruction and pour the liquid from each specified bottle into the corresponding container in the correct order.}

\taskitem{Data Source}{null (eval-only)}

\taskitem{Usage}{Eval}
}

\subsection{Memory}

\taskshowcase
{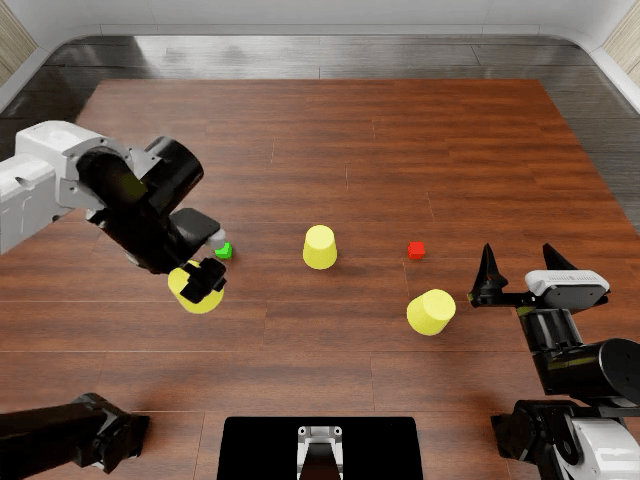}
{cover\_blocks}
{
\taskitem{Instruction}{
Cover the blocks from left to right, remember their colors, then uncover them in the order: red, green, and blue.
}

\taskitem{Description}{There are three covers and three blocks arranged in a random order. The blocks are red, green, and blue. The robot needs to cover the blocks from left to right, remember the color under each cover, and then uncover the blocks in the order of red, green, and blue.}

\taskitem{Data Source}{AutoGen}

\taskitem{Usage}{Train \& Eval}
}

\taskshowcase
{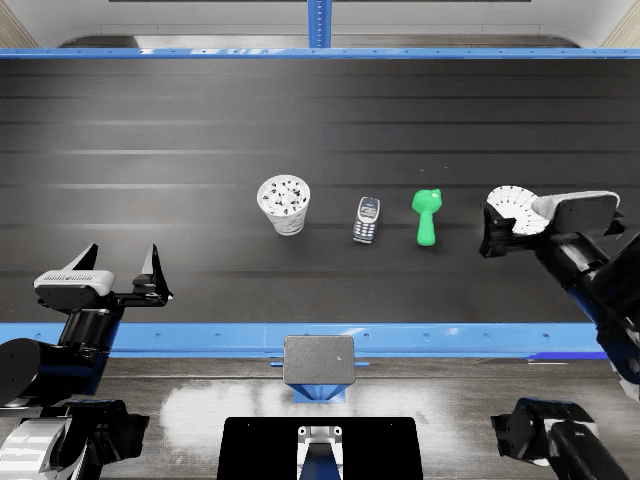}
{match\_and\_pick\_from\_conveyor}
{
\taskitem{Instruction}{
Remember the first object on the conveyor, then pick the matching object when it appears again.
}

\taskitem{Description}{An object first appears on the conveyor and is carried away. The robot needs to remember this object, observe the following objects on the conveyor, and pick the one that matches the first object.}

\taskitem{Data Source}{Teleop}

\taskitem{Usage}{Train \& Eval}
}

\taskshowcase
{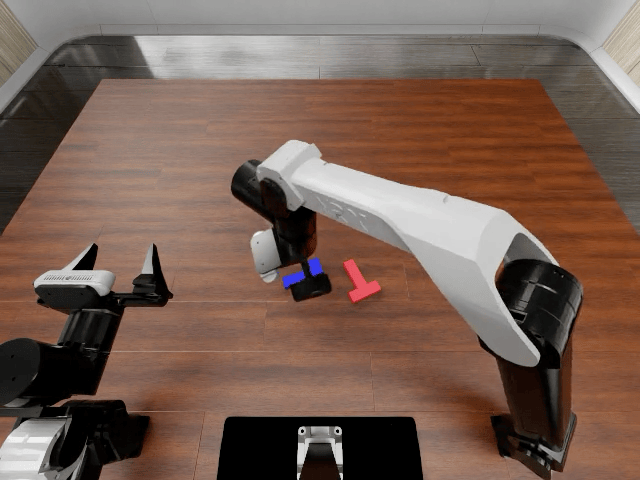}
{swap\_T}
{
\taskitem{Instruction}{
Pick up the two T-shaped blocks, swap their positions, and place them back with the correct orientations.
}

\taskitem{Description}{There are two T-shaped blocks on the table. The robot needs to grasp them with both hands, swap their positions, and place them back so that each block matches the original pose of the other one, including both position and orientation. This is a memory-based task.}

\taskitem{Data Source}{AutoGen}

\taskitem{Usage}{Train \& Eval}
}

\taskshowcase
{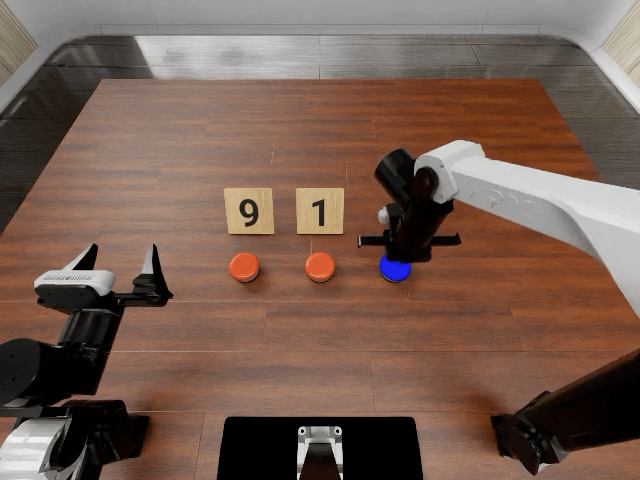}
{press\_by\_number}
{
\taskitem{Instruction}{
Press the two red buttons the required number of times according to the number cards, then press the blue button to confirm.
}

\taskitem{Description}{There are two number cards, two red buttons, and one blue confirmation button. The robot needs to read the numbers, press each red button the corresponding number of times, and then press the blue button to confirm. Pressing the blue button ends the task immediately.}

\taskitem{Data Source}{AutoGen}

\taskitem{Usage}{Train \& Eval}
}

\taskshowcase
{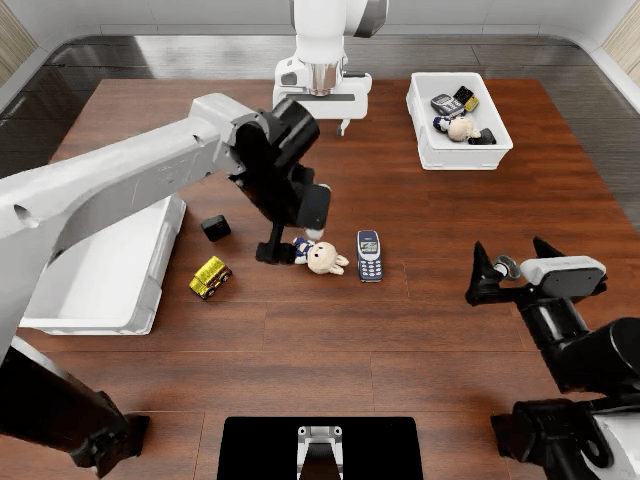}
{imitate\_sorting\_sequence}
{
\taskitem{Instruction}{
Observe the object placement order, remember it, then place the corresponding objects into the basket in the same order.
}

\taskitem{Description}{There are five categories of objects, with five objects on each side. The opposite robot places its objects into the basket on the right side in a certain order. The robot needs to observe and remember this sequence, then place its corresponding objects into the basket in the same order. This is a memory-based imitation task.}

\taskitem{Data Source}{AutoGen}

\taskitem{Usage}{Train \& Eval}
}

\taskshowcase
{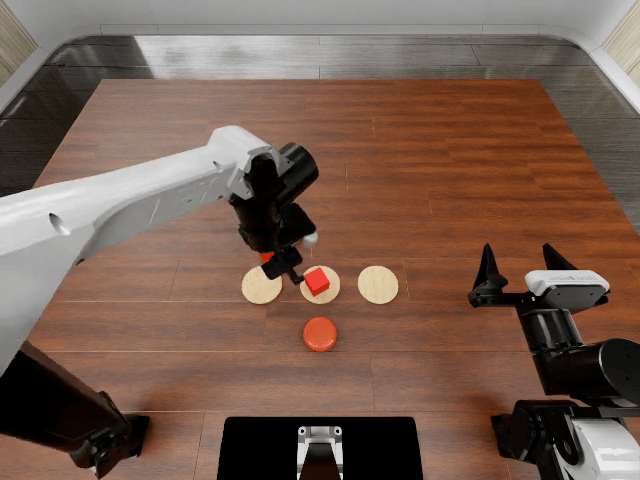}
{swap\_blocks}
{
\taskitem{Instruction}{
Swap the two blocks using the empty mat, pressing the button after each move.
}

\taskitem{Description}{There are three mats, two blocks placed on two of the mats, and one button. The robot needs to swap the positions of the two blocks by using the empty mat as a temporary place. After each move, the robot must press the button.}

\taskitem{Data Source}{AutoGen}

\taskitem{Usage}{Train \& Eval}
}

\subsection{Precision}

\taskshowcase
{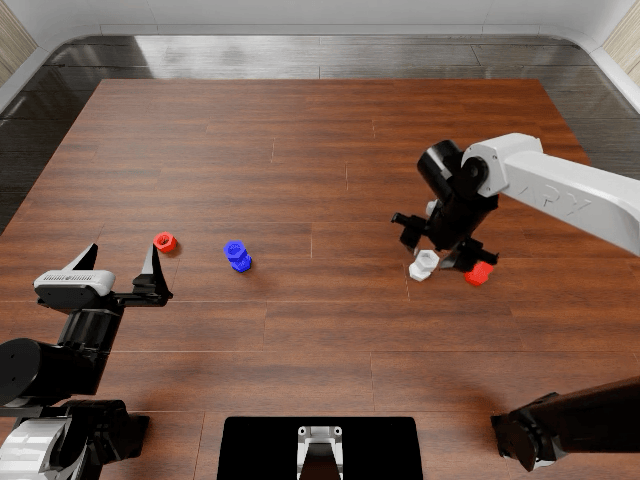}
{fasten\_screws}
{
\taskitem{Instruction}{
Insert and tighten each screw into the nut of the same color.
}

\taskitem{Description}{There are three screws and three nuts on the table. The screws and nuts come from five possible colors: red, blue, gray, yellow, and purple. In each episode, three different colors are selected, and the robot needs to match each screw with the nut of the same color, insert it, and tighten it. The screw positions vary within a small range, while the nut positions vary within a larger range. If a handover is needed, the robot can first place the screw in the middle and then let the other arm pick it up and fasten it.}

\taskitem{Data Source}{AutoGen}

\taskitem{Usage}{Train \& Eval}
}

\taskshowcase
{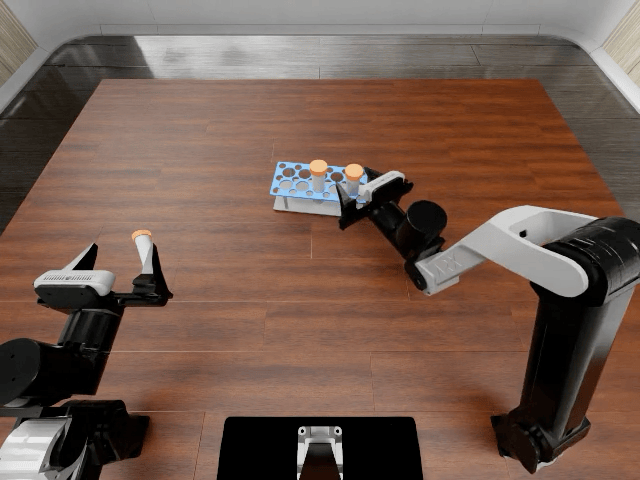}
{insert\_tubes}
{
\taskitem{Instruction}{
Insert the three tubes into the rack one by one.
}

\taskitem{Description}{There is a tube rack and three tubes. The robot needs to pick up each tube in sequence and insert all tubes into the rack.}

\taskitem{Data Source}{AutoGen}

\taskitem{Usage}{Train \& Eval}
}

\taskshowcase
{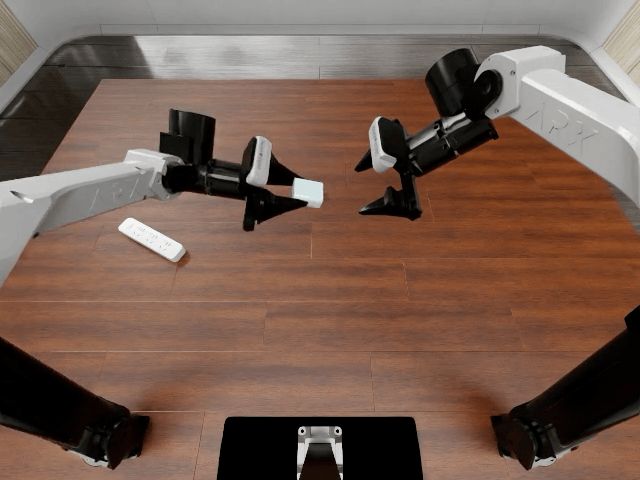}
{plug\_in\_charger}
{
\taskitem{Instruction}{
Plug the charger into the power strip.
}

\taskitem{Description}{There is a charger plug and a power strip. The robot needs to pick up the charger plug and insert it into the power strip.}

\taskitem{Data Source}{AutoGen}

\taskitem{Usage}{Train \& Eval}
}

\taskshowcase
{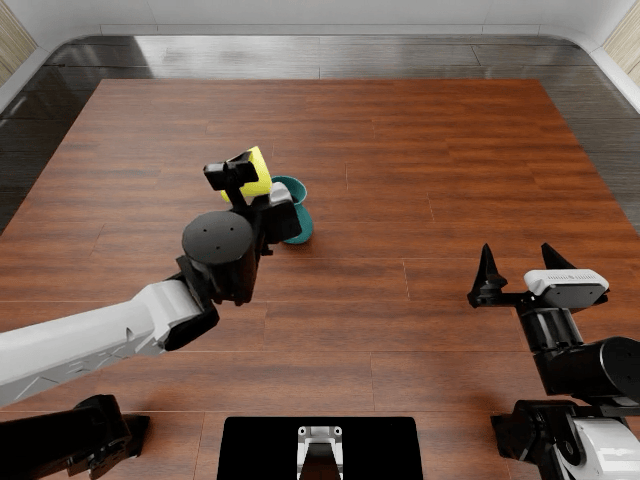}
{pour\_balls\_into\_vase}
{
\taskitem{Instruction}{
Pour all the balls from the cup into the vase.
}

\taskitem{Description}{There is a cup containing many small balls and a vase. The robot needs to pick up the cup and pour all the balls into the vase.}

\taskitem{Data Source}{Teleop}

\taskitem{Usage}{Train \& Eval}
}

\taskshowcase
{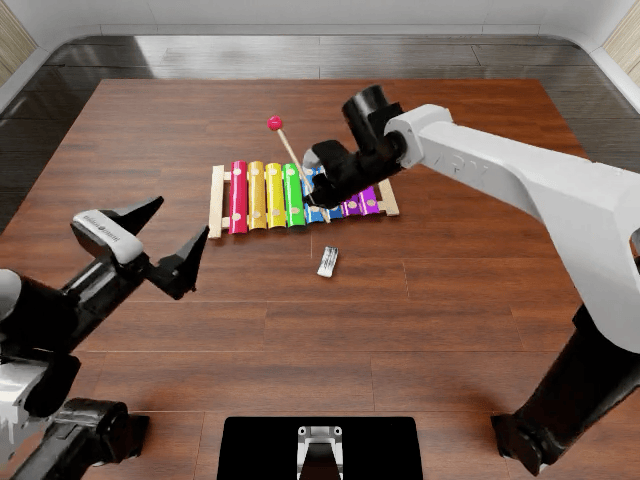}
{play\_Xylophone}
{
\taskitem{Instruction}{
Pick up the mallet and strike all xylophone keys from left to right.
}

\taskitem{Description}{There is a xylophone and a mallet. The robot needs to pick up the mallet with one hand and strike all the xylophone keys from left to right.}

\taskitem{Data Source}{AutoGen}

\taskitem{Usage}{Train \& Eval}
}

\taskshowcase
{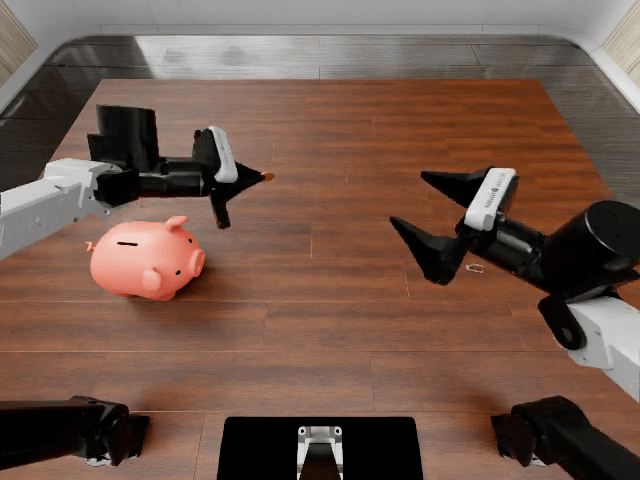}
{deposit\_coin}
{
\taskitem{Instruction}{
Pick up the coin from the holder and insert it precisely into the coin bank.
}

\taskitem{Description}{There is a coin placed on a holder and a coin bank. The robot needs to pick up the coin and accurately insert it into the slot of the coin bank.}

\taskitem{Data Source}{AutoGen}

\taskitem{Usage}{Train \& Eval}
}

\taskshowcase
{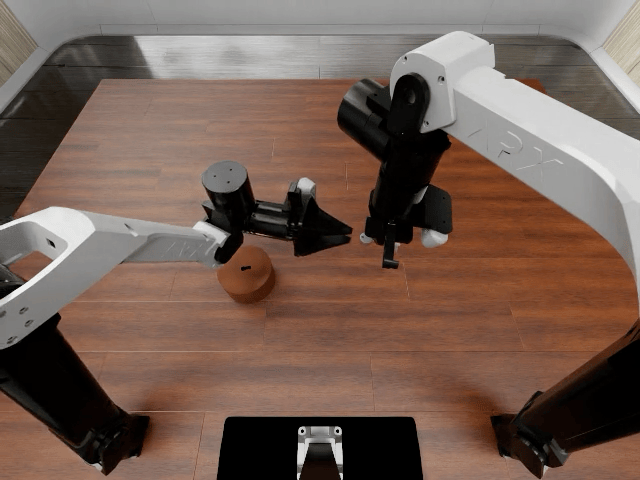}
{insert\_key}
{
\taskitem{Instruction}{
Pick up the thick card, hand it over to the other hand to adjust its pose, then insert it into the card slot.
}

\taskitem{Description}{There is a thick card and a card slot on the table. The robot needs to pick up the card, hand it over to the other hand for pose adjustment, and then insert it accurately into the card slot.}

\taskitem{Data Source}{AutoGen}

\taskitem{Usage}{Train \& Eval}
}

\taskshowcase
{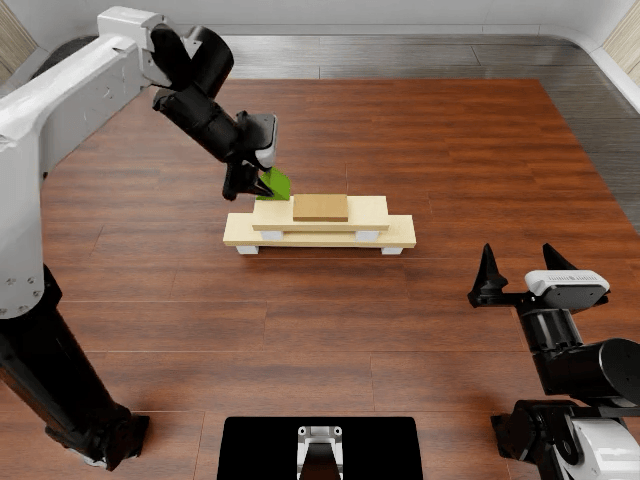}
{build\_tower}
{
\taskitem{Instruction}{
Build a tower using the wooden blocks and wooden boards.
}

\taskitem{Description}{There are wooden blocks and wooden boards on the table. The robot needs to use them to build a stable tower structure.}

\taskitem{Data Source}{AutoGen}

\taskitem{Usage}{Train \& Eval}
}

\subsection{Long-Horizon}

\taskshowcase
{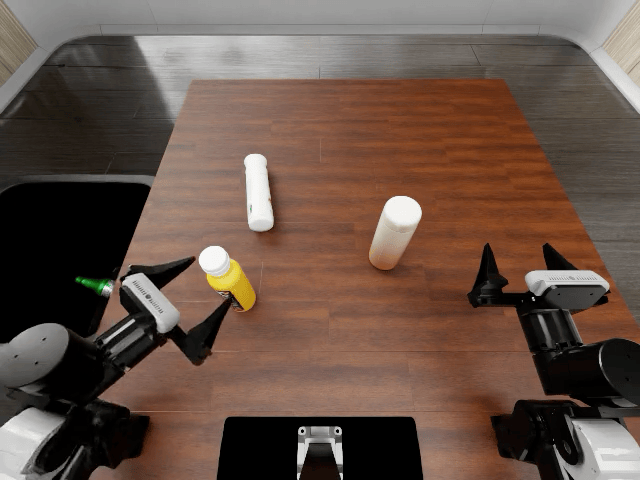}
{put\_bottles\_into\_dustbin}
{
\taskitem{Instruction}{
Pick up the bottles and throw them into the dustbin, using handover when needed.
}

\taskitem{Description}{There are four bottles on the table, and they may be either standing or lying down. A dustbin is placed beside the table. The robot needs to pick up the bottles and throw them into the dustbin. Because of the shifted table layout, bottles on the right side require a handover between the two hands before being discarded.}

\taskitem{Data Source}{Teleop}

\taskitem{Usage}{Train \& Eval}
}

\taskshowcase
{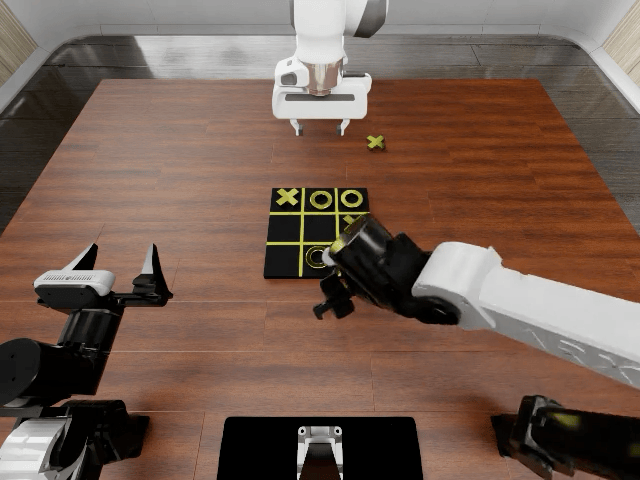}
{play\_tic\_tac\_toe}
{
\taskitem{Instruction}{
Play tic-tac-toe as the first player and fill the board with the opponent.
}

\taskitem{Description}{There is a 3-by-3 tic-tac-toe board. The robot plays as the first player, while the opponent follows a random strategy. The robot and the opponent take turns placing their marks until the board is filled.}

\taskitem{Data Source}{AutoGen}

\taskitem{Usage}{Train \& Eval}
}

\taskshowcase
{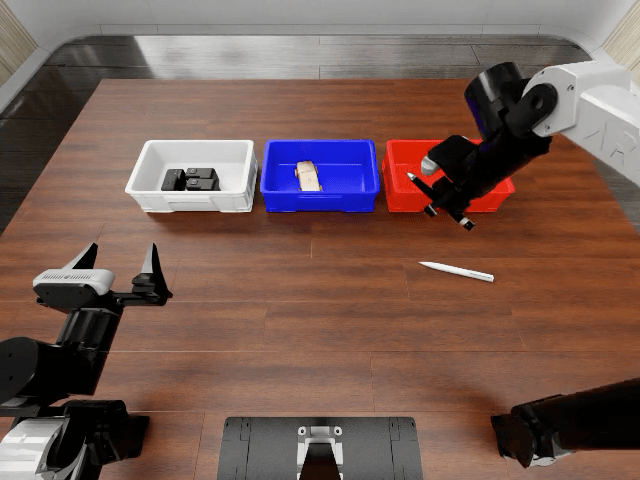}
{classify\_objects}
{
\taskitem{Instruction}{
Sort the objects by category into the three baskets.
}

\taskitem{Description}{There are three categories of objects and three baskets. The robot needs to group the objects by category and place each category into a separate basket. Any basket can be used for any category, as long as objects of the same category are placed together.}

\taskitem{Data Source}{Teleop}

\taskitem{Usage}{Train \& Eval}
}

\taskshowcase
{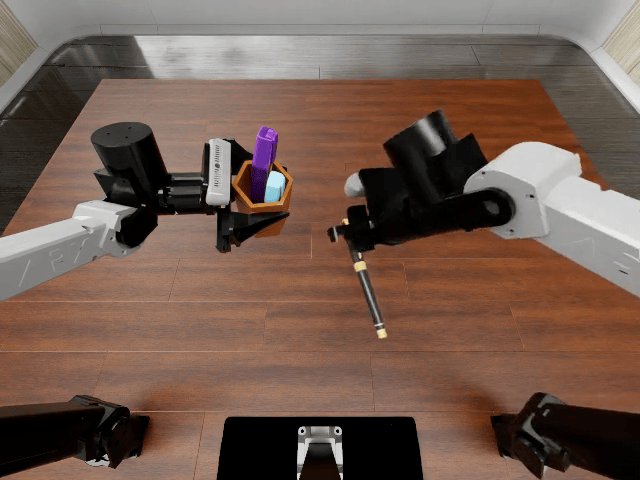}
{fill\_pen\_holder}
{
\taskitem{Instruction}{
Hold the pen holder with one hand, place all pens into it with the other hand, then put it back down.
}

\taskitem{Description}{There is a pen holder and several pens. The robot needs to grasp the pen holder with one hand, use the other hand to place the pens into the holder one by one, and finally put the filled pen holder back on the table.}

\taskitem{Data Source}{Teleop}

\taskitem{Usage}{Train \& Eval}
}

\taskshowcase
{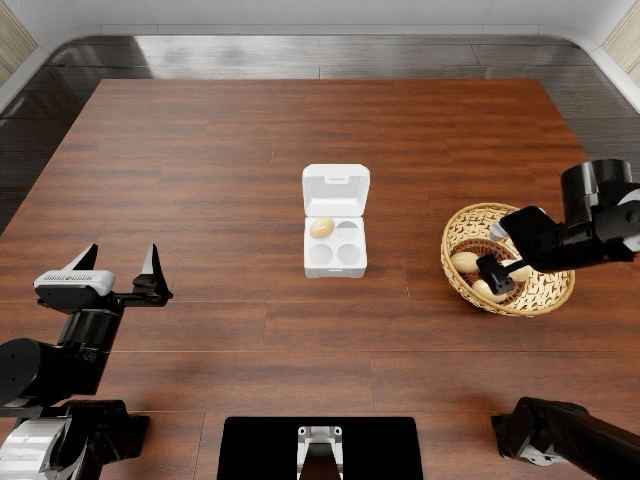}
{fill\_egg\_holder}
{
\taskitem{Instruction}{
Place the four eggs from the basket into the egg holder, then close the lid.
}

\taskitem{Description}{There is a woven basket containing four eggs and an egg holder. The robot needs to pick up the eggs one by one, place all four into the egg holder, and then close the lid.}

\taskitem{Data Source}{Teleop}

\taskitem{Usage}{Train \& Eval}
}

\taskshowcase
{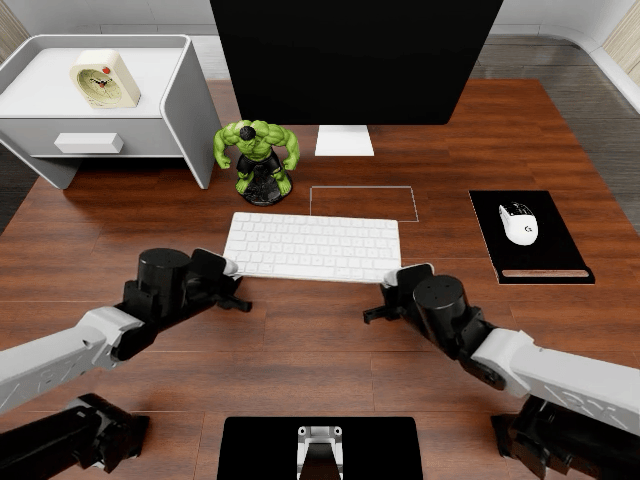}
{organize\_table}
{
\taskitem{Instruction}{
Place the mouse on the mouse pad, push the keyboard into the frame, put the figurine on the stand, place the alarm clock on the drawer, then open the drawer and put all remaining miscellaneous items inside.
}

\taskitem{Description}{There are a computer, a keyboard, a mouse, an alarm clock, a cartoon figurine, three miscellaneous items, and a drawer. The robot needs to organize the table by placing the mouse on the mouse pad, pushing the keyboard into the frame, putting the figurine on the stand, placing the alarm clock on top of the drawer, opening the drawer, and putting all remaining miscellaneous items into it.}

\taskitem{Data Source}{Teleop}

\taskitem{Usage}{Train \& Eval}
}

\taskshowcase
{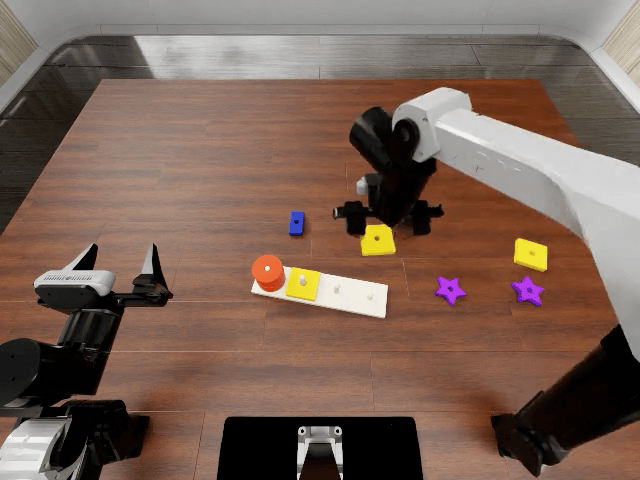}
{play\_stacking\_toy}
{
\taskitem{Instruction}{
Place all stacking toy pieces onto the correct pegs.
}

\taskitem{Description}{There is a stacking toy with four pegs and four types of pieces. The numbers of pieces in the four types are 4, 3, 2, and 1, and each peg matches one type. The robot needs to place all pieces onto their corresponding pegs correctly.}

\taskitem{Data Source}{AutoGen}

\taskitem{Usage}{Train \& Eval}
}

\taskshowcase
{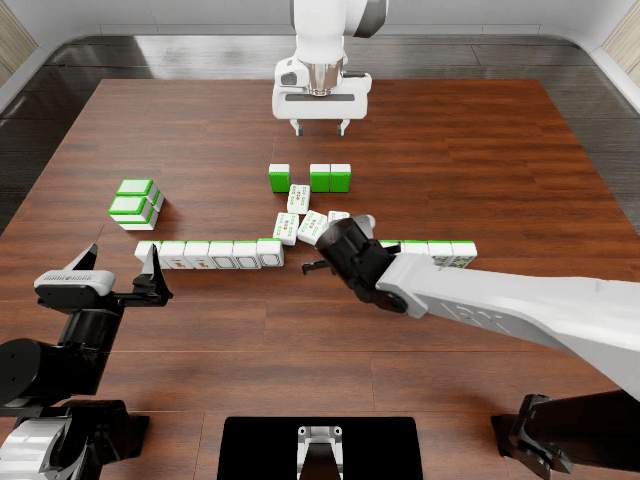}
{make\_kong}
{
\taskitem{Instruction}{
Wait for the opponent to discard a tile, then declare a kong with the matching tiles.
}

\taskitem{Description}{There is a Mahjong setup. The opponent first pushes out a tile. The robot needs to observe the discarded tile, identify the matching tiles on its side, and perform a valid kong action. The setup guarantees that a kong is possible.}

\taskitem{Data Source}{AutoGen}

\taskitem{Usage}{Train \& Eval}
}

\subsection{DLC}

\taskshowcase
{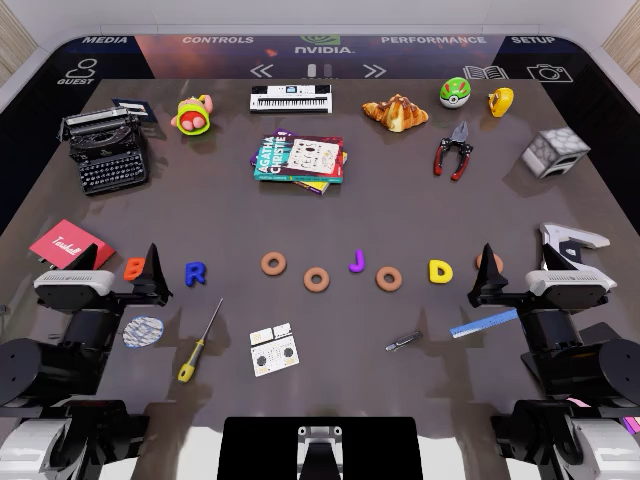}
{dlc}
{
\taskitem{Instruction}{
Arrange the letters to spell "RoboDojo" in a row.
}

\taskitem{Description}{There are multiple separated letters on a cluttered tabletop with a complex background. The robot needs to identify the letters that form "RoboDojo" and arrange them in the correct order in a straight row.}

\taskitem{Data Source}{null (train-only)}

\taskitem{Usage}{Train}
}

%% file: table/real_task_details.tex
\subsection{ARX X5}

\taskshowcase
{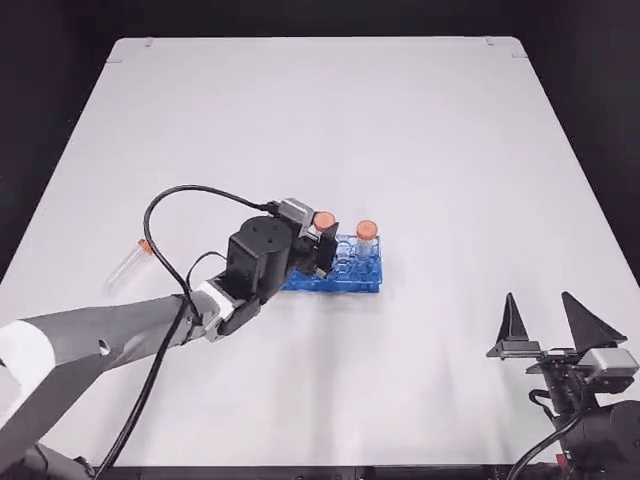}
{insert\_tubes}
{
\taskitem{Instruction}{
Pick up the test tubes on the table and insert them into the test tube rack.
}

\taskitem{Description}{There are multiple test tubes and a test tube rack on the table. The robot needs to pick up the test tubes one by one and insert them into the slots of the test tube rack.}

\taskitem{Data Source}{Teleop}

\taskitem{Usage}{Train \& Eval}
}

\taskshowcase
{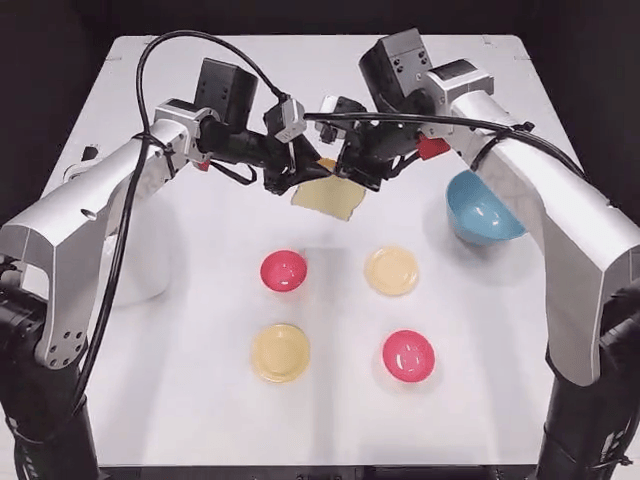}
{make\_bread}
{
\taskitem{Instruction}{
Pick up the two slices of bread from the bowl and place them into the toaster, then place the two small bowls on the plate.
}

\taskitem{Description}{There are two slices of bread in a bowl, a toaster, two small bowls, and a plate on the table. The robot needs to pick up the bread slices and place them into the toaster, then pick up the two small bowls and place them onto the plate.}

\taskitem{Data Source}{Teleop}

\taskitem{Usage}{Train \& Eval}                                                                             
}

\taskshowcase
{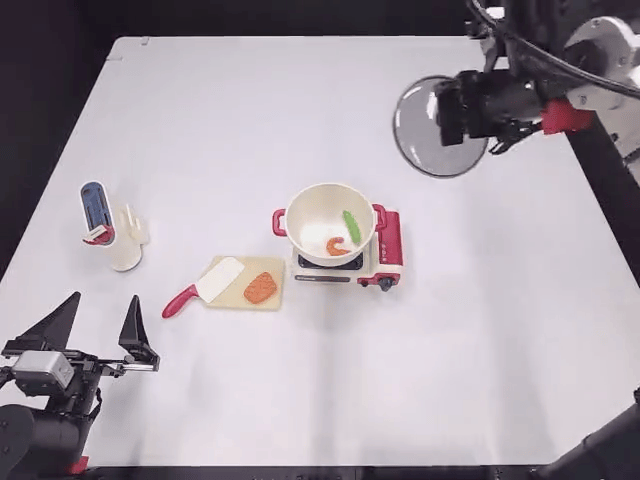}
{make\_food}
{
\taskitem{Instruction}{
Take the cutting board out and place it on the table. Then place the steak and the knife on the cutting board, put the vegetables and shrimp into the pot, place the pot on the stove, and finally cover it with the lid.
}

\taskitem{Description}{There is a cutting board, a steak, a knife, vegetables, shrimp, a pot, a stove, and a lid in the scene. The robot needs to place the cutting board on the table, arrange the steak and knife on it, put the vegetables and shrimp into the pot, move the pot onto the stove, and finally cover the pot with the lid.}

\taskitem{Data Source}{Teleop}

\taskitem{Usage}{Train \& Eval}
}

\taskshowcase
{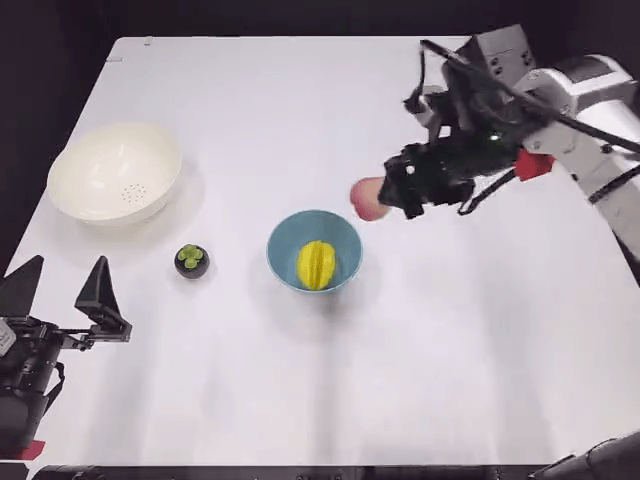}
{pack\_and\_pour\_fruit}
{
\taskitem{Instruction}{
Place all the fruits into the blue bowl, then pour the fruits from the blue bowl into the large white bowl.
}

\taskitem{Description}{There are several fruits, a blue bowl, and a large white bowl on the table. The robot needs to pick up all the fruits and place them into the blue bowl, then transfer them by pouring the fruits from the blue bowl into the large white bowl.}

\taskitem{Data Source}{Teleop}

\taskitem{Usage}{Train \& Eval}
}

\taskshowcase
{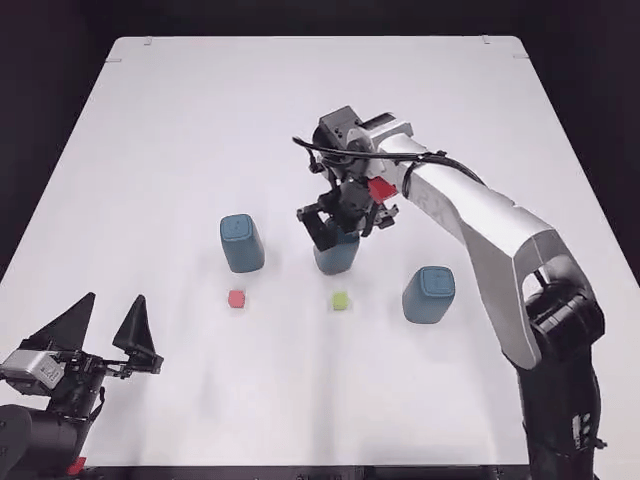}
{cover\_blocks}
{
\taskitem{Instruction}{
Cover the three blocks from left to right, then uncover them again in the order of red, green, and blue.
}

\taskitem{Description}{There are three lids and three colored blocks on the table: red, green, and blue. The robot needs to first cover the blocks from left to right with the lids, memorize the correspondence between the lids and the block colors, and then uncover them again in the order of red, green, and blue.}

\taskitem{Data Source}{Teleop}

\taskitem{Usage}{Train \& Eval}
}

\taskshowcase
{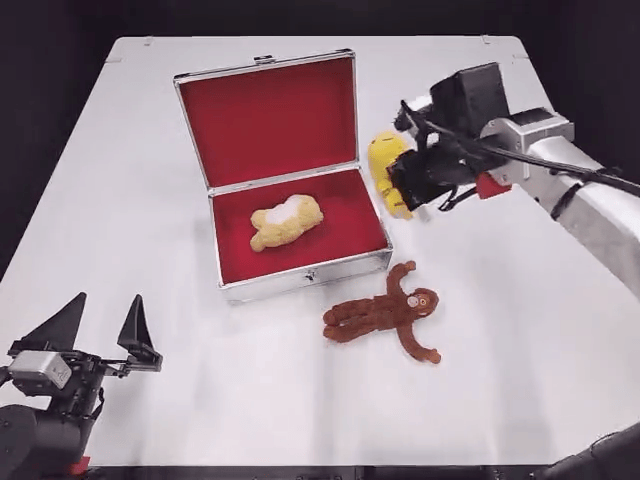}
{store\_in\_safe}
{
\taskitem{Instruction}{
Place all the objects into the safe, then close the safe.
}

\taskitem{Description}{There are multiple objects on the table and a safe nearby. The robot needs to pick up all the objects one by one, place them into the safe, and then close the safe.}

\taskitem{Data Source}{Teleop}

\taskitem{Usage}{Train \& Eval}
}

\subsection{Piper X}

\taskshowcase
{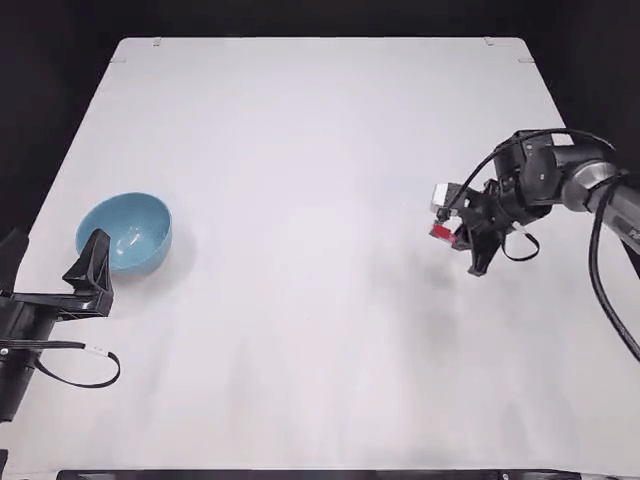}
{disassemble\_LEGO}
{
\taskitem{Instruction}{
Take apart the assembled LEGO structure and place all the pieces into the bowl.
}

\taskitem{Description}{There is an assembled LEGO structure on the table and a bowl nearby. The robot needs to disassemble the LEGO structure and place all the separated pieces into the bowl.}

\taskitem{Data Source}{Teleop}

\taskitem{Usage}{Train \& Eval}
}

\taskshowcase
{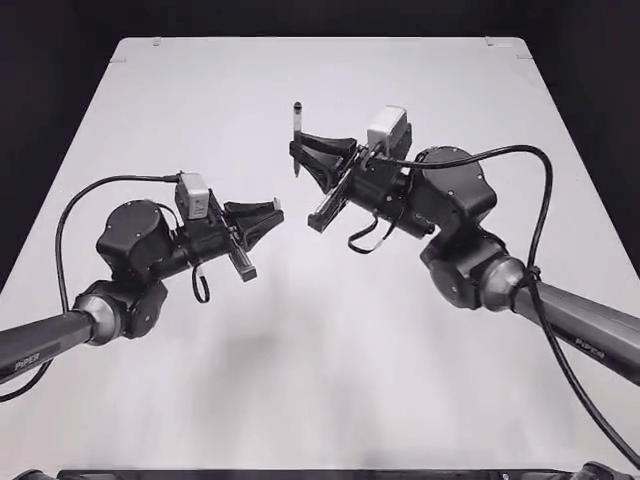}
{cap\_pen}
{
\taskitem{Instruction}{
Pick up the pen and the pen cap, then put the cap on the pen.
}

\taskitem{Description}{There is a pen and a pen cap on the table. The robot needs to pick up both the pen and the cap, align them properly, and put the cap onto the pen.}

\taskitem{Data Source}{Teleop}

\taskitem{Usage}{Train \& Eval}
}

\taskshowcase
{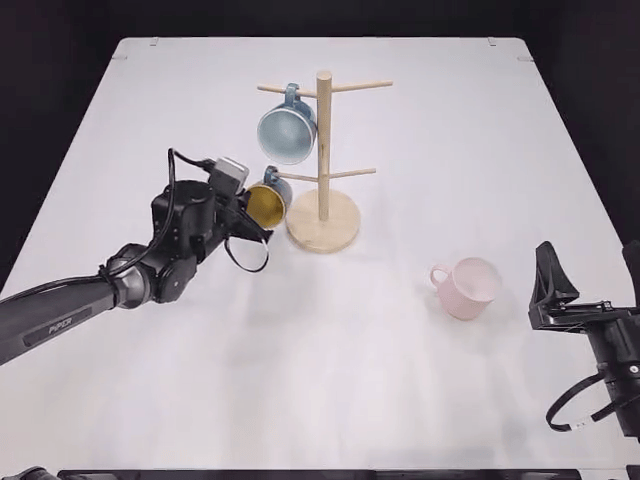}
{hang\_mugs}
{
\taskitem{Instruction}{
Hang the mugs on the mug rack.
}

\taskitem{Description}{There are several mugs and a mug rack on the table. The robot needs to pick up each mug and hang it on the mug rack.}

\taskitem{Data Source}{Teleop}

\taskitem{Usage}{Train \& Eval}
}

\taskshowcase
{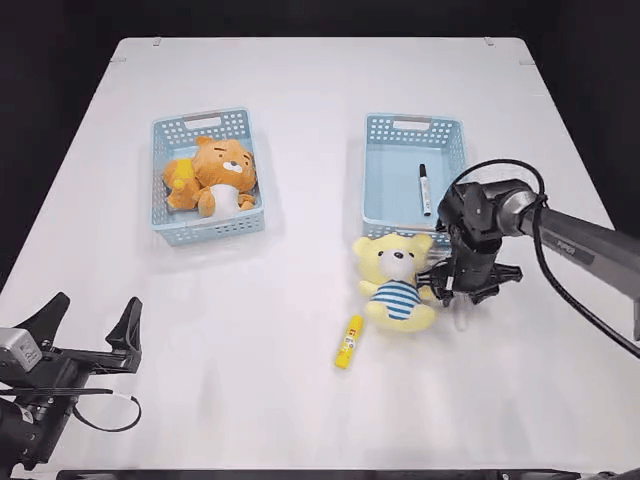}
{classify\_objects}
{
\taskitem{Instruction}{
Classify the two types of objects on the table and place them into the two baskets.
}

\taskitem{Description}{There are two categories of objects and two baskets on the table. The robot needs to identify the category of each object, classify them accordingly, and place each type into the correct basket.}

\taskitem{Data Source}{Teleop}

\taskitem{Usage}{Train \& Eval}
}

\taskshowcase
{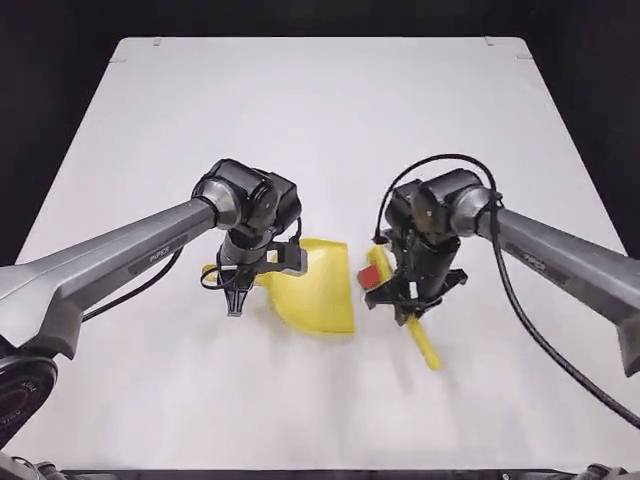}
{sweep\_blocks}
{
\taskitem{Instruction}{
Pick up the broom, hand it over to the right hand, then use the dustpan to sweep the blocks.
}

\taskitem{Description}{There is a broom and a dustpan on the left side, and the blocks are on the right side. The robot needs to first pick up the broom, hand it over to the right hand, then grasp the dustpan with the left hand, and finally sweep the blocks successfully.}

\taskitem{Data Source}{Teleop}

\taskitem{Usage}{Train \& Eval}
}

\taskshowcase
{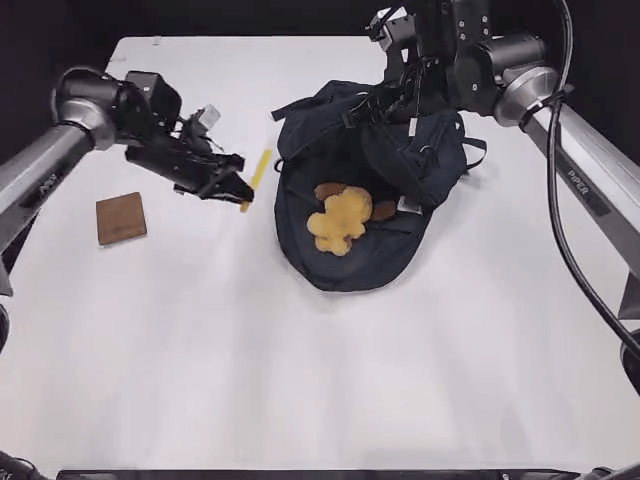}
{pack\_objects\_into\_backpack}
{
\taskitem{Instruction}{
Place all the objects on the table into the backpack.
}

\taskitem{Description}{There are multiple objects on the table and a backpack nearby. The robot needs to pick up all the objects one by one and place them into the backpack.}

\taskitem{Data Source}{Teleop}

\taskitem{Usage}{Train \& Eval}
}

\subsection{Piper}

\taskshowcase
{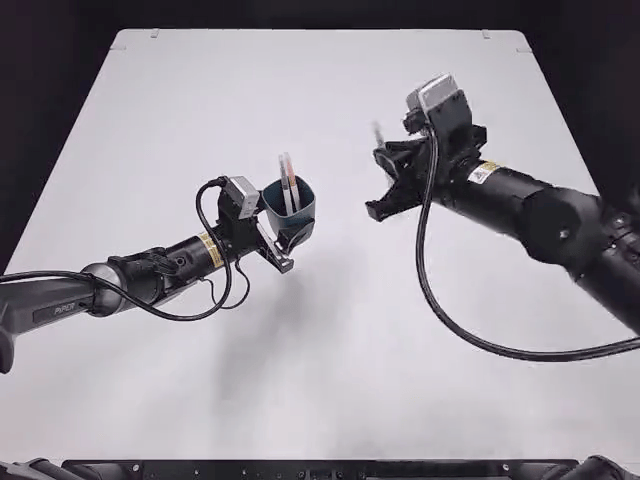}
{fill\_pen\_holder}
{
\taskitem{Instruction}{
Pick up the pen holder and place all the pens into it.
}

\taskitem{Description}{There is a pen holder and several pens on the table. The robot needs to pick up the pen holder and place all the pens into it one by one.}

\taskitem{Data Source}{Teleop}

\taskitem{Usage}{Train \& Eval}
}

\taskshowcase
{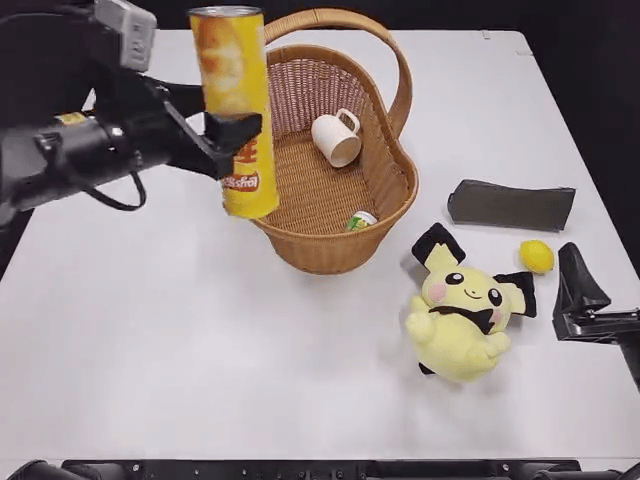}
{put\_objects\_into\_basket}
{
\taskitem{Instruction}{
Place all the objects on the table into the basket.
}

\taskitem{Description}{There are multiple objects on the table and a basket nearby. The robot needs to pick up all the objects one by one and place them into the basket.}

\taskitem{Data Source}{Teleop}

\taskitem{Usage}{Train \& Eval}
}

\taskshowcase
{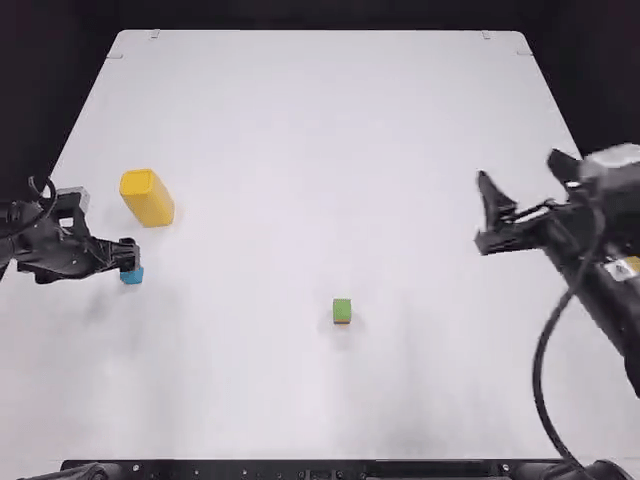}
{stack\_and\_cover\_blocks}
{
\taskitem{Instruction}{
Stack the blocks on the table, then cover them with the cup.
}

\taskitem{Description}{There are several blocks and a cup on the table. The robot needs to stack the blocks into a pile and then place the cup over them to cover the stacked blocks.}

\taskitem{Data Source}{Teleop}

\taskitem{Usage}{Train \& Eval}
}

\taskshowcase
{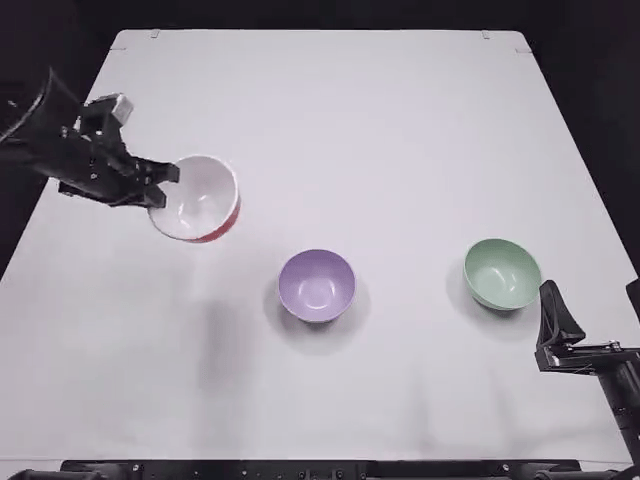}
{stack\_bowls}
{
\taskitem{Instruction}{
Stack the bowls on the table.
}

\taskitem{Description}{There are several bowls on the table. The robot needs to pick them up one by one and stack them neatly.}

\taskitem{Data Source}{Teleop}

\taskitem{Usage}{Train \& Eval}
}

\taskshowcase
{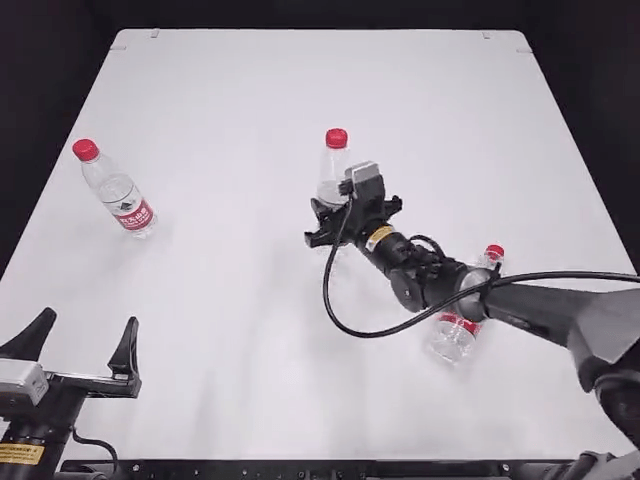}
{stand\_up\_bottles}
{
\taskitem{Instruction}{
Stand the bottle upright.
}

\taskitem{Description}{There is a bottle lying on the table. The robot needs to pick it up and place it upright on the table.}

\taskitem{Data Source}{Teleop}

\taskitem{Usage}{Train \& Eval}
}

\taskshowcase
{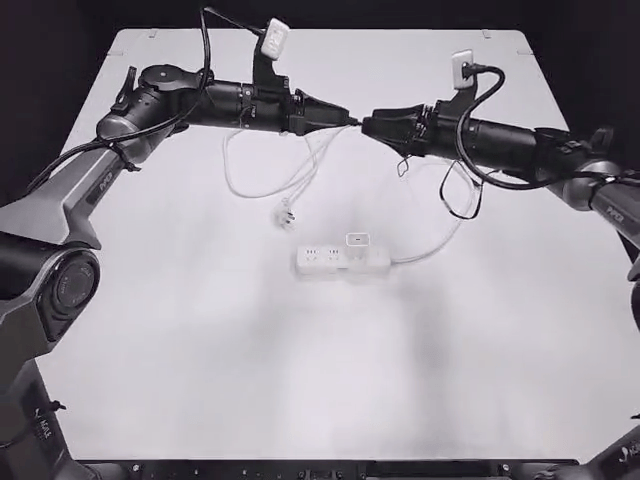}
{insert\_charger}
{
\taskitem{Instruction}{
Insert the charger plug into the power strip, then connect the charging cable to the plug.
}

\taskitem{Description}{There is a charger plug, a charging cable, and a power strip on the table. The robot needs to insert the charger plug into the power strip and then connect the charging cable to the plug.}

\taskitem{Data Source}{Teleop}

\taskitem{Usage}{Train \& Eval}
}

%% file: sec/benchmark_experiment/train_details.tex
\textbf{Hy-Embodied-0.5-VLA.}
We fine-tune Hy-Embodied-0.5-VLA from the foundation checkpoint hosted at \url{https://huggingface.co/tencent/Hy-Embodied-0.5-VLA-UMI}.
For simulation experiments, we follow its official open-source evaluation setting on RoboTwin 2.0, training the model with a batch size of 128 for 200K steps and using 6 historical images sampled every 20 steps for memory encoding.

\textbf{Spatial Forcing.}
We fine-tune Spatial Forcing from the \texttt{gs://openpi-assets/checkpoints/pi05\_base} foundation checkpoint.
For simulation experiments, the model is trained with a batch size of 256 for 60K steps.

\textbf{$\pi_{0.5}$.}
We fine-tune $\pi_{0.5}$ from the \texttt{gs://openpi-assets/checkpoints/pi05\_base} foundation checkpoint.
For simulation experiments, the model is trained with a batch size of 256 for 60K steps.
For real-world experiments, we use the same batch size and train the model for 30K steps.

\textbf{X-VLA.}
We fine-tune X-VLA from the \texttt{2toInf/X-VLA-Pt} foundation checkpoint.
For simulation experiments, the model is trained with a batch size of 256 for 100K steps.
For real-world experiments, we use the same batch size and train the model for 50K steps.

\textbf{Xiaomi-Robotics-0.}
We fine-tune Xiaomi-Robotics-0 from the \texttt{Xiaomi-Robotics-0-Pretrain} foundation checkpoint.
For simulation experiments, the model is trained with a batch size of 256 for 100K steps.
For real-world experiments, we use the same batch size and train the model for 50K steps.

\textbf{X-WAM.}
We fine-tune X-WAM from the \texttt{Wan2.2-TI2V-5B} foundation checkpoint.
For simulation experiments, the model is trained with a batch size of 32 for 40K steps.

\textbf{GigaWorld-Policy.}
We fine-tune GigaWorld-Policy from the \texttt{Wan2.2-TI2V-5B} foundation checkpoint.
For simulation experiments, the model is trained with a batch size of 32.
The video optimization stage is trained for 50K steps (1 epoch), followed by the action optimization stage for 250K steps (2 epochs), resulting in 300K total optimization steps (3 epochs).

\textbf{StarVLA-$\alpha$.}
We fine-tune StarVLA-$\alpha$ from the \texttt{Qwen3-VL-4B-Instruct} foundation checkpoint.
For simulation experiments, the model is trained with a batch size of 128 for 100K steps.
For real-world experiments, we use the same batch size and train the model for 30K steps.

\textbf{GalaxeaVLA.}
We fine-tune GalaxeaVLA from the \texttt{G0Plus\_3B-base} foundation checkpoint.
For simulation experiments, the model is trained with a batch size of 32 for 4 epochs.
For real-world experiments, we use the same batch size and train the model for 4 epochs.

\textbf{LingBot-VLA.}
We fine-tune LingBot-VLA from the \texttt{lingbot-vla-4b} foundation checkpoint.
For simulation experiments, the model is trained with a batch size of 256 for 15K steps.

\textbf{EventVLA.}
We fine-tune EventVLA from the \texttt{Qwen3-VL-4B-Instruct} foundation checkpoint.
For simulation experiments, the model is trained with a batch size of 128 for 150K steps.

\textbf{Fast-WAM.}
We fine-tune Fast-WAM from the \texttt{Wan2.2-TI2V-5B} foundation checkpoint.
For simulation experiments, the model is trained with a batch size of 256 for 20K steps.

\textbf{AHA-WAM.}
We fine-tune AHA-WAM from the \texttt{Wan2.2-TI2V-5B} foundation checkpoint.
For simulation experiments, the model is trained with a batch size of 768 for 25K steps.

\textbf{$\pi_0$.}
We fine-tune $\pi_0$ from the \texttt{gs://openpi-assets/checkpoints/pi0\_base} foundation checkpoint.
For simulation experiments, the model is trained with a batch size of 256 for 60K steps.
For real-world experiments, we use the same batch size and train the model for 30K steps.

\textbf{GR00T-N1.7.}
We fine-tune GR00T-N1.7 from the \texttt{nvidias/GR00T-N1.7-3B} foundation checkpoint.
For simulation experiments, the model is trained with a batch size of 640 for 100K steps.
For real-world experiments, we use the same batch size and train the model for 50K steps.

\textbf{InternVLA-A1.}
We fine-tune InternVLA-A1 from the \texttt{InternVLA-A1-3B} foundation checkpoint.
For simulation experiments, the model is trained with a batch size of 64 for 60K steps.
For real-world experiments, we use the same batch size and train the model for 30K steps.

\textbf{SmolVLA (Single Task).}
We fine-tune SmolVLA (Single Task) from the \texttt{smolvla\_base} foundation checkpoint.
For simulation experiments, the model is trained with a batch size of 512 for 100K steps.

\textbf{GO-1.}
We fine-tune GO-1 from the \texttt{GO-1} foundation checkpoint.
For simulation experiments, the model is trained with a batch size of 96 for 4 epochs, corresponding to 77,484 optimization steps.

\textbf{LDA-1B.}
We fine-tune LDA-1B from the \texttt{LDA-pretrain} foundation checkpoint.
For simulation experiments, the model is trained with a batch size of 128 for 300K steps.

\textbf{MolmoAct2.}
We fine-tune MolmoAct2 from the \texttt{MolmoAct2} foundation checkpoint.
For simulation experiments, the model is trained with a batch size of 128 for 100K steps.

\textbf{ACT (Single Task).}
We train ACT (Single Task) from scratch.
For simulation experiments, the model is trained with a batch size of 128 for 6K steps.

\textbf{H-RDT.}
We fine-tune H-RDT from the \texttt{pretrain-0618/checkpoint-500000} foundation checkpoint.
For simulation experiments, the model is trained with a batch size of 256 for 90K steps.

\textbf{Spirit v1.5.}
We fine-tune Spirit v1.5 from the \texttt{Spirit-v1.5} foundation checkpoint.
For simulation experiments, the model is trained with a batch size of 256 for 50K steps.
For real-world experiments, we use the same batch size and train the model for 30K steps.

\textbf{RDT-1B.}
We fine-tune RDT-1B from the \texttt{rdt-1b} foundation checkpoint.
For simulation experiments, the model is trained with a batch size of 256 for 200K steps.

\textbf{DM0.}
We fine-tune DM0 from the \texttt{DM0-base} foundation checkpoint.
For simulation experiments, the model is trained with a batch size of 32 for 100K steps.
For real-world experiments, we use the same batch size and train the model for 50K steps.